%% file: ms.tex
\documentclass[arxiv,nameyear,final]{econsocart}
\usepackage{xcolor}
\usepackage[most]{tcolorbox}
\usepackage{etoolbox}
\usepackage[skip=4pt]{caption}

\newif\ifshowchanges
\showchangestrue   %  ← flip to false for the clean PDF

\tcbset{
  changebox/.style={
    breakable, enhanced, sharp corners, boxrule=0pt,
    left=1pt,right=1pt,top=1pt,bottom=1pt,
    fonttitle=\scriptsize\bfseries,
    before upper=\vspace{-0.25em},
  },
}

\newtcolorbox{additionbox}[1][]{changebox,colback=green!20,title={#1}}
\newtcolorbox{deletionbox}[1][]{changebox,colback=red!20,  title={#1}}
\newtcolorbox{movebox}[1][]{changebox,colback=yellow!20,  title={#1}}

\usepackage{booktabs}
\usepackage{xcolor}
\usepackage{multirow}
\usepackage{float}
\usepackage{verbatim}
\usepackage{fancyvrb}
\usepackage{graphicx}
\usepackage{amsmath}
\usepackage{fvextra}
\usepackage{longtable}
\usepackage{tcolorbox}
\usepackage{pdflscape} % or \usepackage{lscape}
\RequirePackage[colorlinks,citecolor=blue,linkcolor=blue,urlcolor=blue,pagebackref]{hyperref}

\startlocaldefs

\newcommand{\texttemplate}[0]{\textsc{Tmpl}}
\newcommand{\tokenize}[0]{\textsc{Tok}}
\newcommand{\jobtitle}[0]{\textsc{Title}}

\newif\ifshowcomments
 \showcommentsfalse   % hide comments  (comments are still in Latex code, just not in the compiled PDF)
\ifshowcomments
    \newcommand{\updated}[1]{\textcolor{red}{Last updated: #1}}
    \newcommand{\tianyu}[1]{\textcolor{cyan}{Tianyu: {#1}}}
    \newcommand{\susan}[1]{\textcolor{pink}{Susan: {#1}}}
\else
    \newcommand{\updated}[1]{}
    \newcommand{\tianyu}[1]{}
    \newcommand{\susan}[1]{}
\fi
\endlocaldefs

\begin{document}

\begin{frontmatter}

\title{LABOR-LLM: Language-Based Occupational Representations with Large Language Models}
\runtitle{LABOR-LLM}

\begin{aug}
\author[id=au1,addressref={add2}]{\fnms{Susan}~\snm{Athey}\ead[label=e1]{athey@stanford.edu}}
\author[id=au2,addressref={add1}]{\fnms{Herman}~\snm{Brunborg}\ead[label=e2]{brunborg@stanford.edu}}
\author[id=au3,addressref={add1}]{\fnms{Tianyu}~\snm{Du}\ead[label=e3]{tianyudu@stanford.edu}}
\author[id=au4,addressref={add4}]{\fnms{Ayush}~\snm{Kanodia}\ead[label=e4]{akanodia@stanford.edu}}
\author[id=au5,addressref={add3}]{\fnms{Keyon}~\snm{Vafa}\ead[label=e5]{kvafa@g.harvard.edu}}

%%%%%%%%%%%%%%%%%%%%%%%%%%%%%%%%%%%%%%%%%%%%%%
%% Addresses                                %%
%%%%%%%%%%%%%%%%%%%%%%%%%%%%%%%%%%%%%%%%%%%%%%
\address[id=add1]{
\orgdiv{Institute for Computational and Mathematical Engineering},
\orgname{Stanford University}}

\address[id=add2]{
\orgdiv{Graduate School of Business},
\orgname{Stanford University}}

\address[id=add3]{
\orgdiv{Harvard Data Science Initiative},
\orgname{Harvard University}}

\address[id=add4]{
\orgname{Independent Researcher}}

\end{aug}

\support{We gratefully acknowledge financial support from the Golub Capital Social Impact Lab, the Stanford Institute for Human-Centered Artificial Intelligence, and the Business, Governance, and Society Initiative at Stanford's Graduate School of Business. Herman Brunborg thanks the Aker Scholarship Foundation for financial support. Author order is alphabetical. Tianyu Du was the student leading the project and would be considered first author in the engineering tradition.}

\begin{abstract}
This paper builds an empirical model that predicts a worker's next occupation as a function of the worker's occupational history. Because histories are sequences of occupations, the covariate space is high-dimensional, and further, the outcome (the next occupation) is a discrete choice that can take on many values. To estimate the parameters of the model, we leverage an approach from generative artificial intelligence. Estimation begins from a ``foundation model'' trained on non-representative data and then ``fine-tunes'' the estimation using data about careers from a representative survey. We convert tabular data from the survey into text files that resemble resumes and fine-tune the parameters of the foundation model, a large language model (LLM), using these text files with the objective of predicting the next token (word). The resulting fine-tuned LLM is used to calculate estimates of worker transition probabilities. Its predictive performance surpasses all prior models, both for the task of granularly predicting the next occupation as well as for specific tasks such as predicting whether the worker changes occupations or stays in the labor force. We quantify the value of fine-tuning and further show that by adding more career data from a different population, fine-tuning smaller LLMs (fewer parameters) surpasses the performance of fine-tuning larger models. When we omit the English language occupational title and replace it with a unique code, predictive performance declines.
\end{abstract}



\end{frontmatter}
%%%%%%%%%%%%%%%%%%%%%%%%%%%%%%%%%%%%%%%%%%%%%%%%%%%%%%%%%%%%%%%%%%%%%%%%%
%%%% Main text entry area:
%%%%%%%%%%%%%%%%%%%%%%%%%%%%%%%%%%%%%%%%%%%%%%%%%%%%%%%%%%%%%%%%%%%%%%%%%

\newpage
\section{Introduction}\label{sec:introduction}
This paper introduces a new approach to predicting the evolution of worker careers, specifically the problem of predicting a worker's next occupation as a function of the worker's prior history of occupations and covariates. The problem is challenging because of the high-dimensional feature space: When there are 335 possible occupations, there are $335^t$ possible sequences of occupations in $t$ periods of observation. In addition, the prediction space is large: a predictive model produces 335 probabilities corresponding to the possible next occupations. Historically, economists have often dealt with this complexity by using heuristics to summarize occupational histories with a small number of summary statistics, e.g. most recent occupational category and years of experience (e.g., \citet{blau_gender_2017}), and by focusing on relatively coarse outcomes, such as employment status, (\citet{boskin_conditional_1974}) or characteristics of occupations (\citet{cortes2016have}).

In this paper, we build a model that predicts the next occupation in a sequence, given a history of preceding occupations, while addressing the challenge of the high-dimensional feature space. Our model delivers substantially better predictive performance for the granular task of predicting the next occupation, as well as for auxiliary tasks commonly considered in labor economics, such as predicting whether a worker becomes unemployed or leaves the labor market.  Better predictive performance translates into better estimators of causal effects, decompositions, and structural models; further, a model with granular predictions allows researchers to answer questions about a richer set of outcomes.

We construct our occupation transition model using an approach popularized in the generative artificial intelligence (AI) literature (\citet{bommasani_opportunities_2022}), where we start with a ``foundation model'' (e.g., a large language model (LLM)) with parameters estimated on a large, unrepresentative dataset, and then move to ``fine-tuning,'' where the parameter estimation process continues on a smaller dataset, so that the final parameter estimates incorporate ``foundational'' knowledge from the larger dataset but are tailored to make accurate predictions consistent with the conditional distributions in the smaller dataset.  Most LLMs as of 2025 are built using flexible parametric models (transformer neural networks) with billions to trillions of parameters. The parameters are estimated by maximizing the conditional log-likelihood of the next word in a text, given the history of preceding words. The estimation uses substantial computing resources on a very large dataset of text, e.g., text scraped from the open Internet. Using these parameters as starting values for fine-tuning retains information about the general meaning of text while also specializing the predictions to match the joint distribution of text in the fine-tuning dataset.

Recently, a series of papers (\citet{vafa_career_2024, vafa2025gendergapPNAS}) took language modeling as an inspiration and built a custom foundation model (referred to as ``CAREER'') for occupation transitions using the transformer neural network functional form popularized by LLMs, showing that CAREER improved accuracy substantially relative to existing econometric models. However, that work treated occupation as a categorical variable and did not make use of the textual content of job titles. In this paper, we introduce an approach that directly uses the next-word probability models from open-weight Large Language Models (LLMs), thus allowing us to effectively make use of a much larger model (more parameters) while incorporating the information encoded in the text, which may help the model understand which occupation histories are similar to one another in terms of likely future occupation transitions. We refer to our approach as the \textbf{LA}nguage-\textbf{B}ased \textbf{O}ccupational \textbf{R}epresentations with \textbf{L}arge \textbf{L}anguage \textbf{M}odels (LABOR-LLM) framework.

In contrast to traditional methods that rely on tabular categorical data (e.g., \citet{vafa_career_2024}), our approach leverages the insight that a sequence of occupations can be represented as a sequence of words via a ``text template.'' In our preferred approach, Fine-Tuned LABOR-LLM (FT-LABOR-LLM), we fine-tune an open-source foundation model (Llama-2, by Meta) on textual representations of worker careers constructed from publicly available U.S. survey datasets, the National Longitudinal Survey of Youth (NLSY) and Panel Study of Income Dynamics (PSID), using a next-word prediction objective.

The FT-LABOR-LLM approach involves using estimates of the probability of the next word (conditional on that word being preceded by a particular sequence of words) from the fine-tuned language model to calculate the probability that a given occupation comes next, given career history. Specifically, we use the probability model associated with the fine-tuned LLM to evaluate the probability that the next text in our text template is the text corresponding to a particular occupation, conditional on the preceding text being equal to the text of the text template truncated at the year of interest.

We show that the performance of FT-LABOR-LLM is better than that of CAREER, despite CAREER being custom-designed for the problem and pre-trained on a very relevant corpus of documents, resumes of U.S. workers. Recalling that CAREER in turn substantially outperformed alternatives from the literature, our results show that FT-LABOR-LLM is state-of-the-art in terms of predictive performance. We further show that not only is FT-LABOR-LLM better at the granular next-occupation prediction problem, but it also performs better when applied to predict specific, commonly studied outcomes such as changing occupations, becoming unemployed, or leaving paid employment.  We highlight the importance of the fine-tuning step by showing that, without fine-tuning, off-the-shelf Llama-2 makes plausible-sounding predictions of occupations, but it is not as accurate in terms of the next occupation probability distributions conditional on history, and it ``hallucinates'' invalid job titles because it is not fine-tuned exclusively on labor sequence data. Similar problems arose even with the commercial LLMs that were state-of-the-art at the time our analysis was conducted in 2025.

In the remainder of the paper, we assess the sources of the performance benefits. We begin by assessing the role of model size (number of parameters) and the volume of data.  We show that using a larger LLM as the foundation model, in particular the version of Llama-2 with 13 billion parameters rather than the version with 7 billion parameters, improves predictive performance. However, we show that adding in data from different government surveys (even though they are drawn from different time periods) quickly improves the performance of the smaller model, matching and then surpassing the performance of the larger model. Thus, data (even if non-representative) is a substitute for model size.\footnote{Other papers have shown that more data improves model performance for both pre-training (\citet{vafa_career_2024}, \citet{kaplan2020scaling}) and fine-tuning (\citet{dong2023abilities}, \citet{bucher2024fine}) data.} Since smaller models are less expensive to estimate, and especially cheaper to make predictions from, working with a smaller model has distinct advantages.  More broadly, an open question about the fine-tuning approach is whether the fact that the pre-training dataset is not representative of the target implies that the final estimated model will exhibit bias relative to the true conditional transition probabilities in the population of interest. There may be a tradeoff between using a large, non-representative dataset to better learn underlying structure (e.g., meaning of language, or general information about occupation transitions), and getting a model that makes conditional predictions that are representative of a target dataset of interest. In this paper, we show that if such biases are important, the advantages of the foundation model approach outweigh them in our application, and further, adding non-representative data about occupation transitions in the fine-tuning step improves performance further. The best approach to combining representative and non-representative data in fine-tuning provides an interesting opportunity for future research.

We next assess whether FT-LABOR-LLM is making use of information embedded in the text of the job title. To do so, we replace the job titles with numeric codes in the training data and show that this approach degrades predictive performance substantially. We further establish that demographics, most notably gender, but also the interaction of gender, ethnicity, and region, play an important role in predicting occupation transitions.  Finally, we show that the model's performance is highest when workers' full history is included; predictive performance is degraded unless at least 10 periods of worker history are included.

We also compare our approach to alternative ways we could use a fine-tuned model to make predictions, showing that the alternatives do not perform as well as FT-LABOR-LLM. One alternative approach involves using it to create data-driven low-dimensional embeddings of history (latent factors), and then using those embeddings as if they were observed covariates in a multinomial logistic regression where occupations are the alternatives.

Overall, the success of FT-LABOR-LLM provides an example of how LLMs can be used as foundation models for an economic problem that was traditionally studied using categorical, discrete-choice prediction models. In addition to providing superior predictive performance, the LABOR-LLM approach has some advantages because the pre-training step does not have to be carried out by the individual researcher; rather, open, general-purpose LLMs can be used (or closed models can be used through paid API access, although with less control on the part of the analyst). Since the data used for fine-tuning is textual and does not require a fixed format for either estimation or prediction, the approach allows the researcher to add in additional textual information for some or all observations in either the training set or the test set without changing any code. The approach handles missing data, unbalanced panels, irregular sampling intervals for individuals, and textual information about workers or their jobs, where the only required changes to the pipeline involve modifying the text included in the training or testing data.

How can FT-LABOR-LLM be used? We highlight three categories of economic applications that benefit from a high-quality occupation transition model: estimation of causal effects, decompositions, and structural estimation. Consider first causal effects. Under the assumptions required for identification of average treatment effects with many commonly considered statistical designs (e.g., unconfoundedness, instrumental variables, or panel data), estimating causal effects for a treated group requires the prediction of counterfactual outcomes in the absence of the treatment. The theory of semi-parametric efficiency in estimation (e.g. \citet{chernozhukov2018double}) shows that the performance of efficient treatment effect estimators is determined by the performance of estimators for auxiliary prediction problems, notably estimators of outcome models and treatment assignment (propensity scores) as functions of covariates. Better predictive models lead to more precise estimates and avoid omitted variable bias, and the convergence rates of these estimates must be sufficiently fast to guarantee asymptotic normality.\footnote{In practice, it is common to use cross-validation based on prediction quality to select outcome and propensity models. A large literature compares the performance of estimators using different approaches to outcome modeling and propensity weighting; see, e.g., \citet{athey2018residual} and \citet{athey2024gan} and references therein.} \citet{vafa2025gendergapPNAS} refines the standard results for the case where embedding functions are used to reduce dimensionality while avoiding omitted variable bias induced by the dimension reduction.

In our context, predicting the next occupation may be useful for estimating causal effects in several scenarios: (i) The occupation (or a categorization of the occupation) may be the outcome, e.g. taking a desirable occupation, becoming unemployed (\citet{del2021occupational}, \citet{dabed2023resilience}), entering poverty (\citet{stevens1994dynamics}),
entering education (\citet{frederiksen2007did}), or transitioning industries (\citet{schmidt_prediction_1975}, \citet{dauth2018adjusting}) Future work may use the more granular predictions enabled by FT-LABOR-LLM to ask questions such as, which types of occupational transitions become more likely as a result of an intervention, and which become less likely?  (ii) The occupation (or a categorization of the occupation) may be the ``treatment,'' e.g., researchers may study the effect of being laid off (\citet{jacobson_earnings_1993}, \citet{athey2023heterogeneous}), out of the labor force (\citet{mincer_and_ofek_1982_interrupted_work_careers}), in education (\citet{card1999causal_effect_of_education_on_earnings}), in a union occupation (\citet{oaxaca1975estimation,lewis1986union}), or in an AI-impacted industry or occupation (\citet{webb2019impact}). (iii) As shown in \citet{vafa_career_2024}, if the outcome of interest is wages or another continuous outcome, next-occupation prediction models can still play a supporting role.

Second, accurate next-occupation prediction models are central to a variety of decomposition problems. The decomposition literature (see, e.g., \citet{blau_gender_2017} for a review) asks, how much of the difference in outcomes between two groups can be explained by differences in observable characteristics?  Although the economic interpretation is distinct, the theory of estimation of decomposition terms is isomorphic to estimating causal effects, and so the theory of semi-parametric estimators also implies that high-quality prediction models play a central role in decompositions. Further, omitted variable bias is a major concern in decompositions; \citet{vafa2025gendergapPNAS} demonstrates that traditional models based on heuristics leave out elements of occupation history and only adjust for current occupation, omitting factors that are correlated with both wages and gender.\footnote{For example, two workers whose occupational category is ``manager'' may have different wages but also differ as to whether their career history includes being an engineer or an administrative assistant.}  The literature studies both wages and occupation-related outcomes, e.g., being unemployed (\citet{fairlie_emergence_1999}) or being employed in a particular industry (\citet{schmidt_prediction_1975}). The literature studies decompositions of differences between groups of workers, which may be differentiated by gender (\citet{brown1980incorporating}), race (\citet{fairlie_emergence_1999}), industry (\citet{abowd2012persistent_industry}), education (\citet{hotchkiss2011decomposing_education}), union status (\citet{oaxaca1975estimation, freeman1984unions,lewis1986union}), or other occupation characteristics.  As discussed above, econometric theory tells us that the accurate prediction of, e.g., being in a unionized job is important for efficient estimation of decompositions.  A future study with access to an accurate and granular occupation prediction model might decompose the differences in outcomes for workers transitioning out of AI-affected occupations.

Third, accurate prediction models can be an input for other types of applied econometric models. Some studies rely on counterfactual simulations of occupational transitions to support structural estimates of the impact of factors such as race (\citet{brown1980incorporating}) or concentration (\citet{,schubert2024employer}) on worker outcomes. Many dynamic structural models of labor markets also rely on predictions of worker job transitions among several categories of occupations and employment  (\citet{keane1997career,sullivan2010dynamic}).
In the context of recommendation systems or automated job advice (\citet{de_ruijt_job_2021}), accurate estimates of conditional transition probabilities may be a key building block.

\section{Related Work} \label{sec:related_work}

\paragraph*{Career Trajectory Modeling and Next Occupation Prediction}
In the economics literature, when studies of worker transitions analyze the relationship between worker characteristics and career histories to career transitions, they have traditionally relied on fairly simple predictive models and considered only a few occupation categories.
For example, \citet{boskin_conditional_1974} uses a conditional logistic regression model to analyze the factors affecting workers' transitions among 11 occupational groups, where the factors included estimated earnings, training expenses, and costs due to unemployment.
\citet{schmidt_prediction_1975} uses a multinomial logistic regression to analyze the impact of race, sex, educational attainment, and labor market experience on the probability that individuals transition into one of five different occupational categories, revealing significant effects of these variables on occupational outcomes.
\citet{hall_turnover_1972} examines the dynamics of labor force turnover in the U.S., analyzing the influences of demographic factors, labor demand fluctuations, and job stability on unemployment. To study turnover, the authors consider factors such as race, counts and ages of children, estimated wage, income, age, marital status, location, and employment category (including private wage or salary, government roles, self-employment, and unpaid family work).
\citet{blau_labor_1999} model labor force transitions among older married couples, showing that one spouse's employment status significantly impacts the employment status of the other, with financial incentives and preferences for shared leisure influencing these transitions.
In addition to demographic characteristics, the authors incorporate human capital and education variables, including tenure on the current job and retirement benefits.

\paragraph*{Machine Learning Methods for Next Occupation Prediction}
Prior to the use of foundation models, other authors (e.g., \citet{li_nemo_2017,meng_hierarchical_2019,zhang_attentive_2021}) made use of various neural network architectures for the next-occupation prediction problem, sometimes training on large datasets. For example, \citet{li_nemo_2017} uses a Long Short-Term Memory (LSTM) neural network to predict occupation transitions, where the embedding dimension is 200, and the training set incorporates more than a million individuals. \citet{he2021your} build a model to predict the next occupation position out of 32 frequent position names, as well as job salary and firm size for that position, using a dataset of 70{,}000 resumes. Another approach taken by \citet{zhang2019job2vec} seeks to predict aggregate transition probabilities between pairs of job titles within the same firm. Their approach, which generates embeddings for each job title, does not attempt to condition on individual worker history. These papers do not make use of foundation models; a recent line of work builds domain-specific foundation models for careers, most notably CAREER developed by \citet{vafa_career_2024}, which we discuss in more detail below.

\paragraph*{Foundation Models for Careers}
Building on these machine-learning approaches, a growing literature applies foundation models to domains beyond text (\citet{savcisens2024using,wu2021prototransformer,radford2021learning}). Within labor economics, the most closely related paper to ours is \citet{vafa_career_2024}, who develop CAREER, a custom transformer-based foundation model whose vocabulary is restricted to occupation codes and that incorporates a dedicated state for remaining in the same occupation. The model is pre-trained on 24 million resumes of U.S. workers acquired from Zippia, Inc., a large but non-representative dataset, and is then fine-tuned using data from U.S. government surveys, including the Panel Study of Income Dynamics (PSID) (\citet{PSID2024}) and two cohorts from the National Longitudinal Survey of Youth (NLSY79 and NLSY97) (\citet{BLS_NLSY79_2023,BLS_NLSY97_2024}). \citet{vafa_career_2024} shows that CAREER substantially outperforms existing benchmarks for next-occupation prediction and that the learned embedding of career history transfers to related tasks: when the model is fine-tuned to predict wages, which are not available in the pre-training resume dataset, it improves wage prediction relative to popular wage regression models in labor economics. CAREER has fewer parameters (5.6 million) than FT-LABOR-LLM (7 or 13 billion), and its pre-training dataset, while highly relevant, is much smaller than the corpus used for Llama-2; moreover, CAREER does not make use of textual job titles or descriptions.

\paragraph*{Adapting LLMs to Build Domain-Specific Models}

Adapting pre-trained models to specific domains via fine-tuning has become a prevalent approach for improving the performance of LLMs for specific tasks.
The (full parameter) fine-tuning approach involves further updating all weights of a pre-trained model using domain-specific data and optimization techniques such as gradient descent (\citet{wei_finetuned_2022}).
The pre-training and fine-tuning paradigm has produced state-of-the-art models for dialogue systems (\citet{yi_survey_2024}), code generation (\citet{chen_evaluating_2021}), music generation (\citet{agostinelli_musiclm_2023}), scientific knowledge (\citet{taylor_galactica_2022}), protein structure prediction (\citet{rives_biological_2021}), chemistry (\citet{zhang_chemllm_2024}), medicine (\citet{singhal_large_2022}), and other settings. The literature on the adaptation of LLMs for recommendation systems is also closely related. \citet{geng_recommendation_2022} introduces a general paradigm to adapt the recommendation task to language processing.

Our paper compares our fine-tuning approach to one where LLM embeddings are extracted and treated as covariates in a multinomial logistic regression. This type of approach has been popular in language analysis for a long time; for example, it is used by sentiment classifiers (\citet{reimers-gurevych-2019-sentence}).

Finally, prompt engineering and in-context learning are alternative approaches to fine-tuning LLMs that require minimal computation and avoid the need for direct model access (\citet{brown2020languagemodelsfewshotlearners}). Prompt engineering involves designing specific queries, instructions, or examples within the prompt to direct the model's response. By tuning the language and structure of prompts, researchers can shape the model's output for different applications (\citet{maharjan2024openmedlm}). Researchers can also use in-context learning by providing relevant example data within the prompt itself, priming the model to continue the pattern and apply similar logic to new inputs (\citet{yin2024deeper,bao2023tallrec}). In this paper, we consider an approach in which we prompt off-the-shelf LLMs for a prediction of the next occupation using a textual representation of the worker's career history as the prompt. We show that including example resumes in the prompt helps improve the performance of off-the-shelf pre-trained LLMs, although performance is still worse than FT-LABOR-LLM.

\paragraph*{Other Applications of LLMs to Sequential Prediction Problems in Economics}
LLMs have also been used to model time series data (\citet{jin_time-llm_2024}) and in forecasting.
For instance, \citet{faria2024artificial} investigate the ability of LLMs to produce in-sample conditional inflation forecasts during the 2019–2023 period.

\paragraph*{The Biases and Representativeness of LLMs}A recent literature has emerged that aims to assess whether the outputs of foundation models are representative of larger populations, for example, whether the answers to opinion survey questions are representative of the population (\citet{santurkar_whose_2023, argyle_out_2023}). One proposed approach is to query LLMs with survey responses from long-standing opinion surveys and see how aligned their responses are with the survey average. Our question differs in that we want to know whether the (fine-tuned) LLM can make predictions about occupation transitions that are representative of real-world transitions, conditional on history, which is a more complicated question to answer, as the population conditional probabilities are unknown due to the high-dimensional space of potential histories.

\section{Data}

\subsection{Representative Survey Datasets.}
\label{sec:survey-datasets}
Our primary sources of data are three surveys of workers in the U.S. population. These surveys follow samples of individual workers, where workers are interviewed at regular intervals. The survey samples are constructed to be representative of the U.S. population at particular points in time.

The first dataset we consider is the Panel Study of Income Dynamics (PSID), which began in 1968. The sample of this dataset is intended to be representative of the United States as a whole, and new participants are added to the sample over time. Occupation information is consistently available starting in 1981, so we restrict our attention to survey years starting then.
We further analyze data from two waves of the National Longitudinal Survey of Youth (NLSY). The NLSY 1979 follows a cohort of people aged 14-22 in 1979 throughout their careers. The NLSY 1997 began in 1997 and followed a cohort of individuals aged 12-16 at that time throughout their careers.
We use extracts from these surveys to build longitudinal datasets for individual workers. Details of our dataset construction are reported in Appendix \ref{sec:appendix-data}.

We encode occupations using the \texttt{occ1990dd} system (\citet{occ1990dd}) to map different versions of Census OCC occupational codes to a harmonized set of codes. In addition to the 332 occupations from the \texttt{occ1990dd} taxonomy, we include three special categories: ``education,'' ``out of labor force,'' and ``unemployed.''
We extract demographic characteristics, specifically gender, ethnicity, region of the country, and birth year. To simplify our analysis, we assign each worker a single, unchanging value for each demographic characteristic, typically the first valid value, and we do not allow it to change even if the original survey specifies different values in different survey years. We do not impute occupations for years without survey responses and focus on a single main occupation reported by the subject.

We refer to the cleaned versions of the three survey datasets as PSID81, NLSY79, and NLSY97. Table \ref{tab:data-summary} summarizes the total number of workers and transitions (individual-year survey observations, denoted by $\sum_i T_i$) in each survey dataset. The PSID81 dataset has 10.1 transitions per individual on average (median is 8 and maximum is 29), and the NLSY79 and NLSY97 track relatively fewer workers but have more transitions per individual, with 20.82 (median is 25 and maximum is 29) and 16.56 (median is 19 and maximum is 20), respectively, observations per worker on average.

\begin{table}[t]
    \centering
    \caption{Description of datasets.\updated{10/30/2025}}
    \label{tab:data-summary}
    \begin{tabular}{lccc}
    \toprule
 &    PSID81 & NLSY79 & NLSY97 \\
   \midrule
 Number of Individuals    & 31,056 & 12,479 & 8,984 \\
 Tokens in  $\cup_{i}\texttemplate(x_{i, \leq T_i}, y_{i, \leq T_i})$ & 7,902,511 & 5,406,412 & 3,135,367 \\
 Number of Transitions $\sum_i T_i$   & 313,622 & 259,778 & 148,795 \\
 Tokens in $\cup_{i}\cup_{1 \leq t \leq T_i}\texttemplate(x_{i, \leq t}, y_{i, < t})$ & 69,139,450 & 72,639,496 & 32,368,253 \\
 \hline
 ``First observation'' transitions $t = 1$ & 9.9\% & 4.8\% & 6.0\% \\
 ``Moving'' transitions with $y_{i, t-1} \neq y_{i, t}$ & 38.5\% & 44.5\% & 37.0\% \\
 ``Staying'' transitions with $y_{i, t-1} = y_{i, t}$  & 51.6\% & 50.6\% & 57.0\% \\
    \bottomrule
    \end{tabular}
    \legend{The top panel reports counts of individuals, transitions, and tokens. Token counts are reported separately for text representations used in fine-tuning $\texttemplate(x_{i, \leq T_i}, y_{i, \leq T_i})$ and text representations used for inference $\texttemplate(x_{i, \leq t}, y_{i, < t})$. The bottom panel reports the proportion of transitions corresponding to three transition types: first observation, moving, and staying.}
\end{table}

As previously discussed, we convert individuals' complete career trajectories into natural language paragraphs using a text template. The total number of tokens ranges from 3.1 million to 7.9 million.
The average length of career history $\texttemplate(x_{i, \leq t}, y_{i, < t})$ ranges from 210 to 280 tokens depending on the dataset, while the average length of complete career history $\texttemplate(x_{i, \leq T_i}, y_{i, \leq T_i})$ ranges from 250 tokens to 430 tokens; all of these templates fit well within Llama-2 model's context window of 4,096 tokens.

Transitions between two observations can be categorized into three types: \emph{first observation}, \emph{moving} (i.e., the current occupation is different from the occupation reported in the previous observation), and \emph{staying} (i.e., the current occupation is the same as the occupation reported in the previous observation). Table \ref{tab:data-summary} shows the number of transitions by transition type, highlighting that between 50\% and 60\% of transitions involve staying in the same occupation.

We divide each of the three survey datasets into training (70\%), validation (10\%), and test (20\%) samples, where all of an individual's transitions are allocated to the same set. Results about the performance of models are presented for the test set. Online Appendix \ref{sec:online-appendix-summary-datasets} and Table \ref{tab:demographic-distributions} provide details and summary statistics.

\begin{figure}[!htbp]
    \centering
    \includegraphics[width=0.8\linewidth]{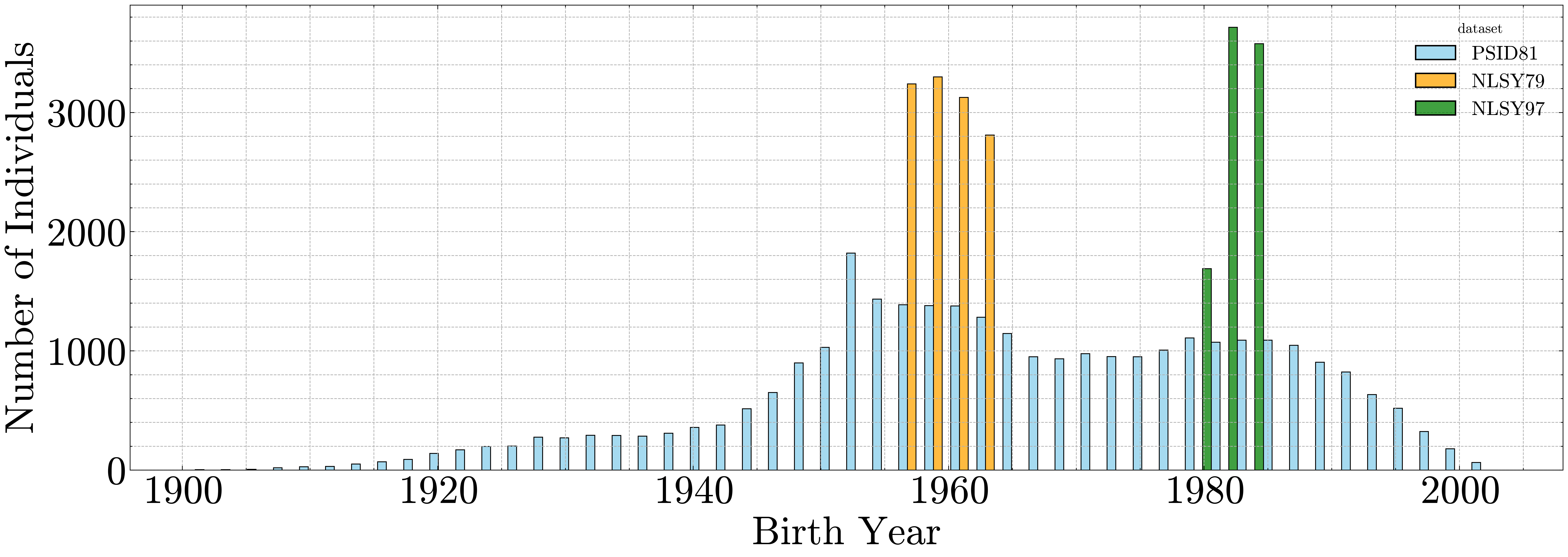}
    \caption{Distributions of individuals' birth years by survey dataset.\updated{12/04/2025}}
    \label{fig:dataset-summary-birth-year}
\end{figure}

Birth cohort affects workers' career trajectories (\citet{wachter2020persistent,doi:10.1177/0003122420966324}). Figure \ref{fig:dataset-summary-birth-year} shows that while birth years of PSID81 individuals span the range covered by NLSY79 and NLSY97, birth years of NLSY individuals are clustered within a small range due to the design of the NLSY surveys.

Table \ref{tab:top-10-occupations-by-dataset} presents the top ten occupations in each dataset, highlighting commonalities and variations across datasets. Notable trends include ``Not in labor force'' ranking highest in all datasets, while occupations like ``In education'' show substantial variation, ranking 9$^{th}$ in PSID but 2$^{nd}$ and 1$^{st}$ in NLSY79 and NLSY97, respectively.

\begin{table}[t]
    \centering
    \caption{Top occupations by dataset.}
    \label{tab:top-10-occupations-by-dataset}
    \resizebox{\textwidth}{!}{
    \begin{tabular}{lcccccc}
    \toprule
     & \multicolumn{2}{c}{PSID81} & \multicolumn{2}{c}{NLSY79} & \multicolumn{2}{c}{NLSY97} \\
     \textbf{Occupation} & Proportion & Rank & Proportion & Rank & Proportion & Rank \\
    \midrule
    Not in labor force & 0.192 & 1 & 0.177 & 1 & 0.122 & 2 \\
    Unemployed & 0.067 & 2 & 0.040 & 4 & 0.034 & 3 \\
    Postmasters and mail superintendents & 0.058 & 3 & 0.045 & 3 & 0.019 & 5 \\
    Coin, vending, and amusement machine servicers and repairers & 0.025 & 4 & 0.025 & 5 & 0.013 & 9 \\
    Secretaries and administrative assistants & 0.022 & 5 & 0.020 & 6 & 0.008 & 16 \\
    Phlebotomists & 0.021 & 6 & 0.017 & 8 & 0.020 & 4 \\
    Telemarketers & 0.016 & 7 & 0.005 & 41 & 0.016 & 7 \\
    Maids and housekeeping cleaners & 0.014 & 8 & 0.011 & 13 & 0.004 & 31 \\
    In education & 0.013 & 9 & 0.136 & 2 & 0.343 & 1 \\
    Elementary and middle school teachers & 0.013 & 10 & 0.008 & 25 & 0.007 & 19 \\
    Painting workers & 0.013 & 11 & 0.015 & 10 & 0.005 & 28 \\
    Sales Representatives Services All Other & 0.011 & 13 & 0.017 & 9 & 0.006 & 25 \\
    Septic tank servicers and sewer pipe cleaners & 0.010 & 15 & 0.018 & 7 & 0.012 & 10 \\
    Cashiers & 0.009 & 21 & 0.011 & 11 & 0.018 & 6 \\
    First-line supervisors/managers of retail sales workers & 0.008 & 28 & 0.005 & 40 & 0.014 & 8 \\
    \bottomrule
    \end{tabular}
    }
    \legend{We take the union of the top ten occupations from each dataset separately (15 occupations in total) and report the proportion of transitions to each occupation in each dataset, as well as the rank of the proportion compared to other occupations in the same dataset. Readers can refer to Appendix Figure \ref{fig:word-cloud-job-titles} for a word cloud of job titles.
\updated{10/30/2025}}
\end{table}

Figure \ref{fig:years-of-experience-by-calendar-year-and-dataset} illustrates the distribution of individual ages across calendar years for each dataset. In the NLSY datasets, the age distribution increases steadily over time, reflecting the longitudinal design that follows the same cohort of individuals. In contrast, the PSID dataset allows for dynamic changes in its subject pool, with individuals entering (e.g., upon becoming the head of a household) and exiting the study. Consequently, the PSID81 dataset exhibits a broader but more temporally stable age distribution. This figure highlights the degree of overlap in age distributions across the three datasets, suggesting that the occupation transition probabilities may be systematically different across the datasets.

\begin{figure}[!htbp]
    \centering
    \includegraphics[width=\textwidth]{figures_and_tables_revision/figure_2_distribution_of_individuals_ages_by_calendar_year_of_observation.png}
    \caption{Distribution of individuals' ages by calendar year of observation. \updated{10/30/2025}}
    \label{fig:years-of-experience-by-calendar-year-and-dataset}
\end{figure}

\subsection{Large-Scale Resume Data}

In this paper, we re-implement the full pre-training and fine-tuning pipeline of the CAREER model so that we can carry out the fine-tuning step on identical survey datasets. Pre-training CAREER involves using a proprietary resume dataset of 24 million resumes acquired from Zippia Inc.\footnote{Zippia is a data-driven career intelligence platform that leverages analytics to provide personalized job recommendations, salary insights, and career development resources. The company aggregates labor market data to offer tailored guidance for job seekers, aiming to optimize their career decisions and employability. Other vendors providing similar data include Kaggle \url{https://www.kaggle.com/datasets/snehaanbhawal/resume-dataset} and Revelio \url{https://www.data-dictionary.reveliolabs.com/index.html}.} As described in Online Appendix \ref{sec:online-appendix-career}, we follow the approach of \citet{vafa_career_2024} to prepare and clean the data from Zippia Inc. This pre-training resume data represents resumes from the Zippia data as annual sequences of \texttt{occ1990dd} occupations, with tie-breaking rules for multiple jobs per year. Covariates include the year of each job, the latest educational degree, and the location, standardized following the approach we use for cleaning the survey datasets. Missing covariates are replaced by a special token, and missing occupational years are dropped. The final dataset comprises 245 million transitions (that is, individual-year observations).

\section{Occupation Models}\label{sec:model}

\subsection{Notation for Occupation Models.}
We refer to a model that predicts an individual's next occupation as a function of career history and other individual characteristics as an \textbf{occupation model}. Let $t\in \{1,..,T_i\}$ correspond to a year in which an individual $i$ was surveyed and otherwise met our filtering requirements, where $T_i$ denotes the total number of individual-year observations of this individual. Note that our cleaned survey datasets do not, in general, have observations in every calendar year. We refer to an observation of an individual's occupation as a ``transition,'' with some abuse of terminology since we use the term for the first observation and also even when the individual stays in the same occupation. Let $\text{year}_{i, t}$ denote the calendar year corresponding to transition $t$ for individual $i$. We represent occupations as discrete variables, in particular following the \texttt{occ1990dd}, a variant of the OCC occupational classification system of \citet{occ1990dd}, as described in Appendix \ref{sec:appendix-data}. Let $\mathcal Y$ denote the set of all occupations, and let $y_{i,t} \in \mathcal{Y}$ represent the occupation that an individual $i$ has at transition index $t$.
We let $y_{i, < t} =(y_{i,1}, \dots, y_{i, t-1})$ denote an individual's occupation sequence prior to their $t$'th observation (for $t = 1$, define $y_{i, < t} = \emptyset$). Let $\mathcal{X}_\text{inv}$ be the support of time-invariant covariates (in our application, gender, race/ethnicity, region, and birth year, denoted by $x_i$), while
$\mathcal{X}_\text{var}$ is the support of time-varying covariates (in our application, education and calendar year, denoted by $x_{i, t}$). Let $x_{i, \leq  t} = (x_i, x_{i,1}, \dots, x_{i, t}) \in \mathcal{X}_\text{inv} \times \mathcal{X}_\text{var}^t$ denote the time-invariant covariates and time-varying covariates up to and including $t$. We refer to $(x_{i, \leq  t}, y_{i, < t},)$ as the worker's career history at transition $t$.

The probability that the worker's next occupation is $y_{i,t}$, conditional on the worker's history, is written $P(y_{i,t} \mid x_{i, \leq t}, y_{i, < t})$.

\subsection{Assessing Predictive Performance of Occupation Models}

We evaluate an occupation model's performance by comparing its predictions of an individual's next occupation to their actual next occupation. Specifically, we evaluate models by computing their perplexity, a commonly used metric in Natural Language Processing (NLP).\footnote{We primarily evaluate models using perplexity rather than more traditional metrics such as accuracy or F1 because occupational transitions are inherently imbalanced: around 50\% of workers stay in the same occupation (see Table \ref{tab:data-summary}). Well-calibrated models will therefore almost always predict staying (i.e., $\text{argmax}_{y \in \mathcal{Y}} \hat{P}(y \mid x_{i, \leq t}, y_{i, < t}) = y_{i,t-1}$), making accuracy simply reflect the empirical staying rate rather than meaningful model differences. Perplexity instead uses the actual predicted probability $\hat{P}(y_{i,t} \mid x_{i, \leq t}, y_{i, < t})$, rewarding models that assign higher probabilities to correct occupations regardless of whether they involve staying or moving.}
The perplexity is a negative monotonic transformation of the sample log-likelihood, with lower perplexity indicating that a model's predictions are more accurate. Formally, the perplexity of an occupation model $\hat{P}$ on a set of transitions (individual-year observations) for units $i=1,..,I$ is given by
\begin{align*}
    \text{perplexity} =\exp \left\{-\frac{1}{\sum_{i} T_i} \sum_{i=1}^I \sum_{t=1}^{T_i} w_{it} \left[ \log \hat{P}(y_{i,t} \mid x_{i, \leq t}, y_{i, < t}) \right] \right\},
\end{align*}
where $w_{it}$ denotes the sampling weight for the individual relative to a population of interest. In this paper, for simplicity, we set $w_{it} = 1$.
Note that a completely uninformative model that assigns uniform mass to each possible occupation would achieve a perplexity of $|\mathcal{Y}|$.
We consider additional evaluation metrics (such as calibration) in Section \ref{sec:comparing-performance}. We also consider predictive performance for several alternative outcomes of interest to economists, such as future unemployment status and coarse occupational categories; see Online Appendix \ref{sec:online-appendix-alt-eval-tasks}.

\subsection{Quantifying Uncertainty in Performance Metrics}
Model performance uncertainty comes from two sources: (i) randomness in model training (both the particular training sample and the stochastic elements of SGD), and (ii) randomness in the sample draw. We quantify (i) with fine-tuning multiple models on bootstrapped training sets and (ii) by evaluating the same model on bootstrapped test sets. Because full fine-tuning is computationally costly, we report test-set bootstrap standard errors for all models and provide training-set bootstrap standard errors only for the three focal models discussed in Section \ref{sec:comparing-performance}. See Appendix \ref{sec:appendix-bootstrap} for more details on both the training- and test-set-bootstrap standard errors.

\section{Large Language Models as Foundation Models}
\label{sec:llms}
\subsection{LLM Notation}
 The LLMs we use in this paper are trained to perform next-word prediction.  Let $\mathcal{W}$ be the allowable set of words and punctuation, while $\cup_{j=1}^{\infty} \mathcal{W}^j$ is the set of sequences of words.  In practice, LLMs work in the space of ``tokens,'' where words can be transformed into a sequence of tokens using a process called ``tokenization.'' We let $\mathcal{V}_\text{LLM}$ be the set of all possible tokens (also known as the vocabulary set) for a particular LLM. Popular commercial-scale LLMs typically use vocabulary sets with 10,000 to 200,000 tokens; for example, $|\mathcal{V}_\text{Llama-2}| = 32,000$.
Let \tokenize$:\cup_{j=1}^\infty \mathcal{W}^j \rightarrow \cup_{j=1}^\infty \mathcal{V}^j$ denote the function mapping a sequence of words to a sequence of tokens.

The LLMs we consider can be viewed as estimating the probability that the next token equals $v_{j+1}$, conditional on a sequence of $j$ tokens (i.e., the prompt). LLMs also impose restrictions on the ``context length,'' or the maximum length of a sequence that can be conditioned on, which we denote $C$, with particular values $C_\text{LLM}$ imposed by different LLMs. Then, we let $\hat{P}^\mathcal{V}_\text{LLM}: \mathcal{V}_\text{LLM} \times \cup_{j \leq C_\text{LLM}} \mathcal{V}_\text{LLM}^j \to [0, 1]$ denote the LLM's estimate of the probability of the next token conditional on the input sequence, which for particular values of $v_1, \dots, v_{j+1}$ is written $\hat{P}^\mathcal{V}_\text{LLM}(v_{j+1} \mid v_1, \dots, v_j)$.  The conditional probability $\hat{P}^\mathcal{V}_\text{LLM}(v_{j+1}, \dots, v_{j+j'} \mid v_1,\dots, v_j)$ for $j + j' \leq C_\text{LLM}+1$ can be derived from individual next-token predictions using the chain rule.

\subsection{Text Templates}
\label{sec:text-template}

In this section, we describe the \textbf{text template} function we use to convert a worker's history of occupations and covariates into a sequence of words and punctuation. This text template can be combined with a tokenizer and an LLM to make next occupation predictions, as described in the next section.

We let $\jobtitle: \mathcal{Y} \to \cup_{j=1}^\infty \mathcal{W}^j$ denote the mapping from an occupation to its English-language title. Note that this mapping should be bijective (one-to-one).
For example, the title of the occupation with \texttt{occ1990dd} code 95 is ``nurse practitioners.''\footnote{Even though the \texttt{occ1990dd} system does not include job titles directly, one can crosswalk it to, for example, the Standard Occupational Classification (SOC) system, and use the list of job titles attached to each SOC code provided by the Bureau of Labor Statistics (\href{https://www.bls.gov/OES/CURRENT/oes_stru.htm}{https://www.bls.gov/OES/CURRENT/oes\_stru.htm}).}
The number of tokens needed to represent a job title depends on the tokenizer; using the Llama-2 tokenizer, the number ranges from 2 to 28 in the survey datasets we analyze, with an average length of 8.3 tokens (and a standard deviation of 4.8).\footnote{``Grinding, Lapping, Polishing, and Buffing Machine Tool Setters, Operators, and Tenders, Metal and Plastic'' (28 tokens) and ``Cutting, punching, and press machine setters, operators, and tenders, metal and plastic'' (24 tokens) are the two longest job titles.
The shortest job titles include ``Cooks'', ``Bakers'', ``Tellers'', and ``Designers''.}

To represent covariates as text, we express an individual's educational status using values such as \texttt{graduate degree}. Online Appendix \ref{sec:online-appendix-occupation-titles} provides the full mapping between \texttt{occ1990dd} codes and their job titles.

Building on this strategy, we define a \textbf{text template} function, $\texttemplate(x_{i, \leq t}, y_{i, < t})$, that transforms an individual's career history into a textual summary.
The text template incorporates additional punctuation, line breaks, and metadata, as detailed in Appendix \ref{sec:appendix-text-template}, and illustrated in the following example.

{
\scriptsize\begin{Verbatim}[breaklines=true,commandchars=+\[\]]
<A worker from the PSID dataset>
The following information is available about the work history of a female black or african american US worker residing in the south region.
The worker was born in 1963.
The worker has the following records of work experience, one entry per line, including year, education level, and the job title:
1984 (some college): Cooks
1985 (some college): Cooks
1987 (some college): Food servers, nonrestaurant
1989 (some college): +underline[+texttt[Cleaners of vehicles and equipment]]
+underline[+texttt[<END OF DATA>]]
\end{Verbatim}
}

The example above is defined as the text representation of the \textbf{complete career history} of the individual, denoted $\texttemplate(x_{i, \leq t}, y_{i, \leq T_i})$, where $T_i$ represents the number of transitions for individual $i$. These complete career histories are used for model fine-tuning, as discussed in Section \ref{sec:fine-tuning-clm}.

Note that the individual can stay in the same occupation for multiple records (e.g., 1984 and 1985 in the example); the text representation explicitly reflects this information.
Additionally, the dataset could miss an individual for certain years in her career trajectory; in this case, the text template will skip those years (e.g., 1987 and 1995 in the example).

We also create the text representation of the \textbf{career history} of the same individual prior to the $t^\text{th}$ occupation, denoted $\texttemplate(x_{i, \leq t}, y_{i, < t})$, by truncating the complete career history.
For example, to obtain an LLM's predictions of an individual's occupation in 1989 given the covariates and occupation history, we would use as input the text above, removing the text ``Cleaners of vehicles and equipment" and everything afterward (i.e., the underlined part in the example). That is, we apply the text template function to all previous occupation and covariate information, and conclude with a partial row for the occupation to be predicted.

On average in the survey datasets we consider, the text representation of workers' complete career histories contains around 250 to 500 tokens using the Llama-2 tokenizer, which fits within the context window of Llama-2 models for fine-tuning.
For inference, the prompt encoding of an individual's career history, i.e., $\tokenize(\texttemplate(x_{i, \leq t}, y_{i, < t}))$, consists of 200 to 300 tokens on average. Detailed summary statistics on token counts can be found in Online Appendix \ref{sec:online-appendix-summary-tokenized-datasets}.

\subsection{Using LLMs for Occupation Modeling}
\label{sec:UsingLLMs}
In this paper, we use LLMs in three ways. First, we use an LLM to directly produce a ``predicted occupation'' in response to a ``prompt.'' More precisely, if we first map job codes to text (the English language job title) using the text template function described in the previous subsection, and then use a tokenizer to translate the resulting sequences of past jobs into a sequence of tokens, an LLM will produce a textual ``response'' that is a sequence of tokens. That sequence may or may not correspond to a valid occupation, but we can, in principle, further transform the output in various ways to interpret it as an occupation. Of course, a textual response or a single predicted occupation is not an estimate of the probability of a sequence of tokens.  Some commercial LLMs allow the user to set a ``temperature'' parameter when submitting a prompt, where a particular setting is designed to approximate sampling from the distribution of responses. Probabilities can then be estimated by repeatedly prompting the LLM. We do not follow this approach in this paper; instead, we restrict attention to LLMs where probabilities (or, where relevant, embeddings) can be directly obtained by the analyst.

Second, for those LLMs for which it is possible, we directly obtain the probability assigned to a given token. This functionality may be enabled in the setup of an open model such as Llama-2, or it may be exposed through an API in the case of a closed model such as ChatGPT-4.\footnote{For example, \url{https://cookbook.openai.com/examples/using_logprobs} explains how to use the logprobs parameter in OpenAI API requests to evaluate token probabilities, allowing analysis of model confidence and alternative predictions for improved understanding of text generation.} For example, for the LLM Llama-2 7 billion parameter model, denoted Llama-2-7B, we let $\hat{P}^\mathcal{V}_\text{Llama-2-7B}(\text{``Engineer''} \mid \text{``Software''})$ denote the estimated probability that ``Engineer'' follows the single-token sequence ``Software''.
To obtain the probability of the next occupation given a sequence of prior occupations, we first use the text template function and the tokenizer to translate the occupation history into a sequence of tokens; similarly, we translate the title of a particular next occupation $y_{i,t+1}$ into a sequence of tokens.
The estimated next-token probability model associated with the LLM, denoted by $\hat{P}^\mathcal{V}_\text{LLM}(\cdot \mid v_1, \dots, v_j): \mathcal{V}_\text{LLM} \to [0, 1]$, can be applied several times to determine (using the chain rule of probability) an estimate of the probability that the sequence of tokens induced by $y_{i,t+1}$ follows the sequence of tokens induced by $(y_{i,1},\dots,y_{i,t})$.
A language-based next-token prediction model thus induces an associated occupation model, as follows. See Appendix \ref{sec:appendix-predict-as-tokens} for details.
\begin{align}
    \begin{aligned}
    &\hat{P}_\text{LLM}(y_{i,t} \mid x_{i, \leq t}, y_{i, < t})
    \overset{\text{def}}{=} \hat{P}^\mathcal{V}_\text{LLM}(\tokenize(\jobtitle(y_{i,t})) \mid \tokenize(\texttemplate(x_{i, \leq t}, y_{i, < t}))).
    \end{aligned}
    \label{eq:LLMProbModel}
\end{align}

Third, some LLMs make it possible to extract a lower-dimensional ``embedding'' of text, where any sequence of tokens is associated with a real-valued vector. For example, for the Llama-2-7B LLM that we use in this paper, input text is represented as a vector of 4,096 floating-point numbers. Formally, we let $\mathcal{E}_\text{LLM}: \cup_{j \leq C_\text{LLM}} \mathcal{V}_\text{LLM}^j \to \mathbb{R}^{d_\text{LLM}}$ be the ``embedding function'', where $d_\text{LLM}$ denotes embedding dimension. The composite function $\mathcal{E}_\text{LLM} \circ \tokenize$ generates the embedding of any input string of words (i.e., the ``prompt'').%\footnote{The exact implementation of the embedding function differs for encoder-decoder transformer architectures (e.g., the original transformer by  \textbf{Vaswani et al., 2017}) and decoder-only transformers (e.g., CAREER and Llama). In encoder-decoder transformers, obtaining an embedding for the input token sequence is straightforward, as the encoder explicitly computes a sequence-level representation. In contrast, for decoder-only transformers, the standard practice is to use the embedding associated with the last token of the input sequence. This embedding captures the most comprehensive representation of the input context and serves as the basis used by the language model's prediction head for predicting the next token. This is the procedure followed by this paper.}

\subsection{Estimation of Transformer-Based Foundation Models and Fine-Tuning}
\label{sec:TransEst}

Both CAREER and our preferred model, FT-LABOR-LLM, make use of the transformer neural network functional form for the occupation transition model. In addition, both make use of the foundation model and fine-tuning approach popularized in the artificial intelligence literature. FT-LABOR-LLM builds on the Llama-2 series of LLMs. Here, we provide an overview of the general approach used to estimate and fine-tune Llama.

A transformer model can be viewed as a multinomial logistic regression model where the outcome $v_j$ is selected from the set of tokens $\mathcal{V}$ and the covariates are the previous tokens $v_{1:j}$ in the sequence. To estimate the probabilities associated with each next token given history, the model first maps the history $v_{1:j}$ into an embedding using a parameterized embedding function (defined in Section \ref{sec:UsingLLMs})
$z_j = \mathcal{E}_\text{tsf}(v_{1:j};\gamma)\in \mathbb{R}^{d_\text{tsf}}$. The embedding function is specified as a transformer neural network where the parameters determine how strongly tokens in earlier and later positions should be weighted in forming the prediction of the next token, and these weights vary with the specific tokens in the history.
Given an embedding $z_j$, the probability that the next token equals $v \in \mathcal{V}$ can be written as
$$
\hat{P}^{\mathcal{V}}_{\text{Transformer}}(v_{j+1}=v \mid v_{1:j})
= \frac{\exp\!\big(z_j^\top \hat{\beta}_v\big)}
       {\sum_{v' \in \mathcal{V}} \exp\!\big(z_j^\top \hat{\beta}_{v'}\big)},
$$
where $\hat{\beta}_v \in \mathbb{R}^{d_\text{tsf}}$ is the parameter vector associated with the outcome that the next token is $v$.
The parameters in the embedding function $\hat\gamma$ and the coefficients $\{\hat{\beta}_v\}_{v \in \mathcal{V}}$ are estimated jointly using maximum likelihood. Each token is an outcome and thus corresponds to an observation. It is standard practice to estimate the parameters using variations of stochastic gradient descent (\citet{touvron_llama_2023}), where observations are batched into relatively small batches and for each batch, parameters are updated in the direction that increases the likelihood, where the direction is determined by calculating the gradient of the objective function evaluated at the current batch of observations.

Estimating the parameters of a foundation model, sometimes referred to as ``pre-training,'' refers to estimation using a large, potentially unrepresentative dataset. Fine-tuning the model refers to estimating the same parameters using a second, distinct dataset that better represents the eventual prediction tasks of interest, but where the starting values for parameter estimation are the ending values from the pre-training phase. The fine-tuning phase typically proceeds through multiple ``epochs,'' where an epoch is a complete pass through the fine-tuning data. Using more epochs is analogous to placing a higher weight on the fine-tuning data. Performance of the model on a held-out validation set (distinct from the test set) is used to select the number of epochs.

\subsubsection{Fine-Tuning FT-LABOR-LLM}
We use the term FT-LABOR-LLM to refer to the combination of a base model (either Llama-2-7B or Llama-2-13B) and fine-tuning data, as well as to refer to the union of the fine-tuned models we evaluate in this paper.
Consider the implementation of fine-tuning for FT-LABOR-LLM (see Appendix \ref{sec:appendix-fine-tune-details} for details). We fine-tune the LLMs to predict each token of a textual summary of worker careers, including punctuation and metadata, so that the FT-LABOR-LLM learns the structure of the text template as well as the conditional probabilities of tokens corresponding to occupations.
For each individual $i$, we use the text template discussed in Section \ref{sec:text-template} to build a text representation of their entire career, denoted as $\texttemplate(x_{i, \leq T_i}, y_{i, \leq T_i})$. In estimation, for each $i$, each token $(i,t)$ is a distinct outcome and corresponds to a distinct observation in training.

\begin{figure}[!htbp]
    \centering
    \includegraphics[width=\linewidth]{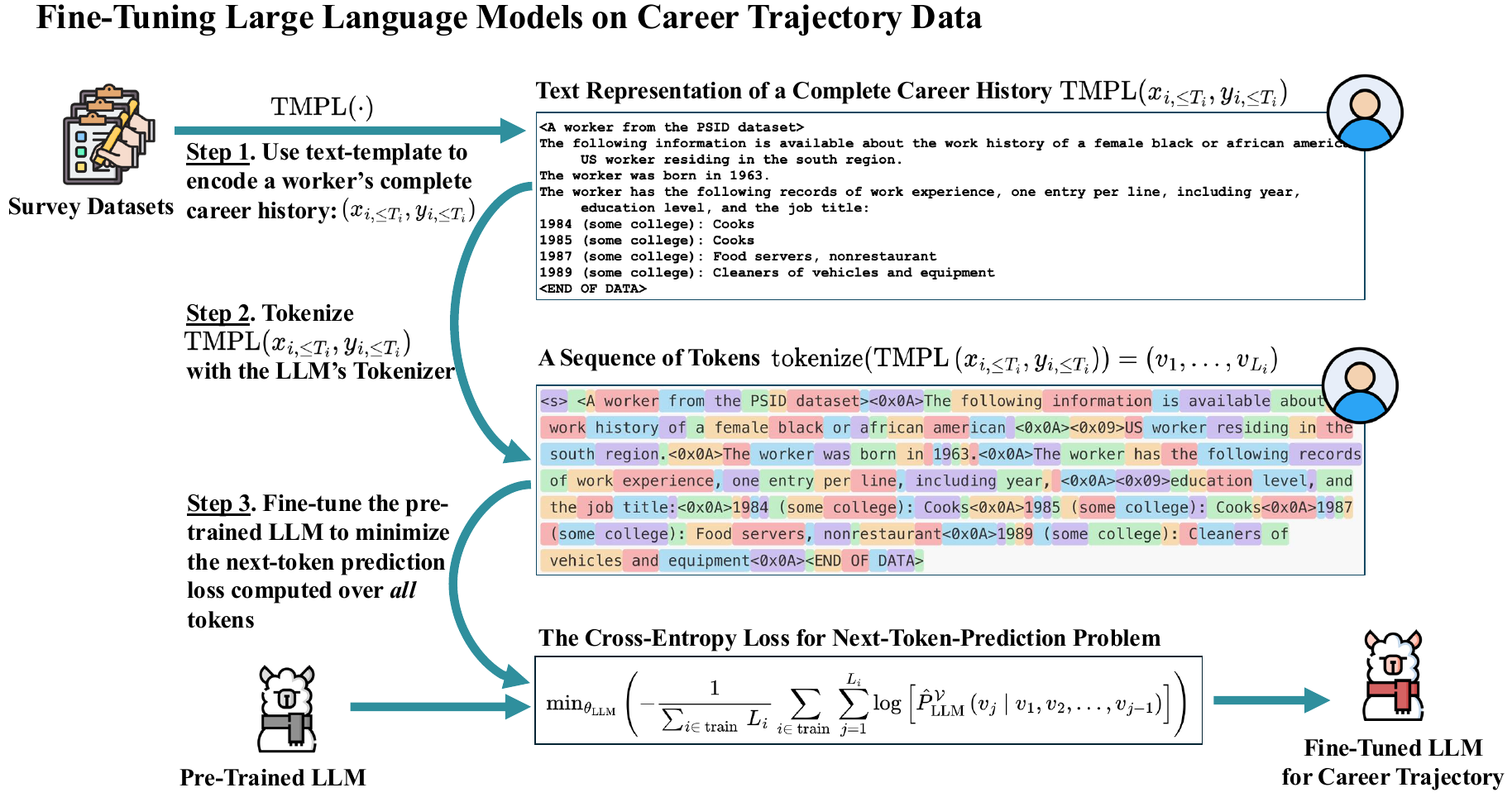}
    \caption{Illustration of the model fine-tuning procedure.\updated{10/30/2025}}
    \label{fig:fine-tune}
\end{figure}

We fine-tune the two Llama-2 models (7B and 13B) separately on each of the three training set text templates, resulting in three sets of fine-tuned models. We refer to the occupation models derived from these fine-tuned models through \ref{eq:LLMProbModel} as FT-7B and FT-13B, dropping the ``Llama-2'' nomenclature because we only fine-tune Llama-2 models. The fine-tuning procedure is illustrated in Figure \ref{fig:fine-tune}. The resulting FT-LABOR-LLMs are themselves LLMs, and we create estimates of $\hat{P}_\text{LLM}$ based on each of them.

In practice, for estimation, we simply upload the textual resume files to the LLM hosting service Together AI and make use of their fine-tuning service. When fine-tuning FT-LABOR-LLM we use either three or five epochs, whichever has better performance in the validation set. For making predictions, we apply the estimated transformer neural network to resumes in the test set.

\subsubsection{CAREER: Fine-tuning a Transformer-Based Foundation Model}
\label{sec:transformers}

In this section, we provide details on the CAREER model, which builds on the same transformer architecture described in Section \ref{sec:TransEst}.
The CAREER model specializes this architecture for occupational sequences in several ways.
First,
CAREER restricts the vocabulary to the set of occupations, $\mathcal{V}_\text{CAREER} = \mathcal{Y}$, and directly models sequences of occupations rather than sequences of words.
Second, at each step, CAREER embeds not only the previous occupation token but also covariates relevant for predicting the next occupation.
These include static attributes (e.g., gender, region, birth year) and time-varying attributes (e.g., calendar year, most recent education level).
Third, CAREER replaces the simple multinomial logistic over $\mathcal{Y}$ with a two-stage nested logistic structure: it first predicts whether the individual stays in the current occupation or moves; and conditional on a move, it outputs a distribution over new occupations in $\mathcal{V}_\text{CAREER} \setminus \{v_J\}$. This captures the empirical pattern that workers exhibit a high probability of remaining in the same occupation across periods.
Fourth, CAREER is trained in two phases.
It is first pre-trained on 24 million occupation sequences from resumes to learn general career patterns (i.e., the foundation model), and then fine-tuned on smaller, representative survey datasets to align the model with the target population.
Online Appendix \ref{sec:online-appendix-career} provides more details.
\section{Comparing Performance of Occupation Models}
\label{sec:comparing-performance}
In this section, we compare the performance of alternative occupation models.

\subsection{LLM Embeddings as Features in Multinomial Logistic Regression Models}
\label{sec:lm-as-embedding-engine}
This section exploits LLMs for occupational modeling by training a multinomial logistic regression model with $\mathcal{Y}$ as the set of alternatives and embeddings derived from off-the-shelf (that is, not fine-tuned) LLMs as covariates. Formally, let $z_{i, t}$ be an embedding vector (defined in Section \ref{sec:UsingLLMs}) of the career history $x_{i, \leq t}, y_{i, < t}$ derived from an LLM. For each occupation $y \in \mathcal{Y}$, the model estimates a parameter $\hat\beta_y$ such that
$
    \hat{P}_\text{MNL}(y_{i, t} \mid x_{i, \leq t}, y_{i, < t}) = \frac{\exp(z_{i, t}^\top \hat\beta_{y_{i, t}})}{\sum_{y' \in \mathcal{Y}} \exp(z_{i, t}^\top \hat\beta_{y'})}.
    \label{eq:MNL}
$
The parameters $\{\hat\beta_y\}_{y \in \mathcal{Y}}$ are estimated via maximum likelihood.

We first convert the career history $(x_{i, \leq t}, y_{i, < t})$ to natural language using the text template described in Section \ref{sec:text-template}. We then pass the text to an LLM and extract the model's embedding, $\mathcal{E}_\text{LLM}(\texttemplate(x_{i, \leq t}, y_{i, < t})) \in \mathbb{R}^{d_\text{LLM}}$.
This approach requires that the researcher has access to the embeddings from the LLM either through an API or by using an open-weight model. We consider a wide range of off-the-shelf models to embed career histories into embedding vectors, including Llama-2-7B/13B, Llama-3.1-8B, Llama-3.2-1B/3B, as well as the \texttt{text-embedding-3-large} text embedding model provided by OpenAI.\footnote{This model was released January 2024 and was the latest available as of December, 2025.} We then train a multinomial logistic regression on top of these embeddings for the next occupation prediction task.
See Appendix \ref{sec:appendix-embedding-based-approach} for details.

Table \ref{tab:perplexity-embedding-approach} compares performance across baseline models. CAREER outperforms all of the embedding-based multinomial logistic regression models.\footnote{For CAREER, predictions were made directly; we do \textbf{not} use CAREER as an embedding engine and build multinomial logistic regression on top of the embeddings.}

\begin{table}[!htbp]
    \centering
    \caption{\updated{10/30/2025} Test set perplexity for embedding-based approaches vs. CAREER.}
    \label{tab:perplexity-embedding-approach}
    \resizebox{\textwidth}{!}{
        \begin{tabular}{lcccc}
        \toprule
         & \multicolumn{1}{r}{\textbf{Dataset}} & PSID81 & NLSY79 & NLSY97 \\
         & \multicolumn{1}{r}{\textbf{Number of Test Set Transitions}}  & 61,759 & 51,593 & 29,949 \\
        \midrule
        \textbf{Model} & \textbf{Embedding Dimension $d_\text{LLM}$} & & & \\
        \midrule
        Empirical Transition Frequency & -- & 14.636 (0.224) & 14.264 (0.271) & 10.047 (0.169) \\
        OpenAI Text Embedding & 3,072 & 11.176 (0.191) & 12.065 (0.245) & 9.278 (0.189) \\
        OTS Llama-2-7B & 4,096 & 10.184 (0.169) & 10.764 (0.216) & 8.216 (0.164) \\
        OTS Llama-2-13B & 5,120 & 10.167 (0.169) & 10.704 (0.203) & 7.995 (0.152) \\
        OTS Llama-3.1-8B & 4,096 & 9.916 (0.162) & 10.517 (0.203) & 7.890 (0.151) \\
        OTS Llama-3.2-1B & 2,048 & 9.919 (0.164) & 10.378 (0.200) & 7.875 (0.146) \\
        OTS Llama-3.2-3B & 3,072 & 9.789 (0.156) & 10.283 (0.199) & 7.663 (0.141) \\
        \midrule
        CAREER (without pre-training) & -- & 8.996 (0.144) & 9.143 (0.171) & 6.780 (0.107) \\
        CAREER (\citet{vafa_career_2024}) & -- & 8.575 (0.132) & 8.551 (0.157) & 6.346 (0.100) \\
        \bottomrule
        \end{tabular}
    }
    \legend{Test-set-bootstrap standard errors are reported in parentheses.}
\end{table}

To interpret these results and the value added from incorporating text embeddings, we also compare performance against another simple baseline, a simple empirical transition probability model. Let $\#^\text{(train)}\{y\}$ and $\#^\text{(train)}\{y \to y'\}$ denote the count of occupation $y$ and the transitions $y \to y'$ in the training data, respectively. This model estimates the probability of transitioning from occupation $y$ to $y'$ (where all individuals are in the ``null'' occupation when $t=0$) as
$
\hat{P}_\text{Empirical}(y_{i, t} \mid x_{i, \leq t}, y_{i, < t}) =
\frac{\#^\text{(train)}\{y_{i, t-1} \to y_{i, t}\}+1}{\#^\text{(train)}\{y_{i, t-1}\}+1},
$ adding 1 to the numerator and denominator to handle zero counts.
This model predicts based solely on the immediately preceding occupation without using any covariates. Across all datasets, Table \ref{tab:perplexity-embedding-approach} shows that the embedding-based models have better performance than the empirical transition model, but worse than CAREER.

\subsection{Occupation Models Derived From Fine-Tuned Language Models}
\label{sec:fine-tuning-clm}

In this section, we analyze the performance of  occupational models based on LLMs fine-tuned on text templates created from our survey datasets.

\subsubsection{Comparing Performance Across Alternative Foundation Model Approaches}
\label{sec:comparingperformance}
Table \ref{tab:direct-token-prediction} reports the test set perplexity of the FT-LABOR-LLM occupation models along with several baselines. We see that FT-7B and FT-13B achieve substantially lower perplexity than CAREER, which was pre-trained on 24 million resumes and fine-tuned for occupation modeling on survey data. The differences between CAREER and FT-7B are about ten times larger than the test-set-bootstrap standard errors (defined in Section \ref{sec:appendix-bootstrap}) for PSID81 and NLSY79, while they are similar in size to the standard error for NLSY97, a substantially smaller dataset. FT-13B exhibits even larger performance improvements.
Online Appendix \ref{sec:online-appendix-edu-group} shows that FT-7B and FT-13B generally achieve lower perplexity than CAREER within education-defined subgroups, with CAREER slightly ahead only in the relatively small NLSY97 college subgroup.

As previewed in Section \ref{sec:appendix-bootstrap}, one question that naturally arises is whether sampling variation in the training set and randomness in the fine-tuning estimation algorithm lead to substantial variation in estimates of performance difference. In Appendix \ref{sec:appendix-bootstrap}, we carry out a small experiment with training-set-bootstrapping. The training-set-bootstrap standard errors for perplexity of FT-7B are 0.051, 0.058, and 0.020 for PSID81, NLSY79, and NLSY97, respectively (where to facilitate other comparisons, we included birth year in the estimation and these models are fine-tuned using pooled training set). These standard errors are smaller than those reported in Table \ref{tab:direct-token-prediction}. We calculate the training-set-bootstrap standard error for the difference between FT-7B and FT-13B only for PSID81, and found a standard error of 0.029, larger than that corresponding test set standard error. This exercise suggests that variation due to training is not negligible, and small performance differences that appear to be statistically distinguishable from zero using test-set-bootstrap standard errors could in fact arise due to training uncertainty. Due to the large cost of training-set-bootstrapping, we report test-set-bootstrap standard errors in the rest of the paper, but we are cautious in interpreting marginally significant results.

\begin{table}[hbtp]
    \centering
    \caption{
    Test-set perplexity and perplexity improvement for fine-tuned vs. baseline models.
    \updated{10/30/2025}
    }
    \label{tab:direct-token-prediction}
    \begin{tabular}{lccc}
        \toprule
        \multicolumn{1}{r}{\textbf{Dataset}} & PSID81 & NLSY79 & NLSY97 \\
        \multicolumn{1}{r}{\textbf{Number of Test Set Transitions}} & 61,772 & 51,593 & 29,951 \\
        \midrule
        \textbf{Perplexity}\\
        \midrule
        Empirical Transition Frequency & 14.647 (0.224) & 14.264 (0.271) & 10.050 (0.169) \\
        CAREER (without pre-training) & 8.999 (0.144) & 9.143 (0.171) & 6.785 (0.107) \\
        CAREER (\citet{vafa_career_2024}) & 8.575 (0.132) & 8.551 (0.157) & 6.350 (0.100) \\
        FT-7B & 8.184 (0.126) & 8.329 (0.147) & 6.350 (0.101) \\
        FT-13B & 8.140 (0.126) & 8.282 (0.145) & 6.326 (0.100) \\
        \midrule
        \textbf{Perplexity Improvement}\\
        \midrule
        PPL(CAREER)-PPL(FT-7B) & 0.391 (0.019) & 0.222 (0.024) & 0.001 (0.018) \\
        PPL(CAREER)-PPL(FT-13B) & 0.435 (0.020) & 0.269 (0.023) & 0.025 (0.016) \\
        PPL(FT-7B)-PPL(FT-13B) & 0.044 (0.011) & 0.047 (0.012) & 0.024 (0.009) \\
        \bottomrule
    \end{tabular}
    \legend{
    Test-set-bootstrap standard errors are in parentheses.
    }
\end{table}

Next, we turn to analyzing the sources of differences between the models. We compare performance for the task of predicting the binary outcome of whether workers move to a different job $\text{move}_{i,t} = \mathbf{1}\{y_{i,t} \neq y_{i,t-1}\}$; and separately, we analyze performance conditional on a transition involving a move. Changing occupations may be an outcome of particular interest in empirical studies of, e.g., the effects of layoffs, training programs, or technological change.

A standard way to evaluate the performance of alternative prediction models for binary outcomes is to compare the area under the Receiver Operating Characteristic curve (AUC-ROC) in the test set, which ranges from 0 (the worst possible model) to 1 (the best possible model). Table \ref{tab:roc-group-separate-datasets} shows that the FT-7B model has AUC-ROC of 0.784, slightly greater than CAREER at 0.776. The empirical transition frequency benchmark model has a much lower AUC-ROC of 0.639.

\begin{table}[!htbp]
    \caption{Area Under the ROC Curve (AUC-ROC). \updated{10/30/2025}}
    \label{tab:roc-group-separate-datasets}
    \centering
    \begin{tabular}{lcccc}
    \toprule
    \multicolumn{1}{r}{\textbf{Dataset}} & PSID81 & NLSY79 & NLSY97 & Aggregated  \\
    \multicolumn{1}{r}{\textbf{Number of Test Set Transitions}} & 55,560 & 49,096 & 28,153 & 132,809 \\
    \midrule
    Empirical & 0.653 & 0.636 & 0.604 & 0.639 \\
    OTS Llama-2-7B (with title) & 0.714 & 0.716 & 0.680 & 0.709 \\
    CAREER (without pre-training) & 0.768 & 0.762 & 0.739 & 0.761 \\
    CAREER & 0.778 & 0.778 & 0.763 & 0.776 \\
    FT-7B & 0.789 & 0.787 & 0.764 & 0.784 \\
    \bottomrule
    \end{tabular}
    \legend{For the off-the-shelf model, we use the Llama-2-7B model with the list of job titles included in the prompt.}
\end{table}

To assess how well-calibrated each model is, we split observations into ten groups based on deciles of predicted probability of changing occupations $\hat{P}(\text{move}_{i,t})$ (i.e., the next occupation $y_{i, t}$ is different from the previous one $y_{i, t-1}$), denoted as $G_1, G_2, \dots, G_{10}$. Then, for each group, we compute the empirical percentage of movers.
If a model is well-calibrated, the average predicted $\hat{P}(\text{move}_{i,t})$ should match the actual proportion of movers within the corresponding group in the test set.
We further calculate the average (over deciles) of the calibration error \resizebox{0.5\linewidth}{!}{$\sqrt{\frac{1}{10} \sum_{i=j}^{10} \left[\left(\sum_{(i, t) \in G_j} \mathbf{1}\{\text{move}_{i, t}\} \right) - \left(\sum_{(i, t) \in G_j} \hat{P}_\text{model}(\text{move}_{i,t})\right) \right]^2}$}.

\begin{figure}[!htbp]
    \centering
    \includegraphics[width=0.4\linewidth]{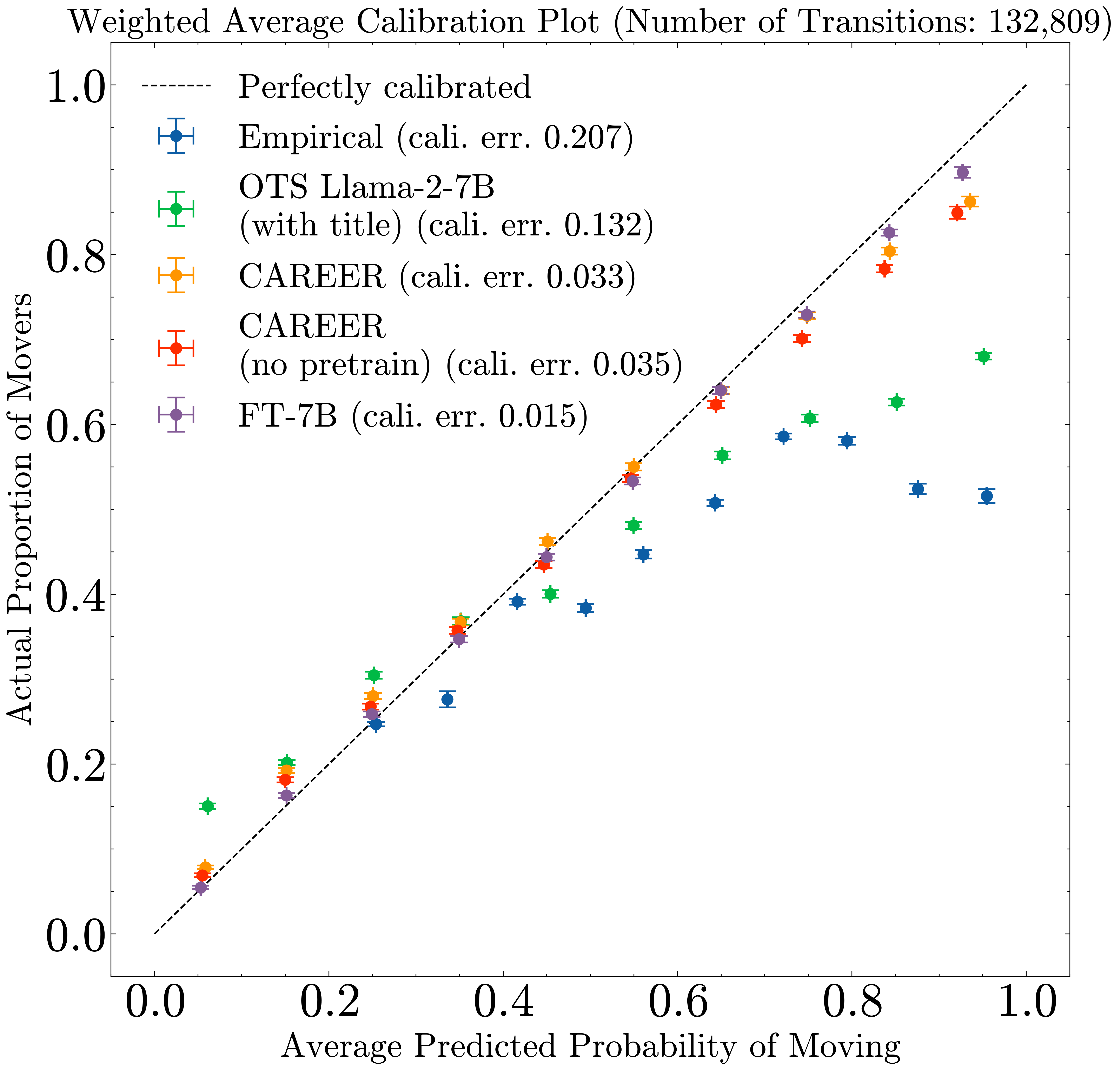}
    \caption{Calibration plots of baseline and fine-tuned models on the task of predicting staying in an occupation vs. moving occupations. \updated{10/30/2025}}
    \label{fig:binary-calibration-plot}
\end{figure}

Figure \ref{fig:binary-calibration-plot} illustrates calibration plots of the empirical transition frequency baseline, CAREER model, FT-7B model, as well as their corresponding calibration errors. The diagonal line in the plot represents a perfectly calibrated model.

We observe that our FT-7B model is better calibrated in predicting staying versus moving than the CAREER model, which underestimates moving in some groups and overestimates it in others. The CAREER model has a two-stage prediction design (i.e., predict staying versus moving, then next occupation sequentially), and the training process of CAREER pays special attention to enforcing the model calibration. In contrast, our LLM fine-tuning does not give special treatment to matching the empirical probability of staying, so it is somewhat surprising that it is better calibrated in this dimension than CAREER; with its extremely large parameter space, the FT-7B model appears to learn these probabilities without special treatment in the model.

Table \ref{tab:conditional_mover_perplexities} reports perplexity conditional on moving for alternative models, while the bottom panel reports differences between FT-LABOR-LLM models and CAREER. We see that the FT-7B and FT-13B models outperform all other models on PSID81 and NLSY79, and FT-13B achieved comparable performance to CAREER on NLSY97. Note that occupation transitions conditional on moving are inherently harder to predict, explaining the higher level of perplexity in this table.

\begin{table}[!htbp]
    \centering
    \caption{Test-set perplexity and perplexity improvement for fine-tuned vs. baseline models, conditional on moving. \updated{10/30/2025}}
    \label{tab:conditional_mover_perplexities}
    \begin{tabular}{lccc}
    \toprule
    \multicolumn{1}{r}{\textbf{Dataset}} & PSID81 & NLSY79 & NLSY97 \\
    \multicolumn{1}{r}{\textbf{Number of Test Set Transitions}} &  24,030  & 23,023 & 10,960 \\
    \midrule
    \textbf{Perplexity} & & & \\
    \midrule
    Empirical Transition Frequency & 59.982 (0.925) & 66.149 (0.800) & 72.311 (1.498) \\
    CAREER (without pre-training) & 26.813 (0.522) & 33.892 (0.503) & 40.822 (0.951) \\
    CAREER & 24.244 (0.419) & 30.312 (0.439) & 35.875 (0.769) \\
    FT-7B & 22.531 (0.387) & 29.419 (0.426) & 36.346 (0.810) \\
    FT-13B & 22.304 (0.389) & 28.978 (0.418) & 36.066 (0.807) \\
    \midrule
    \textbf{Perplexity Improvement}\\
    \midrule
    PPL(CAREER)-PPL(FT-7B) & 1.713 (0.125) & 0.893 (0.143) & -0.471 (0.221) \\
    PPL(CAREER)-PPL(FT-13B) & 1.940 (0.130) & 1.334 (0.155) & -0.191 (0.209) \\
    PPL(FT-7B)-PPL(FT-13B) & 0.227 (0.071) & 0.441 (0.095) & 0.280 (0.156) \\
    \bottomrule
    \end{tabular}
    \legend{Estimated conditional probabilities are calculated using Bayes' rule. Test sets are restricted to include individual-year observations that satisfy $y_{i, t} \neq y_{i, t-1}$.
    Test-set-bootstrap standard errors are in parentheses.
   }
\end{table}

Beyond studying whether workers change occupation, it may be of interest to study other outcomes derived from occupational transitions, such as transitions into unemployment, or transitions out of working (either to unemployment, education, or out of the labor force).  Online Appendix \ref{sec:online-appendix-alt-eval-tasks} shows that FT-LABOR-LLM, despite being fine-tuned only for the more granular next-occupation prediction task, achieves similarly strong relative performance for these alternative outcomes of interest.

Because the surveys we study were typically conducted every other year, FT-LABOR-LLM typically needs to make predictions about transitions separated in time by two years. However, it is possible to use FT-LABOR-LLM to make predictions about transitions for years that we did not directly observe, including the years between surveys. A potential limitation of FT-LABOR-LLM is that it may not be internally consistent when predicting transitions when survey observations are separated in time; the prediction that comes out of the LLM is not constrained to be equal to the result if we were to make sequential predictions for each year and combine them via Bayes' rule. In particular, the probability FT-LABOR-LLM models assign to $y_{i,t}$ when $\text{year}_{i,t}=\text{year}_{i,t-1}+2$ is not necessarily equal to the estimated probability by applying the model year-by-year, composing its predictions about $\text{year}_{i,t}=\text{year}_{i,t-1}+1$ and predictions about $\text{year}_{i,t}$ conditional on potential occupations in $\text{year}_{i,t-1}+1$. In Appendix \ref{sec:appendix-gap-year}, we compare the model's direct predictions about a transition across two years with those constructed based on a sequence of two one-year-ahead predictions and show that the correlation between the two predictions is 0.93, meaning that the model appears to correctly account for the gap in calendar time when making predictions. We leave further exploration of this issue for future work.

\subsubsection{The Value of Fine-Tuning: Using Off-The-Shelf Large Language Models as Occupation Models}

In this section, we evaluate the importance of fine-tuning. We report results about the performance of occupation models based on off-the-shelf (rather than fine-tuned) LLMs, applying Equation (\ref{eq:LLMProbModel}) to estimate $\hat{P}_\text{LLM}$ for several alternative LLMs.\footnote{To improve computational efficiency for prediction, we quantize all LLMs in this paper to 8-bit precision (i.e., compress model weights to 8-bit integers to speed inference with minimal precision loss) while running model inference. We perform full-precision inference on a few subsamples, and the difference in performance was small. See Online Appendix \ref{sec:online-appendix-full-prec-quant} for more details on full-precision versus quantized model analysis.} Because evaluating perplexity requires accessing a model's assigned probabilities, we restrict attention to open-source LLMs where it is possible to obtain predicted probabilities directly, with the exception of Section \ref{sec:validjobtitles} of Online Appendix \ref{sec:online-appendix-detailed-ots}, where we evaluate the ability of OpenAI gpt-4o-mini to produce valid job titles in response to a prompt.

In particular, we study open-source LLMs from the Llama family of models: Llama-2, Llama-3.1, and Llama-3.2. For example, Llama-2 models were trained by Meta on approximately 2 trillion tokens of text, much of it from the Internet, and are among the most capable open-source LLMs currently available (\citet{touvron_llama_2023}). We do not study bigger models such as Llama-2-70B and Llama-3.1-405B because evaluating this model across many variations requires substantial cost.

\begin{table}[!htbp]
\centering
\caption{Test-set perplexity for off-the-shelf LLMs vs. CAREER. \updated{10/30/2025}}
\label{tab:main-text-baseline-llm-performances}
\begin{tabular}{lccc}
\toprule
\multicolumn{1}{r}{\textbf{Dataset}} & PSID81 & NLSY79 & NLSY97 \\
\multicolumn{1}{r}{\textbf{Number of Test Set Transitions}} & 61,772 & 51,593 &  29,951 \\
\midrule
\textbf{Model} & & & \\
\midrule
OTS Llama-2-7B & 364.655 (12.676) & 288.466 (9.203) & 237.219 (7.141) \\
OTS Llama-2-13B & 140.046 (4.931) & 123.778 (4.438) & 101.510 (3.114) \\
OTS Llama-3.1-8B & 126.532 (4.510) & 126.429 (4.736) & 90.360 (2.777) \\
OTS Llama-3.2-1B & 467.296 (22.175) & 395.788 (19.200) & 256.675 (11.288) \\
OTS Llama-3.2-3B & 167.218 (6.396) & 147.557 (5.619) & 114.283 (3.825) \\
\bottomrule
\end{tabular}

\legend{Test-set-bootstrap standard errors are reported in parentheses.
}

\end{table}

Table \ref{tab:main-text-baseline-llm-performances} contains the perplexity of off-the-shelf LLMs, which are substantially higher than other models discussed.
\footnote{For reference, a completely uninformative model that assigns uniform mass to each possible occupation would achieve a perplexity of $|\mathcal{Y}|$, which is 335.}
The unsatisfactory performance of off-the-shelf LLMs can be attributed to two factors: off-the-shelf LLMs are not adapted to the career trajectory distributions in our survey dataset, and these LLMs do not know the set of valid job titles to predict.
To better understand the poor performance of the model based on off-the-shelf LLMs, we assess the responses that the LLMs provide when prompted with examples of tokenized text templates. Online Appendix \ref{sec:online-appendix-llm-trajectory-examples} provides some examples. While the responses appear plausible, the LLMs also assign mass to strings that are not valid job titles.

Online Appendix \ref{sec:online-appendix-detailed-ots} explores alternative prompting strategies that direct LLMs to restrict prediction to valid occupations when computing occupation probabilities. While we observe notable reductions in perplexity with these prompting strategies, the resulting model perplexities remain orders of magnitude above those achieved by CAREER. As mentioned, LLMs can sometimes hallucinate and produce invalid job titles; however, these alternative prompting strategies help mitigate this issue. Off-the-shelf Llama variants and GPT-4o-mini generate valid job titles with probabilities climbing from roughly 0.6 to above 0.9 as prompts add explicit job-title lists and up to ten in-context resume examples. Online Appendix \ref{sec:online-appendix-detailed-ots} presents more details.

\section{Value of Data and Model Size}
In this section, we analyze the roles of model complexity (number of parameters) and of quantity of data in determining performance. As discussed in the introduction, analysts using fine-tuned LLMs will need to consider costs of computation in the fine-tuning process, as well as when making predictions from the model, costs which increase with model complexity. These costs may be traded off against improved accuracy from more complex models.
Another tradeoff arises when acquiring more data: extra data may come from a different context and thus correspond to a different data-generating process.
Incorporating non-representative data in fine-tuning may or may not improve performance.

We empirically evaluate these tradeoffs by varying the datasets used for fine-tuning, for example, by combining datasets, while holding the three test sets fixed.
To facilitate our discussion, we use $\mathcal{D}_{\text{data}}^\text{(split)}$ to denote a particular split of the dataset $\omega$, for example, $\mathcal{D}_\text{PSID81}^\text{(train)}$ represents the training split of the PSID81 dataset.
We explore the consequences of fine-tuning based on a different survey dataset, or combinations of survey datasets, than the survey from which the test set is drawn. Recall that all three of the survey datasets we analyze are approximately representative of the U.S. population, but as shown in Section \ref{sec:survey-datasets}, they incorporate different distributions of calendar year, as well as different conditional distributions of birth year for each calendar year, where we include both of these variables in the text templates for the analyses in this section.

In our first exercise, reported in Table \ref{tab:cross-dataset-eval}, we evaluate models fine-tuned using the training split of one survey data, $\mathcal{D}_\omega^\text{(train)}$, and the test split from another dataset, $\mathcal{D}_{\omega'}^\text{(test)}$, with $\omega \neq \omega'$.  This exercise shows how training from data with very different distributions of birth year and calendar year affects performance; since FT-7B and FT-13B are trained using information about both of these variables and the transformer neural network allows for rich interactions, in principle, the model could be flexible enough to predict well across distributions. In particular, the PSID81 dataset has overlap in terms of calendar year and birth year with both NLSY datasets, and it is substantially larger overall. However, the results illustrate significantly degraded predictive performance when the training data and test data are from different survey datasets.

\begin{table}[!htbp]
    \centering
    \caption{Fine-tuning on training set of dataset $\omega$ and evaluating on test split of dataset $\omega'$. \updated{10/30/2025 (no change)}}
    \label{tab:cross-dataset-eval}
    \begin{tabular}{llccc}
    \toprule
     \multicolumn{2}{r}{\textbf{Evaluation Dataset}} & PSID81 & NLSY79 & NLSY97 \\
    \multicolumn{2}{r}{\textbf{Number of Test Set Transitions}} & 61,772 & 51,593 & 29,951 \\
    \midrule
    \textbf{Foundation Model} & \textbf{Fine-tuning Dataset} \\
    \midrule
    FT-7B & PSID81 & 8.184 (0.126) & 10.705 (0.198) & 10.523 (0.154) \\
    FT-7B & NLSY79 & 9.929 (0.154) & 8.329 (0.147) & 7.965 (0.123) \\
    FT-7B & NLSY97 & 12.638 (0.213) & 11.274 (0.228) & 6.350 (0.101) \\
    \midrule
    FT-13B & PSID81 & 8.140 (0.126) & 10.165 (0.190) & 9.246 (0.135) \\
    FT-13B & NLSY79 & 10.069 (0.154) & 8.282 (0.145) & 7.603 (0.114) \\
    FT-13B & NLSY97 & 12.849 (0.211) & 10.929 (0.217) & 6.326 (0.100) \\
    \bottomrule
    \end{tabular}
    \legend{Test-set-bootstrap standard errors in parentheses.}
\end{table}

Next, we evaluate the value of data by first pooling all training data from survey datasets together, so that $\mathcal{D}_\text{all}^\text{(train)} = \bigcup_{\omega \in \text{\{PSID81, NLSY79, NLSY97\}}}\mathcal{D}_\omega^\text{(train)}$.
Then, we sample $P\%$ of individuals from $\mathcal{D}_\text{all}^{\text{(train)}}$ and use the sample to fine-tune a Llama-2-7B model. Finally, we evaluate the FT-LABOR-LLM on the test split of each survey dataset separately.
Table \ref{tab:value-of-info-pooled-training-data} summarizes the performance of these models. The model's performance improves as we increase the amount of training data (i.e., raise the value of $P$), and the returns to data are diminishing. On the test split of dataset $\omega$, models fine-tuned on the aggregated dataset eventually outperform the model fine-tuned on the corresponding training set $\mathcal{D}_{\omega}^\text{(train)}$, when $P \geq 80$. In addition, the models fine-tuned on the pooled data with FT-7B eventually outperform FT-13B trained on the individual baseline training sets, showing that adding data, even data from different distributions, can substitute for model complexity. Note, however, that the improvement on PSID81 is small enough (0.06) relative to the test-set-bootstrap standard error that the uncertainty derived from training may be large enough to overturn the statistical significance of the result. Indeed, in Appendix \ref{sec:appendix-bootstrap} we find a training-set-bootstrap standard error of 0.055 for this improvement, which together with test-set uncertainty would render the improvement not statistically significant.

\begin{table}[!htbp]
    \centering
    \caption{Fine-tuning on $P\%$ of the mixture of training splits of three datasets.\updated{10/30/2025 (no change since 12/02/2024)}}
    \label{tab:value-of-info-pooled-training-data}
    \begin{tabular}{lccc}
    \toprule
    \multicolumn{1}{r}{\textbf{Evaluation Dataset}} & PSID81 & NLSY79 & NLSY97 \\
    \multicolumn{1}{r}{\textbf{Number of Test Set Transitions}} &  61,772 & 51,593 & 29,951  \\
    \midrule
    \textbf{Perplexity} & & & \\
    \midrule
    FT-7B with Corresponding Training Set & 8.184 (0.126) & 8.329 (0.147) & 6.350 (0.101) \\
    FT-13B with Corresponding Training Set & 8.140 (0.126) & 8.282 (0.145) & 6.326 (0.100) \\
    \midrule
    FT-7B with 20\% of Pooled Data & 8.774 (0.137) & 8.831 (0.162) & 6.526 (0.103) \\
    FT-7B with 40\% of Pooled Data & 8.386 (0.130) & 8.480 (0.152) & 6.342 (0.100) \\
    FT-7B with 60\% of Pooled Data & 8.256 (0.127) & 8.335 (0.149) & 6.256 (0.098) \\
    FT-7B with 80\% of Pooled Data & 8.151 (0.126) & 8.258 (0.147) & 6.214 (0.097) \\
    FT-7B with 100\% of Pooled Data & 8.082 (0.124) & 8.207 (0.146) & 6.194 (0.097) \\
    \midrule
    \textbf{Perplexity Improvement} & & &\\
    \midrule
    PPL(FT-13B)-PPL(FT-7B-20\%) & -0.635 (0.024) & -0.549 (0.026) & -0.200 (0.016) \\
    PPL(FT-13B)-PPL(FT-7B-40\%) & -0.246 (0.017) & -0.198 (0.016) & -0.016 (0.013) \\
    PPL(FT-13B)-PPL(FT-7B-60\%) & -0.116 (0.014) & -0.053 (0.015) & 0.070 (0.012) \\
    PPL(FT-13B)-PPL(FT-7B-80\%) & -0.011 (0.013) & 0.024 (0.014) & 0.112 (0.012) \\
    PPL(FT-13B)-PPL(FT-7B-100\%) & 0.057 (0.014) & 0.075 (0.015) & 0.132 (0.013) \\
    \bottomrule
    \end{tabular}
    \legend{Test-set-bootstrap standard errors in parentheses.}
\end{table}

In Appendix \ref{sec:appendix-VOI}, we consider another variation of the analysis, incrementally adding pooled data to the full baseline training set for a given survey. We find that adding the data from other surveys to the full baseline training set immediately improves performance, and increasing the training set size by 30\% allows FT-7B to match or surpass the performance of FT-13B.
\section{Sources of Performance Gains}\label{sec:analyses}

Our analyses demonstrate that our best-performing model, directly predicting occupations through text tokens using FT-LABOR-LLM, achieves superior perplexity scores compared to the previous state-of-the-art CAREER model. This section delves deeper into the sources of performance differences.

\subsection{Language Models using Numeric Job Titles}
\label{sec:numerictitles}
 One key difference between LLMs and baseline models, besides the number of parameters, is that the LLMs have an understanding of textual data.
This section examines whether LLMs' performance is driven by their rich, deep neural network architecture or their advanced understanding of the meaning of occupations based on textual data.
To do so, we create an alternative prediction space with ``numeric job titles'' only.
We assign each occupation $y \in \mathcal{Y}$ a randomly chosen numeric job title (in contrast to their original literal job title) from \texttt{job\_000}, \texttt{job\_001}, ....(e.g., \texttt{Cashiers} is mapped to \texttt{job\_045}); all numeric job titles have three digits.
Then, we replace all original literal job titles in the text representation with their corresponding numeric job titles, denoted as $\texttemplate^\text{(numeric)}(x_{i, \leq t},y_{i, < t})$. Online Appendix \ref{sec:online-appendix-numerical-job-titles} provides an example of career history text representations with numeric job titles.

For each survey dataset, we fine-tune the Llama-2-7B model using the training corpus with numeric job titles only, and denote that fine-tuned model as FT-7B-NUMERIC; then, we compare FT-7B-NUMERIC to FT-7B fine-tuned on the corresponding training split of a single survey data.
While evaluating performance, we use the conditional probability of numeric job titles assigned by the LLM. For example, the predicted probability of the next occupation being cashier is $P_\text{LLM}(\texttt{job\_045} \mid \texttemplate^\text{(numeric)}(x_{i, \leq t},y_{i, < t}))$ instead of $P_\text{LLM}(\text{Cashier} \mid \texttemplate(x_{i, \leq t},y_{i, < t}))$, where historical occupations are also replaced with numbers.

Table \ref{tab:numeric-7B} shows the performance of FT-7B-NUMERIC, which performed much worse than the FT-7B model using literal job titles. Our results indicate that an important contributor to the LLM's performance comes from the LLM's ability to use the textual content of the occupation title to make predictions; using numeric job titles disassociates this knowledge from the prediction task and hurts performance significantly.

\begin{table}[!htbp]
    \centering
    \caption{Test-set perplexity and perplexity improvement on literal vs. numeric job titles.\updated{10/30/2025 (no change since 11/11/2024)}}
    \label{tab:numeric-7B}
    \begin{tabular}{lccc}
    \toprule
    \multicolumn{1}{r}{\textbf{Evaluation Dataset}} & PSID81 & NLSY79 & NLSY97 \\
    \multicolumn{1}{r}{\textbf{Number of Test Set Transitions}} &  61,772 & 51,593 & 29,951 \\
    \midrule
    \textbf{Perplexity}\\
    \midrule
    PPL(FT-7B) &  8.184 (0.126) & 8.329 (0.147) & 6.350 (0.101) \\
    PPL(FT-7B-NUMERIC) & 8.830 (0.141) & 9.129 (0.168) & 6.720 (0.105) \\
    \midrule
    \textbf{Perplexity Improvement}\\
    \midrule
    PPL(FT-7B-NUMERIC)-PPL(FT-7B) & 0.647 (0.027) & 0.800 (0.031) & 0.370 (0.021) \\
    \bottomrule
    \end{tabular}
    \legend{
    Test-set-bootstrap standard errors are in parentheses.}
\end{table}

\subsection{Additional Analyses}
\label{sec:additiona-analyses}

This section describes several additional analyses that shed light on the sources of performance improvements.

We evaluate the importance of demographic variables (gender, ethnicity, and region) for the fine-tuned model's predictive performance by randomly replacing these covariates in the test set with values from the validation set and measuring the impact on perplexity. The analysis reveals that randomizing gender alone significantly degrades performance (e.g., increasing perplexity by  $\sim 12\%$ in PSID81), while ethnicity alone has about one-quarter of gender's effect, and region has minimal impact. Simultaneously randomizing all three variables yields the greatest performance decline, demonstrating that complex interactions among demographic covariates are crucial for prediction. Appendix \ref{sec:appendix-sensitivity-to-input-features} provides more details.

In Appendix \ref{sec:appendix-value-of-career-history}, we test how much historical information the model uses to predict the next occupation by predicting with prompts that only include the last $k \in \{5, 10, 15, 20, 25\}$ observations and evaluate the perplexity lift within age-banded test subsets. Results indicate that adding history from five to 10-15 prior records lowers perplexity sharply, while additional context delivers diminishing returns: for later career transitions ($t > 20$), most predictive information is captured within just 10-15 years of history.

Appendix Table \ref{tab:embeddings-literal-numeric} shows the extent to which the embeddings created by FT-7B fine-tuned using PSID81, the largest dataset, incorporate more information about the meaning of job titles. One way to approach this analysis is to assess the predictive power of these embeddings on a task that relates to the interpretation of the titles. We consider a particular task that requires such an understanding: predicting which part of the occupation code hierarchy a particular occupation falls into (this information was not used in LABOR-LLM, although it may have been one part of the enormous pre-training corpus for the original Llama models). We compare the predictions derived from a multinomial logistic regression using as features embeddings extracted from FT-7B and CAREER. We show that the embeddings from FT-7B have a test-set accuracy of 82\% for predicting the correct SOC group for an occupation, which is better than CAREER (78\%).

In a second exercise (Appendix \ref{sec:appendix-GRF}), we characterize the economic conditions under which the language-based LABOR-LLM (FT-13B) outperforms the CAREER baseline for ``mover'' transitions in the test split of the PSID81 dataset. While CAREER is a sophisticated transformer that effectively captures the ``structural syntax'' of occupational sequences via embeddings, it only relies on discrete categorical occupational codes. In contrast, LABOR-LLM leverages the semantic richness of natural language encoded in the literal titles. Using a Generalized Random Forest (GRF) to analyze heterogeneity in model performance, we find the language-based model provides the highest value-added where transitions are less predictable: during ``complex'' labor market adjustments that cross occupational boundaries (different SOC groups) or involve theoretically distant skill sets (low O*NET similarity). This suggests that the LLM uncovers semantic linkages between occupations that administrative codes obscure. Furthermore, LABOR-LLM's advantage grows with the length of the career history (i.e., for later transitions in the worker's career history), suggesting that textual representations better encode human capital accumulation than the categorical occupation codes of CAREER.

\section{Conclusion}
This paper proposes a novel approach, LABOR-LLM, to the problem of predicting a worker's next occupation conditional on history. The best-performing version of this approach, FT-LABOR-LLM, translates the tabular data about a worker's history from publicly available U.S. surveys (PSID and NLSY) into text files that resemble resumes, and then fine-tunes the Llama-2 open-weight foundation models on that corpus. Then, to estimate the probability that the next occupation is a particular occupation, say ``engineer,'' given worker history, the approach prompts the fine-tuned LLM with the textual version of the worker's history, and extracts the probability that the LLM assigns to the text ``engineer'' as the next word. We show that, off-the-shelf, without fine-tuning, this approach performs poorly even when OpenAI's API is used. However, the fine-tuning leads this approach to outperform all existing benchmarks. The fine-tuning eliminates the problem of occupation title ``hallucinations,'' but more importantly, it leads the model to make accurate predictions about conditional probabilities in held-out test data.  Accurate, fine-grained predictions enable economists to ask and answer more nuanced questions, and to improve the quality of causal inference analyses that rely on accurate predictions.

The paper explores some of the sources of the strong performance of FT-LABOR-LLM, showing that representative data is important, but that adding more data (even non-representative data) can lead a smaller model (in terms of number of parameters) to outperform a larger one. The paper also shows that FT-LABOR-LLM makes use of many years of history, even a worker's early career history, to improve prediction quality. Our results illustrate that the approach can be effective in datasets of moderate size (tens of thousands of transitions), leveraging the general information about occupations embedded in the open-weight LLM's embeddings of the text of job titles and resumes.

An advantage of the FT-LABOR-LLM approach is that all data and software necessary to apply this approach is available publicly, including the weights of the LLM, so that the main cost in practice is the cost of the computing for fine-tuning and making predictions. Low-cost cloud-based services are available (we used the service provided by Together AI) that enable fine-tuning by simply uploading documents; with these services, no coding is required for the training step, and minimal original coding is required to obtain predictions from the fine-tuned LLM.  Thus, researchers can focus on analyzing the results and performing downstream empirical exercises. However, a limitation to this approach is that fine-tuning can become expensive as the dataset size grows, and repeatedly fine-tuning (for example, to bootstrap standard errors) can be prohibitively expensive.

It is important to recognize that LLM-based approaches share a common limitation with all predictive models used in economic analysis: their performance can deteriorate when the underlying data-generating process changes. These models rely on historical patterns and, therefore, may generalize poorly during periods of structural change, such as rapid AI adoption or other macroeconomic disruptions, when past transitions no longer reflect future dynamics.
This limitation underscores the need for caution in such settings and points to future work, such as combining LLM-based prediction with structural economic models.

Using a textual representation of resumes has the advantage of incorporating other types of data into the analysis. For example, some datasets include job descriptions or descriptions of accomplishments as part of a resume; the LABOR-LLM framework makes this type of information straightforward to incorporate.

An approach based on publicly available foundation models may also be useful in other settings, for example, any economic prediction problems that involve discrete outcomes with many alternatives and where the alternatives may be associated with meaningful textual descriptions. A sequence of purchases made by a consumer may have a similar structure. Our paper also illustrates the importance of fine-tuning: off-the-shelf LLMs may make plausible sounding predictions, but without fine-tuning they are unlikely to give accurate conditional probabilities for any particular dataset of interest.

%%%%%%%%%%%%%%%%%%%%%%%%%%%%%%%%%%%%%%%%%%%%%%
%% Single Appendix:            %%
%%%%%%%%%%%%%%%%%%%%%%%%%%%%%%%%%%%%%%%%%%%%%%
%\begin{appendix}
%\section*{???} %% if no title is needed, leave empty \section*{}.
%\end{appendix}
%%%%%%%%%%%%%%%%%%%%%%%%%%%%%%%%%%%%%%%%%%%%%%
%% Multiple Appendices:        %%
%%%%%%%%%%%%%%%%%%%%%%%%%%%%%%%%%%%%%%%%%%%%%%
\newpage
\begin{appendix}

\section{Quantifying Uncertainty in Performance Metrics}
\label{sec:appendix-bootstrap}
When comparing the performance of alternative occupation models, we wish to quantify the uncertainty about estimates of performance.
The randomness in measured perplexity for a given model arises from several sources: sampling variation in the training data, randomness in the fine-tuning pipeline (e.g., data shuffling for a stochastic gradient descent optimizer), and sampling variation of the test data.

To estimate the uncertainty arising from the first two sources, we bootstrap the training set used for fine-tuning (sampling at the individual level) and estimate the variation in measures of the performance of models across bootstrap samples. We refer to the resulting standard errors as ``training-set-bootstrapped.'' To capture sampling variation of the training set, we sample with replacement.

According to the support team of \href{https://www.together.ai/}{Together AI}, the platform we use to fine-tune LLMs, the randomness in their fine-tuning process arises mainly from randomizing the order of observations in the process of optimizing via stochastic gradient descent; each instance of re-tuning a bootstrap sample will include randomization of this type.\footnote{Unfortunately, at the time of this writing, there is no way to specify the random seed for reproducibility. The support team also mentioned ``adapter weight initialization'' as another source of randomness in the fine-tuning pipeline, which is only relevant if one is fine-tuning using the Low-Rank Adaptation (LoRA) technique. We are doing full-parameter fine-tuning instead.} Because fine-tuning is very expensive to carry out, we conduct an experiment for three of the models, as described below in Section \ref{sec:comparingperformance} and Appendix \ref{sec:appendix-bootstrap}.

To estimate the uncertainty arising from sampling variation in the test set, we bootstrap the test set and refer to the resulting standard errors as ``test-set-bootstrap.''  We sample at the individual level with replacement and, in the analysis we report below, use 100 bootstrap replications.

We employ a similar bootstrapping approach to calculate the test-set-bootstrap standard errors for the differences in perplexities between the two models. We select bootstrap samples at the individual level, compute the perplexities for both models on the bootstrap sample, and calculate the standard deviation of the difference in perplexities.

\subsection{Test Set Bootstrap for Test Set Variations}
The purpose of our bootstrapping approach for the test set is to estimate the sensitivities of our metrics (e.g., perplexities and differences in perplexities) to changes in the test set distribution. In the main paper, we first report the metric (e.g., perplexity) computed using all observations in our test set. Then, we create $B$ bootstrap samples of the test set, sampled on the individual level, to estimate the standard error of the metric.
For each bootstrap iteration $b$, we sample individuals in the test set with replacement, then we collect all individual-year observations associated with these sampled individuals and the log-likelihood values assigned to these observations by the model. We use these log-likelihood values to compute the $b^\text{th}$ bootstrap value of the metric (e.g., perplexity). After repeating the process above $B$ times, we estimate the standard error using the standard deviation of the $B$ bootstrap values.

We call this procedure the test-set-bootstrap, and report the standard error estimation from the test-set-bootstrap along with our metrics in this paper.

\subsection{Training Set Bootstrap for Uncertainty Training Set and Training Pipeline}
Similarly, the purpose of our bootstrapping approach for the training set is to quantify the uncertainty in model performance due to training set variation and randomness in the training pipeline, primarily due to data shuffling (according to the support team at Together AI). We create 12 bootstrapped training sets by sampling the pooled training set (i.e., the union of the training splits of PSID81, NLSY79, and NLSY97) with replacement.
Let $\mathcal{D}_\text{mixture}^\text{(train, s)}$ denote the bootstrapped training data of the mixture dataset (sampled with replacement), generated using the random seed $s \in \{0, 1, ..., 11\}$.
We fine-tune 12 versions of the Llama-2-7B models using these mixture dataset bootstrapped training sets and evaluate their performance using the \emph{complete} test split of each dataset. We call this procedure the training-set-bootstrap.

To better understand how uncertainty from the training set impacts the comparisons we make in this paper, we also fine-tune 12 versions each of our Llama-2-7B and Llama-2-13B models using the PSID81, our largest survey dataset, subset of each $\mathcal{D}_\text{mixture}^\text{(train, s)}$, denoted by $\mathcal{D}_\text{PSID81}^\text{(train, s)}$.\footnote{We did not perform the same exercise for NLSY79 and NLSY97 due to the computational cost.} We fine-tune these additional models to make two comparisons. First, we compare the FT-7B fine-tuned on the mixed data to the same model fine-tuned on only one dataset to understand how the training-set-bootstrap impacts our value of information analysis. Second, we compare the Llama-2-7B model fine-tuned on PSID81 to the Llama-2-13B model fine-tuned on the same data to learn how our analysis of smaller versus larger models is impacted by the training-set-bootstrap. The results for all three models and the two comparisons are shown in Table \ref{tab:train-set-bootstrap-results}.

We observe that uncertainty from the training-set-bootstrap is lower than the uncertainty from the test-set-bootstrap for perplexity; however, the training-set-bootstrap uncertainty is relatively higher than the test-set-bootstrap uncertainty when it comes to the perplexity differences.
Because the computational cost is prohibitive since it involves multiple rounds of large language model fine-tuning, we did not perform the training-set-bootstrap in the main paper.

\begin{table}[!htbp]
    \caption{Test-set perplexity for models fine-tuned 12 times, with training-set-bootstrap standard errors. \updated{10/30/2025 (no change since 12/04/2024)}}
    \label{tab:train-set-bootstrap-results}
    \centering
    \begin{tabular}{lccc}
        \toprule
        \multicolumn{1}{r}{\textbf{Evaluation Dataset}} & PSID81 & NLSY79 & NLSY97 \\
\multicolumn{1}{r}{\textbf{Number of Test Set Transitions}} &  61,772 & 51,593 & 29,951 \\
        \midrule
        \textbf{Perplexity} & & & \\
        \midrule
        FT-7B with $\mathcal{D}_\text{mixture}^\text{(train, s)}$ & 8.373 (0.051)  & 8.486 (0.058) & 6.358 (0.020) \\
        FT-7B with $\mathcal{D}_\text{PSID81}^\text{(train, s)}$ & 8.492 (0.034) & - & - \\
        FT-13B with $\mathcal{D}_\text{PSID81}^\text{(train, s)}$ & 8.458 (0.021) & - & - \\
        \midrule
        \textbf{Perplexity Improvement} &  & & \\
        \midrule
        PPL(FT-13B with $\mathcal{D}_\text{PSID81}^\text{(train, s)}$) - PPL(FT-7B with $\mathcal{D}_\text{mixture}^\text{(train, s)}$) & 0.085 (0.055) & - & - \\
        PPL(FT-7B with $\mathcal{D}_\text{PSID81}^\text{(train, s)}$) - PPL(FT-13B with $\mathcal{D}_\text{PSID81}^\text{(train, s)}$) & 0.034 (0.029) & - & - \\
        \bottomrule
    \end{tabular}
    \legend{Numbers in parentheses show the standard deviation of metrics (i.e., perplexity or perplexity difference) computed using 12 random seeds, i.e., the training-set-bootstrap standard error. These standard deviations measure \textit{solely} uncertainties due to training set variation and randomness in the training pipeline.
    }
\end{table}

\section{Details for Text Template }
\label{sec:appendix-text-template}

This appendix describes the text template in more detail. The text template starts with a preamble that describes the individual's static covariates, then lists the individual's education level and occupation for each calendar year. Specifically:
\begin{enumerate}
    \item The first line describes the source of the data, e.g., \texttt{<A worker from the PSID dataset>}.
    \item The second line describes the individual's demographic characteristics, e.g., \texttt{The following information is available about the work history of a female black or african american US worker residing in the south region}.
    \item The third line describes the individual's birth year, e.g., \texttt{The worker was born in 1963}.
    \item The fourth line describes the structure of the resume, i.e., \texttt{The worker has the following records of work experience, one entry per line, including year, education level, and the job title:}. This line is constant for all individuals and is useful for the LLM to understand the format of the subsequent rows of work experience.
    \item Starting from the fifth line, each line summarizes the information of the worker from a wave of the survey she participated in, including the calendar year, education level, and title of her main occupation reported in that survey year. Specifically, it is in the format \texttt{YEAR (EDUCATION): JOB TITLE}, e.g., \texttt{1984 (some college): Cooks}.
    \item The template ends with the line \texttt{<END OF DATA>}.
\end{enumerate}

The following example shows a complete text template of an individual worker. For more examples, see Online Appendix \ref{sec:online-appendix-text-template-example}.

{
\footnotesize\begin{Verbatim}[breaklines=true]
<A worker from the PSID dataset>
The following information is available about the work history of a female black or african american US worker residing in the south region.
The worker was born in 1963.
The worker has the following records of work experience, one entry per line, including year, education level, and the job title:
1984 (some college): Cooks
1985 (some college): Food servers, nonrestaurant
1986 (some college): Cleaners of vehicles and equipment
1988 (some college): Food servers, nonrestaurant
1989 (some college): Bus drivers
1990 (some college): Food servers, nonrestaurant
1991 (some college): Unemployed
1992 (some college): Painting workers
1993 (some college): Painting workers
1994 (some college): Court, municipal, and license clerks
1996 (some college): Septic tank servicers and sewer pipe cleaners
<END OF DATA>
\end{Verbatim}
}

The survey dataset may have missing data for certain individuals in some years, as described in Appendix \ref{sec:appendix-data}. This missingness can occur if a worker did not respond to a particular wave of the survey but participated in later waves. Additionally, some surveys, such as the NLSY and PSID, have transitioned from annual to biennial surveys in recent years, resulting in gaps for certain years.

The text template only has rows corresponding to the years when the individual was observed.

\section{Details for Obtaining the Probability Assigned to a Token}
\label{sec:appendix-predict-as-tokens}
In this appendix, we explain the details of how to directly leverage LLMs' next token prediction capabilities to predict future occupations using job titles described in Section \ref{sec:UsingLLMs}.
To obtain the predicted probability of the next occupation, we first tokenize each job title, $\text{title}_y$, into a sequence of tokens.
Suppose the string $\text{title}_y$ is tokenized into $n$ tokens $\{ \text{token}_y^{(1)}, \text{token}_y^{(2)}, \dots, \text{token}_y^{(n)} \}$. Then, the unnormalized probability of predicting $y$ is the likelihood the language model assigns to the token sequence $\{ \text{token}_y^{(1)}, \text{token}_y^{(2)}, \dots, \text{token}_y^{(n)} \}$ as the continuation of the text representation $\texttemplate(x_{i, \leq t}, y_{i, < t})$.
The predicted probability can further be expanded using the chain rule of probability, as shown in Equation (\ref{eq:predication}).
\begin{align}
    \begin{aligned}
        &\hat{P}^\mathcal{V}_\text{LLM}(\tokenize(\jobtitle(y)) \mid \texttemplate(x_{i, \leq t}, y_{i, < t})) \\
        &= \hat{P}^\mathcal{V}_\text{LLM}(\{ \text{token}_y^{(1)}, \text{token}_y^{(2)}, \dots, \text{token}_y^{(n)} \} \mid \texttemplate(x_{i, \leq t}, y_{i, < t})) \\
        &= \prod_{j=1}^{n} \hat{P}^\mathcal{V}_\text{LLM}(\text{token}_y^{(j)} \mid \texttemplate(x_{i, \leq t}, y_{i, < t}), \text{token}_y^{(1)}, \text{token}_y^{(2)}, \dots, \text{token}_y^{(j-1)} )
    \end{aligned}
    \label{eq:predication}
\end{align}
The $\hat{P}^\mathcal{V}_\text{LLM}(\text{token}_y^{(j)} \mid \texttemplate(x_{i, \leq t}, y_{i, < t}), \text{token}_y^{(1)}, \text{token}_y^{(2)}, \dots, \text{token}_y^{(j-1)})$ is operationalized by (1) appending all tokens $\text{token}_y^{(1)}, \text{token}_y^{(2)}, \dots, \text{token}_y^{(j-1)}$ to the text representation $\texttemplate(x_{i, \leq t}, y_{i, < t})$ and (2) querying the likelihood the language model assigned to $\text{token}_y^{(j)}$ as the next token conditioned on all the previous tokens.

For example, the title ``software engineer'' may be tokenized into two tokens, one for ``software'' $\in \mathcal{V}_\text{LLM}$ and one for ``engineer'' $\in \mathcal{V}_\text{LLM}$.\footnote{This is for illustration purposes only, how the LLM's tokenizer splits the phrase ``software engineer'' depends on the exact LLM used.}
Equation (\ref{eq:software-engineer-pred-example}) illustrates how to obtain the conditional probability assigned to ``software engineer''.
\begin{align}
    \begin{aligned}
    &\hat{P}^{\mathcal{V}}_\text{LLM}(\text{``software engineer''} \mid \text{prompt tokens}) \\
    &=
    \hat{P}^{\mathcal{V}}_\text{LLM}(\text{``software''} \mid \text{prompt tokens})
    \hat{P}^{\mathcal{V}}_\text{LLM}(\text{``engineer''} \mid \text{prompt tokens}, \text{``software''})
    \end{aligned}
    \label{eq:software-engineer-pred-example}
\end{align}

It is worth noting that we cannot guarantee that the model only assigns positive probabilities to valid job titles.
In fact, given the presence of the softmax function in our language model, $\hat{P}^\mathcal{V}_\text{LLM}(\cdot \mid \texttemplate(x_{i, \leq t}, y_{i, < t}))$ is strictly positive for any sequence of tokens of any length.
Therefore, the sum of all possible job titles' probabilities is not necessarily one.
We would need the following normalization to calculate the probability of predicting $y_t$ so that predicted probabilities on all job titles sum to one.
\begin{align}
    \hat{P}_\text{LLM}^\text{normalized} (y_{i,t} \mid x_{i, \leq t}, y_{i, < t}) = \frac{\hat{P}^\mathcal{V}_\text{LLM}(\tokenize(\jobtitle(y)) \mid \texttemplate(x_{i, \leq t}, y_{i, < t}))}{\sum_{y' \in \mathcal{Y}} \hat{P}^\mathcal{V}_\text{LLM}(\tokenize(\jobtitle(y') \mid \texttemplate(x_{i, \leq t}, y_{i, < t}))}
    \label{eq:predication-normalization}
\end{align}

The normalization operation in Equation (\ref{eq:predication-normalization}) is computationally expensive, since we need to perform LLM inference $|\mathcal{Y}|$ times. In this paper, we do not perform this normalization and we use the predicted probability from Equation (\ref{eq:predication}) directly. It is worth noting that since the denominator in Equation (\ref{eq:predication-normalization}) is less than one (since the total probability mass on the subset of job title tokens is less than the total probability mass on all tokens), $\hat{P}^\mathcal{V}_\text{LLM} \leq \hat{P}_\text{LLM}^\text{normalized}$. As a result, test perplexity for LLMs reported in the paper \emph{under-estimates} the performance of these LLMs.
To assess the extent of this potential issue, we analyze the total probability mass assigned to the set of all possible job titles on a randomly selected 10\% subset of the data; we did not scale to the full dataset due to the computational cost of computing the predictive probability of all occupations.
We compute the sum of predicted probabilities for all valid titles $y \in \mathcal{Y}$ for each observation $(i, t)$ in this subset, then compute the normalization constant $\sum_{y' \in \mathcal{Y}} \hat{P}^\mathcal{V}_\text{LLM}(\tokenize(\jobtitle(y') \mid \texttemplate(x_{i, \leq t}, y_{i, < t}))$. The average values of normalization constants in all three datasets are close to 1, with detailed values in Table \ref{tab:normalization-effect}. This indicates that the fine-tuned LLM concentrates nearly all of its probability mass (> 99\%) on valid job title tokens.
Furthermore, we compare the perplexity computed using un-normalized probabilities (i.e., $\hat{P}^\mathcal{V}_\text{LLM}(\tokenize(\jobtitle(y)) \mid \texttemplate(x_{i, \leq t}, y_{i, < t}))$) against the perplexity using normalized probabilties (i.e., $\hat{P}_\text{LLM}^\text{normalized} (y_{i,t} \mid x_{i, \leq t}, y_{i, < t})$) on this 10\% subset.
Table \ref{tab:normalization-effect} summarizes the perplexity computed from unnormalized and normalized predictive probabilities. The results suggest that the lack of explicit normalization leads to a negligible underestimation of model performance. Note that these perplexity values are calculated on a subset of the test data and is different from the main results reported using the full test set.

\begin{table}[!htbp]
    \centering
    \caption{Test-Set Perplexity Comparison: Un-normalized vs. Normalized Predictive Probabilities.}
    \label{tab:normalization-effect}
    \begin{tabular}{lccc}
    \toprule
    \textbf{Evaluation Dataset} & PSID81 & NLSY79 & NLSY97 \\
    \midrule
    Average of $\sum_{y' \in \mathcal{Y}} \hat{P}^\mathcal{V}_\text{LLM}(\tokenize(\jobtitle(y') \mid \texttemplate(x_{i, \leq t}, y_{i, < t}))$ & 0.999035 & 0.999068 & 0.998893 \\
    Perplexity from $\hat{P}^\mathcal{V}_\text{LLM}(\tokenize(\jobtitle(y)) \mid \texttemplate(x_{i, \leq t}, y_{i, < t}))$ & 7.833197 & 8.170692 & 6.227802 \\
    Perplexity from $\hat{P}_\text{LLM}^\text{normalized} (y_{i,t} \mid x_{i, \leq t}, y_{i, < t})$ & 7.824903 & 8.162442 & 6.220557 \\
    Underestimation of LABOR-LLM Performance & 0.008294 & 0.008250 & 0.007244 \\
    \bottomrule
    \end{tabular}
    \legend{Perplexity values are calculated on a 10\% subset of the test data.}
\end{table}

\section{Details on Embedding-Based Approach}
\label{sec:appendix-embedding-based-approach}

This appendix provides the details of the embedding-based approach reported on in Section \ref{sec:lm-as-embedding-engine}.
To extract embeddings from the Llama models (fine-tuned and off-the-shelf), we use the final-layer model embedding at the last token of each sequence of tokens.
For OpenAI embeddings, we used the latest \texttt{text-embedding-3-large} model at the time the analysis was conducted (November $12^\text{th}$, 2024), which was released in January, 2024, and was still the latest as of December, 2025; details are available at \url{https://platform.openai.com/docs/guides/embeddings}.

We estimate the multinomial logistic regression using Bayesian Optimization to find the optimal learning rate in the log-uniform space $[10^{-6},10^{-2}]$. The embeddings are high-dimensional with thousands of dimensions. We also explore using embeddings of 16, 64, or 256 dimensions, using PCA to reduce our embeddings, in addition to the full-dimensional embeddings, and pick the best-performing model from our validation set.\footnote{We explore random forest with 50 Bayesian Optimization calls and uniform parameters $[20,400]$ estimators, $[5,50]$ maximum depth, $[0.01,0.9]$ minimum samples split, $[0.01,0.9]$ minimum samples leaf. Performance is significantly worse than that of multinomial logistic regression.}

Note that the embedding-based model cannot predict occupations that are not in the training set; therefore, we drop transitions of occupations that are present in the test set, but not the training set in Table \ref{tab:perplexity-embedding-approach}. The train/test split that we use to report results in this paper has 13 transitions in the test set for PSID81 and two for NLSY97 that are dropped due to having occupation codes that are not in the training set. These few observations have a negligible impact on our perplexity metric, as it is inherently robust to individual data points. The language-based model addresses this issue by producing predictive probabilities that are inherently valid for all job titles, including those not represented in the training set. In later tables, there will be 13 more transitions in the PSID81 and two more transitions in NLSY97.

\section{Details on Fine-Tuning}
\label{sec:appendix-fine-tune-details}
This section discusses the details of fine-tuning LLMs in this paper and additional results showing how the number of epochs, i.e., complete passes through the entire training dataset during the training process, impacts model performance.

For each individual $i$ in the training split, we construct a text representation of her complete career history $\texttemplate(x_{i, \leq T_i}, y_{i, \leq T_i})$ as described in Section \ref{sec:text-template}.
We use these text representations as the corpus to fine-tune the language models. In the fine-tuning process, there is one observation for each token in each $\texttemplate(x_{i, \leq T_i}, y_{i, \leq T_i})$ in the training corpus. The loss function not only considers the model's prediction on tokens corresponding to job titles, but also on tokens corresponding to everything else in the text representation to improve models' understanding of our text template data structure.
We use $\texttemplate(x_{i, \leq T_i}, y_{i, \leq T_i})$ from individuals in the validation split to evaluate the performance of the fine-tuned models after each fine-tuning epoch.

For each model reported in the paper, we deploy two different training strategies: full-parameter automated mixed precision fine-tuning for three epochs (where in the context of fine-tuning, an epoch is a single complete pass through a dataset) and the same for five epochs. During the fine-tuning, we evaluate the model's validation loss after each training epoch, and keep the model checkpoint (saved snapshot of a model's parameters) that attains the lowest validation loss for evaluation.

All models in this paper were fine-tuned using the two strategies mentioned, and we always report the model from the better-performing strategy.

Consider now some additional details about fine-tuning, which mirrors the pre-training process. First, note that our description of CAREER in Section \ref{sec:transformers} gives a high-level overview of the functional form of a transformer model, where the ``vocabulary'' of CAREER is occupations instead of tokens from English words. Now consider estimation details. In current practice, LLMs are usually trained so that the parameters of the transformer neural network maximize log-likelihood, which in the case of language models, where outcomes are encoded as indicator variables for tokens, is equivalent to minimizing cross-entropy loss (\citet{touvron_llama_2023}). In stochastic gradient descent, observations are grouped into small batches. Given parameter estimates from prior batches, within each new batch, the gradient of the loss with respect to the parameters is evaluated for each observation in the batch (where the gradient is evaluated at the previous parameter estimates). The parameters are then updated using an adaptive version of stochastic gradient descent where updates are made using moving averages; see \citet{touvron_llama_2023} for details.

In our fine-tuning, we use a batch size of 32, the initial learning rate of $10^{-5}$ (which determines the step size for each update of model parameters), and a linear learning rate decay (which determines how the learning rate changes across epochs, see e.g., \citet{jin2023rethinking}) from the initial learning rate to zero learning rate. Such learning rate scheduling of linear decaying is enforced by Together AI, and we do not have control over it at the time of fine-tuning. It is worth noting that given the linear learning rate decay, the checkpoints corresponding to the first three epochs in the three epoch settings are different from the first three epochs in the five epoch settings.

We also experiment with fine-tuning the model for more epochs while taking the checkpoint corresponding to the lowest validation loss. We observed escalating validation loss (i.e., over-fitting) after four to five epochs. Due to the prohibitive computational cost, we only fine-tuned Llama-2-7B models using the pooled training data for five (reported in the main paper), six, eight, and ten epochs. Table \ref{tab:fine-tune-more-epochs} summarizes the perplexities of the best model checkpoint, according to the validation loss, in these settings. We do not observe significant improvement in model performance, if any, while fine-tuning the model for more epochs.

\begin{table}[hbtp]
    \centering
    \caption{Test-set perplexity of FT-7B fine-tuned for 5, 6, 8, or 10 epochs.\updated{12/05/2025, no change from 11/25/2024}}
    \label{tab:fine-tune-more-epochs}
    \begin{tabular}{llll}
    \toprule
    \multicolumn{1}{r}{\textbf{Evaluation Dataset}} & PSID81 & NLSY79 & NLSY97 \\
    \multicolumn{1}{r}{\textbf{Number of Test Set Transitions}} & 61,772 & 51,593 & 29,951 \\
    \midrule
    FT-7B Best Checkpoint of 5 Epochs & 8.083 (0.124) & 8.208 (0.146) & 6.194 (0.097) \\
    FT-7B Best Checkpoint of 6 Epochs & 8.122 (0.125) & 8.223 (0.147) & 6.214 (0.096) \\
    FT-7B Best Checkpoint of 8 Epochs & 8.098 (0.124) & 8.190 (0.145) & 6.191 (0.098) \\
    FT-7B Best Checkpoint of 10 Epochs & 8.143 (0.124) & 8.240 (0.147) & 6.217 (0.098) \\
    \bottomrule
    \end{tabular}
    \legend{FT-7B model is trained on the union of the three survey datasets. Test-set-bootstrap standard errors are in parentheses.}
\end{table}

\section{Additional Results on Gap Year Prediction}
\label{sec:appendix-gap-year}
This section provides additional analyses of FT-LABOR-LLM's prediction behavior when there is a gap between the calendar years of the current occupation and the previous occupation in a transition.
Let $\text{year}_{i, t}$ denote the calendar year of the $t^{\text{th}}$ transition of individual $i$, where $\text{year}_{i, t-1}$ is the calendar year of the previous transition (only defined for $t > 1$). Specifically, we focus on transitions with $t > 1$ such that $\text{year}_{i, t} = \text{year}_{i, t-1} + 2$, i.e., the gap size is exactly one calendar year, to reduce computational resource requirements. To create the dataset, we randomly sample 500 transitions from the test split of each survey dataset, resulting in a total of 1,500 transitions.

Using the FT-LABOR-LLM model fine-tuned on the mixture training data, we compute the predicted probability of landing at occupation $y_{i, t}$ in calendar year $\text{year}_{i, t}$ as:
$$
\hat{P}(y_{i, t} \text{ in } \text{year}_{i, t} \mid y_{i, t-1} \text{ in } \text{year}_{i, t}-2),
$$
where covariates $x_{i, \leq t}$ and past occupations $y_{i, < t-1}$ are omitted in the conditional part for simplicity. This is referred to as the \textbf{direct prediction}.
We also compute the \textbf{compound prediction} as:
\[
\sum_{y' \in \mathcal{Y}} \hat{P}(y_{i, t} \text{ in } \text{year}_{i, t} \mid y' \text{ in } \text{year}_{i, t}-1 \land y_{i, t-1} \text{ in } \text{year}_{i, t}-2) \times \hat{P}(y' \text{ in } \text{year}_{i, t}-1 \mid y_{i, t-1} \text{ in } \text{year}_{i, t}-2).
\]
Computing the compound prediction for a single transition requires approximately $2 \times |\mathcal{Y}| \approx 700$ model inferences, making this experiment computationally expensive.

Finally, we compare the agreement between the direct prediction and the compound prediction using the 1,500 transitions; the log probabilities are found to be highly correlated, with a correlation coefficient of $0.93$.

\section{Additional Results for the Value of Information}
\label{sec:appendix-VOI}
In this section, we report on a complementary exercise to that conducted in Table \ref{tab:cross-dataset-eval} of the main paper. Instead of either fine-tuning on a single dataset or the union of all datasets, we start from each baseline survey training dataset and create new training datasets that mix in additional data from the other two surveys. Specifically, we take the training split of dataset $\omega$, $\mathcal{D}_\omega^{\text{(train)}}$ and mix it with $P\% \times |\mathcal{D}_\omega^{\text{(train)}}|$ additional training samples from training splits of the other two datasets $\mathcal{D}_{\omega'}^{\text{(train)}} \cup \mathcal{D}_{\omega''}^{\text{(train)}}$. We fine-tune Llama-2-7B models using the merged training data, and then evaluate the model's performance on the test split $\mathcal{D}_{\omega}^{\text{(test)}}$.

Table \ref{tab:value-of-info-extra-training-data} summarizes the performance of these models fine-tuned with additional training data; adding sufficient non-representative data leads to improvements over the models fine-tuned with only data representative of the test set.

\begin{table}[!htbp]
    \centering
    \caption{
    Test-set perplexity of fine-tuning model on full training split plus $P\%$ training data from other sources.\updated{12/05/2025, no change from 12/02/2024}}
    \label{tab:value-of-info-extra-training-data}
    \begin{tabular}{llll}
    \toprule
    \multicolumn{1}{r}{\textbf{Evaluation Dataset}} & PSID81 & NLSY79 & NLSY97 \\
\multicolumn{1}{r}{\textbf{Number of Test Set Transitions}} &  61,772 & 51,593 & 29,951 \\
    \midrule
    \textbf{Perplexity} & & & \\
    \midrule
    FT-7B with $P=0$ & 8.184 (0.126) & 8.329 (0.147) & 6.350 (0.101) \\
    FT-13B with $P=0$ &  8.140 (0.126) & 8.282 (0.145) & 6.326 (0.100) \\
    \midrule
    FT-7B with $P=10$ & 8.179 (0.127) & 8.315 (0.147) & 6.333 (0.099) \\
    FT-7B with $P=30$ & 8.109 (0.124) & 8.293 (0.147) & 6.293 (0.099) \\
    FT-7B with $P=50$ & 8.094 (0.123) & 8.281 (0.148) & 6.278 (0.098) \\
    FT-7B with $P=70$ & 8.087 (0.124) & 8.266 (0.146) & 6.260 (0.099) \\
    \midrule
    \textbf{Perplexity Improvement} & & & \\
    \midrule
    PPL(FT-13B)-PPL(FT-7B with $P=10$) & -0.039 (0.014) & -0.033 (0.013) & -0.007 (0.010) \\
    PPL(FT-13B)-PPL(FT-7B with $P=30$) & 0.030 (0.014) & -0.011 (0.012) & 0.033 (0.010) \\
    PPL(FT-13B)-PPL(FT-7B with $P=50$) & 0.045 (0.013) & 0.001 (0.013) & 0.047 (0.010) \\
    PPL(FT-13B)-PPL(FT-7B with $P=70$) & 0.053 (0.014) & 0.017 (0.013) & 0.065 (0.010) \\
    \bottomrule
    \end{tabular}
    \legend{Test-set-bootstrap standard errors are in parentheses.}
\end{table}

\section{Details on Sensitivity to Input Features}
\label{sec:appendix-sensitivity-to-input-features}

In this section, we evaluate the importance of demographic variables for predictive performance. This exercise is not straightforward for complex, nonlinear models. If we find that including a covariate in the estimation of a model improves predictive quality on a test set, that implies that the covariate both matters in the true (unknown) data generating process, and that the predictive model makes use of the covariate in prediction. However, if excluding a covariate does not affect predictive quality, we cannot be sure whether something in the estimation process failed to capture a relationship that is present in the true data generating process (e.g., mis-specification or noise), or whether the covariate is simply not important once other covariates are incorporated. Although it is straightforward to assess whether an individual covariate has predictive power in isolation using very simple models, understanding whether it has predictive power conditional on other covariates relies on modeling. Thus, negative results about the importance of a covariate require additional analysis to confirm whether, in fact, that covariate has predictive power. Here, we do not explore the latter question.

We evaluate the importance of demographic variables for FT-7B fine-tuned on $\mathcal{D}_\text{all}^\text{(train)}$, our best-performing predictive model.  We apply an approach common in the machine learning literature, which entails holding fixed the estimated model, and replacing covariates with randomly assigned values in the test set, then assessing the impact on predictive performance of the model when the model is applied to the modified test set. \footnote{Note that this exercise is imperfect; an alternative would be to re-estimate the model omitting a covariate, since a model might increase the loadings on correlated covariates when a particular covariate is omitted. However, re-estimating the model comes with computational cost. Thus, we focus here on exercises that can be carried out without re-estimation.}

We explore the importance of three static variables in our text representations: gender, ethnicity, and indicators for four regions of the country. To implement the randomization of the test set demographics, we create an alternative version of the test set in which, for each unit, we replace the vector of demographics with a randomly drawn vector of demographics from units in the validation set and assign the unit those demographics. We repeat this exercise with alternative combinations of variables.

Table \ref{tab:randomized-variables} presents the results. Randomly modifying gender hurts the performance of FT-LABOR-LLM significantly. For PSID81, randomizing gender labels increases perplexity by 1 (about 12\% above baseline), while ethnicity has about a quarter of the effect. For NLSY79 and NLSY97 test sets, gender has a similar impact, but ethnicity has a much lower effect. For PSID81, there is substantial additional degradation in performance from the interaction of gender and ethnicity, while NLSY79 sees ethnicity and region having larger effects when randomized jointly rather than individually. For all three survey datasets, the three-way interaction of gender, ethnicity, and region results in the largest impact, with the incremental effect of including all three covariates over two of them is substantial for PSID81 and NLSY79. These findings should be interpreted in light of the historical trends in the labor market participation relevant to the time periods covered by the different survey. Overall, these results suggest that complex interactions are important to consider when building predictive models of occupation, suggesting that simple, additive regressions of the type commonly used in labor market applications may omit important predictors.

\begin{table}[t]
    \centering
    \caption{Test-set perplexity and perplexity improvement on actual vs. randomized demographic characteristics. \updated{12/05/2025, change to 3 decimal places}}
    \label{tab:randomized-variables}
    \resizebox{\textwidth}{!}{
    \begin{tabular}{lccc}
    \toprule
    \multicolumn{1}{r}{\textbf{Evaluation Dataset}} & PSID81 & NLSY79 & NLSY97 \\
    \multicolumn{1}{r}{\textbf{Number of Test Set Transitions}} & 61,772 & 51,593 & 29,951 \\
    \midrule
    \textbf{Perplexity}\\
    \midrule
    No modification / Actual & 8.184 (0.126) & 8.330 (0.147) & 6.349 (0.101) \\
    \midrule
    Randomized ethnicity & 8.450 (0.130) & 8.400 (0.148) & 6.385 (0.100) \\
    Randomized gender & 9.223 (0.151) & 9.179 (0.167) & 6.896 (0.117) \\
    Randomized region & 8.202 (0.126) & 8.378 (0.148) & 6.356 (0.101) \\
    \midrule
    Randomized gender and ethnicity & 9.373 (0.152) & 9.226 (0.167) & 6.936 (0.117) \\
    Randomized gender and region & 9.295 (0.152) & 9.280 (0.169) & 6.901 (0.117) \\
    Randomized ethnicity and region & 8.434 (0.129) & 8.437 (0.149) & 6.388 (0.100) \\
    \midrule
    Randomized all variables & 9.438 (0.153) & 9.329 (0.170) & 6.935 (0.117) \\
    \midrule
    \textbf{Perplexity Improvement} & & & \\
    \midrule
    PPL(Randomized ethnicity)-PPL(Actual) & 0.267 (0.012) & 0.070 (0.007) & 0.036 (0.006) \\
    PPL(Randomized gender)-PPL(Actual) & 1.039 (0.033) & 0.849 (0.032) & 0.547 (0.027) \\
    PPL(Randomized region)-PPL(Actual) & 0.018 (0.004) & 0.048 (0.005) & 0.007 (0.002) \\
    \midrule
    PPL(Randomized gender and ethnicity)-PPL(Actual) & 1.190 (0.034) & 0.896 (0.034) & 0.587 (0.027) \\
    PPL(Randomized gender and region)-PPL(Actual) & 1.111 (0.034) & 0.950 (0.036) & 0.551 (0.027) \\
    PPL(Randomized ethnicity and region)-PPL(Actual) & 0.251 (0.011) & 0.107 (0.008) & 0.039 (0.006) \\
    \midrule
    PPL(Randomize all)-PPL(Actual) & 1.254 (0.036) & 0.999 (0.037) & 0.585 (0.027) \\
    \bottomrule
    \end{tabular}
    }
    \legend{
    The foundation model is FT-7B fine-tuned on the union of the training sets of the surveys without any modification of demographic features. Test-set-bootstrap standard errors are in parentheses.}
\end{table}

\section{Details on the Value of Longer Career Histories}
\label{sec:appendix-value-of-career-history}

In this section, we assess the predictive value of observing a worker's full history as recorded in the survey, relative to truncating the history to include only more recent observations. This question helps shed light on the sources of model performance with respect to the ability of the transformer model to capture relevant information from long histories; it also informs survey design, since following individuals over long time periods is expensive.

We proceed by evaluating how the predictive quality of FT-7B fine-tuned on $\mathcal{D}_\text{all}^\text{(train)}$, our best-performing predictive model, changes when we make predictions about $y_{i, t}$ using time-invariant covariates, $x_i$, time-varying covariates and occupations reported in the $k$ \emph{most recent} observations of $\{x_{i, \tau}\}_{\tau=t-k}^t$, $\{y_{i, \tau}\}_{\tau=t-k}^{t-1}$,
reporting $$\hat{P}_\text{LLM}\left(y_{i, t} \mid x_i, \{x_{i, \tau}\}_{\tau=t-k}^t, \{y_{i, \tau}\}_{\tau=t-k}^{t-1}\right).$$ With $k = t-1$, the model has access to all available history. We first create different subsets of individual-year observations from the \emph{test set} of each dataset, defining the following non-overlapping subsets of individual-year observations $S^{\text{(test)}}_{t_\text{min} < t \leq t_\text{min} + 5} = \{(i, t) \in \mathcal{D}^{\text{(test)}} \mid t_\text{min} < t \leq \text{min} + 5\}$ for $t_\text{min} \in \{5, 10, 15, 20 ,25\}$.
The NLSY97 dataset covers a shorter time span, therefore, $S^{\text{(test)}}_{20 < t \leq 25}$ and $S^{\text{(test)}}_{25 < t \leq 30}$ are defined as empty sets for NLSY97.
Given a $S^{\text{(test)}}_{t_\text{min} < t \leq \text{min} + 5}$, for each observation $(i, t) \in S^{\text{(test)}}_{t_\text{min} < t \leq \text{min} + 5}$, we create text templates consisting of only the $k$ \emph{most recent} observations of individual $i$ prior to her $t^\text{th}$ observation: $\texttemplate(x_i, \{x_{i, \tau}\}_{\tau=t-k}^t, \{y_{i, \tau}\}_{\tau=t-k}^{t-1})$. For values of $k$, we consider multiples of five such that $k \leq t_\text{min}$ (e.g., $k \in \{5, 10, 15, 20\}$ if $t_\text{min} = 20$).
A greater value of $k$ exposes the model to more information about the individual's career history and should lead to an improved prediction accuracy.

We then assess perplexity in the test set for different subsets of constructed test data defined by values of $(k,t_\text{min})$:
\begin{align*}
    \tilde{S}^{\text{(test)}}_{t_\text{min} < t \leq t_\text{min}+5, k} = \left\{\left(\texttemplate(x_i, \{x_{i, \tau}\}_{\tau=t-k}^t, \{y_{i, \tau}\}_{\tau=t-k}^{t-1}), y_{i, t}\right)
    \right\}_{(i, t) \in S^{\text{(test)}}_{t_\text{min} < t \leq t_\text{min} + 5}}
\end{align*}
where each element of $\tilde{S}^{\text{(test)}}_{t_\text{min} < t \leq t_\text{min}+5, k}$ is a pair of (1) a prompt containing $k$ past observations prior to the $t^\text{th}$ record of individual $i$ and (2) the ground truth occupation individual $i$ has in her $t^\text{th}$ record.

We evaluate the FT-LABOR-LLM models using the prompt-label pair in \emph{each} $\tilde{S}^{\text{(test)}}_{t_\text{min} < t \leq t_\text{min}+5, k}$ \emph{separately}. Within each $\tilde{S}$ group, we query the likelihood that the language model assigns to the ground truth job title as the continuation of the text prompt, $\hat{P}_\text{LLM}(\jobtitle (y_{i, t}) \mid \texttemplate(x_i, \{x_{i, \tau}\}_{\tau=t-k}^t, \{y_{i, \tau}\}_{\tau=t-k}^{t-1}))$, and compute the perplexity using all predictions within that $\tilde{S}$.

For example, consider the following prompt that would be fed into the LLM to predict the fifth occupation using the first four observations.
{
\footnotesize
\begin{Verbatim}[breaklines=true]
<A worker from the PSID dataset>
The following information is available about the work history of a female white US worker residing in the west region.
The worker was born in 1985.
The worker has the following records of work experience, one entry per line, including year, education level, and the job title:
2007 (college): Postmasters and mail superintendents
2009 (college): Athletes, coaches, umpires, and related workers
2011 (college): Education administrators
2013 (college): Child care workers
2015 (college):
\end{Verbatim}
}

If we set $k=2$ most recent previous observations, we would drop the first two observations in the years 2007 and 2009 and feed the following prompt into the LLM to predict the fifth occupation using only the two most recent observations instead of the full prompt above.

{
\footnotesize
\begin{Verbatim}[breaklines=true]
<A worker from the PSID dataset>
The following information is available about the work history of a female white US worker residing in the west region.
The worker was born in 1985.
The worker has the following records of work experience, one entry per line, including year, education level, and the job title:
2011 (college): Education administrators
2013 (college): Child care workers
2015 (college):
\end{Verbatim}
}

Finally, we build a matrix of perplexity metrics assessing model's performance under different levels of exposure to past information.
Table \ref{tab:limit-context-window} summarizes model performance when it only has access to a limited number of past observations while predicting the next occupation.
To better illustrate the result, we compute the perplexity difference between predictions made using prompts with $k \in \{10, 15, 20, 25\}$ and the baseline predictions made using prompts with $k=5$. For example, for PSID81, the data in row $t\in(15,20]$ and column $k=10$ indicates that predictions made on those observations using $k=10$ past observations in prompts for transitions indexed between 15 and 20  achieve a perplexity that is 0.193 (with a test-set-bootstrap standard error of 0.026) lower than the perplexity of predictions using $k=5$ past observations. Truncating the career history thus leads to a significant decrease in predictive performance, although for transitions at the end of a worker's career, most of the predictive benefit is achieved with 10 or 15 years of history.

\begin{table}[t]
    \centering
    \caption{Test-set perplexity improvement from increasing number of historical periods used for prediction. \updated{12/05/2025 replication check}}
    \label{tab:limit-context-window}
    \begin{tabular}{cccccc}
    \toprule
    \textbf{PSID81} & \textbf{Number of Test Set Transitions} & $k=10$ & 15 & 20 & 25 \\
    \midrule
    $t \in (10, 15]$ & 9,180 & 0.182 (0.020) & - & - & - \\
    $t \in (15, 20]$ & 5,288 & \textcolor{red}{0.193 (0.026)} & 0.237 (0.032) & - & - \\
    $t \in (20, 25]$ & 3,214 & 0.075 (0.015) & 0.086 (0.018) & 0.099 (0.019) & - \\
    $t \in (25, 30]$ & 1,008 & 0.050 (0.015) & 0.046 (0.017) & 0.054 (0.019) & 0.049 (0.019) \\
    \midrule
    \textbf{NLSY79} & \textbf{Number of Test Set Transitions} & $k=10$ & 15 & 20 & 25 \\
    \midrule
    $t \in (10, 15]$ & 9,078 & 0.306 (0.035) & - & - & - \\
    $t \in (15, 20]$ & 8,051 & 0.370 (0.035) & 0.442 (0.042) & - & - \\
    $t \in (20, 25]$ & 6,719 & 0.131 (0.017) & 0.166 (0.018) & 0.184 (0.020) & - \\
    $t \in (25, 30]$ &2,617 & 0.079 (0.026) & 0.116 (0.028) & 0.132 (0.029) & 0.132 (0.028) \\
    \midrule
    \textbf{NLSY97} & \textbf{Number of Test Set Transitions} & $k=10$ & 15 & & \\
    \midrule
    $t \in (10, 15]$ & 7,151 & 0.187 (0.031) & - & - & - \\
    $t \in (15, 20]$ & 4,112 & 0.116 (0.030) & 0.179 (0.039) &  - & -\\
    \bottomrule
    \end{tabular}
    \legend{Each row corresponds to a group of individual-year observations $S^{\text{(test)}}_{t_\text{min} < t \leq \text{min} + 5}$, each column corresponds to a value of $k$, and each cell corresponds to the perplexity improvement due to increasing the number of past observations from 5 to $k$. Test-set-bootstrap standard errors are in parentheses.}
\end{table}

\section{Details for Additional Analyses}
\label{sec:appendix-GRF}

In this appendix, we provide additional details on the two exercises we perform in Section \ref{sec:additiona-analyses}. First, we report the details of our analysis to learn the extent to which the embeddings created by FT-7B incorporate information about the meaning of job titles by assessing the predictive power of these embeddings on a task related to the interpretation of job titles.

Specifically, we use different transformer models to generate embedding vectors for all occupations $y \in \mathcal{Y}$, and set up a prediction task to explore how much SOC occupational hierarchy these embeddings encode.
Since the Llama-2-7B embedding dimensions are much higher (e.g., 4,096) than CAREER's native embedding dimension (i.e., 192), we apply PCA dimension reduction to reduce all embeddings to 192 dimensions for fair comparison.
Then, we build a multinomial logistic regression (with an elastic-net regularization) to predict which of the following six SOC groups an occupation belongs to: ``Alternate aggregations'', ``Management, Business, Science, and Arts Occupations'',  ``Service Occupations'', ``Sales and Office Occupations'', ``Natural Resources, Construction, and Maintenance Occupations Production, Transportation, and Material Moving Occupations'', and ``Military Specific Occupations''.
We regularize the multinomial regression using a convex combination of L1 and L2 regularization (i.e., the elastic-net, $\frac{\alpha||\mathbf{\beta}||_1 + (1-\alpha) ||\mathbf{\beta}||_2}{C}$); and we use five-fold cross-validation to choose the best regularization strength $C$ and weight $\alpha$.

Table \ref{tab:embeddings-literal-numeric} shows that LLM embeddings can capture meaningful patterns in occupational hierarchies, highlighting the importance of prior knowledge in the predictions.

\begin{table}[hbtp]
\centering
\caption{Test-set accuracy of predicting correct SOC-group given embeddings. \updated{12/05/2025}}
\label{tab:embeddings-literal-numeric}
\begin{tabular}{llrrl}
\toprule
\textbf{Embedding Method} & \textbf{Test Set Accuracy }\\
\midrule
FT-7B & 81.52\% (0.042) \\
OTS Llama-2-7B & 80.91\% (0.028) \\
CAREER & 78.18\% (0.068) \\
Gaussian Random in $\mathbb{R}^{192}$ & 32.42\% (0.030) \\
OTS Llama-2-7B (Numeric Title) & 32.42\% (0.030) \\
\bottomrule
\end{tabular}
\legend{Test-set-bootstrap standard errors are in parentheses. All models are PCA-ed to 192 dimensions.}
\end{table}

Second, we report the details of our analysis to learn for which types of transitions FT-13B outperforms CAREER in predicting whether an individual ``moves'' occupations.

Specifically, we ask the question: for what kind of mover observations $(i, t)$ with characteristics $(y_{i,t}, x_{i, \leq t}, y_{i, <t})$ do language models outperform the previous specialized transformer? We focus on ``mover'' transitions in the test split of the PSID81 dataset since it is our largest dataset.

To begin, we define our prediction target as the difference in the log-likelihood of the ground truth between predictions from FT-13B and CAREER, as follows:
\begin{align}
   \begin{aligned}
   \Delta \hat{P}_\text{job} =& \log \hat{P}_\text{LLM}(y_{i,t} \mid y_{i,t}\neq y_{i-1,t}, x_{i, \leq t}, y_{i, < t})\\
   &- \log \hat{P}_\text{CAREER}(y_{i,t} \mid y_{i,t}\neq y_{i-1,t}, x_{i, \leq t}, y_{i, < t})
   \end{aligned}
   \label{eq:treatment-effect}
\end{align}
where $\Delta \hat{P}_\text{job}$ quantifies the improvement of FT-13B over the CAREER model for a particular transition $(i, t)$ (i.e., individual-year observation).

We build a predictive generalized random forest  (which embeds sample splitting to avoid overfitting as described in \citet{athey_generalized_2018}) to predict this difference using as covariates the variables in Table \ref{tab:grf-features}. We assign each realization of covariates to a quintile based on the resulting estimates of the difference between the models (i.e., $\Delta \hat{P}_\text{job}$).
The presence of heterogeneity in the quintile-level test set mean differences in log-likelihood indicates that the intensity of differences in performance between FT-13B and CAREER vary as a function of the features of the individual-year observation, denoted $\Phi_{i,t}(y_{i,t}, x_{i, \leq t}, y_{i, <t})$. Note that logged variables are computed as $\log(x+1)$ to avoid $\log(0)$.

\begin{table}[h!]
\centering
\caption{Description of features used in the heterogeneous advantage analysis.}
\label{tab:grf-features}
\resizebox{\textwidth}{!}{
\begin{tabular}{@{}lp{12cm}@{}}
\toprule
\textbf{Feature} & \textbf{Description} \\ \midrule
\textbf{Transition index} & The transition index $t$ of the occupation $y_{i,t}$, which is the number of prior observations in the dataset. With a higher $t$, the models have access to a longer career history while making the prediction. \\
\textbf{(Logged) occupation frequency} & The number of occurrences of occupation $y_{i,t}$ in the dataset. \\
\textbf{(Logged) previous occupation frequency} & The number of occurrences of occupation $y_{i, t-1}$ in the dataset. \\
\textbf{(Logged) empirical transition frequency} & The empirical number of transitions $y_{i, t-1} \to y_{i,t}$, calculated as $\#^\text{(train)}\{y_{i, t-1} \to y_{i,t}\}$. \\
\textbf{(Logged) empirical transition probability} & The empirical probability of transition $y_{i, t-1} \to y_{i,t}$, calculated as $\frac{\#^\text{(train)}\{y_{i, t-1} \to y_{i,t}\}}{\#^\text{(train)}\{y_{i, t-1}\}}$. \\
\textbf{(Logged) number of tokens in job title} & The number of tokens in the job title of occupation $y_{i, t}$. \\
\textbf{(Logged) number of tokens in previous job title} & The number of tokens in the previous job title $y_{i, t-1}$. \\
\textbf{Same SOC group} & Using the SOC hierarchy to cluster $y_{i, t-1}$ and $y_{i,t}$ into $\text{SOC-group}(y_{i, t-1})$ and $\text{SOC-group}(y_{i,t})$. Indicators measure the magnitude of occupation transition: $\mathbf{1}\{\text{SOC-group}(y_{i, t-1}) = \text{SOC-group}(y_{i,t})\}$. \\
\textbf{Same detailed SOC group} & Using the SOC hierarchy to cluster $y_{i, t-1}$ and $y_{i,t}$ into $\text{SOC-detailed-group}(y_{i, t-1})$ and $\text{SOC-detailed-group}(y_{i,t})$. Indicators measure the magnitude of occupation transition: $\mathbf{1}\{\text{SOC-detailed-group}(y_{i, t-1}) = \text{SOC-detailed-group}(y_{i,t})\}$. \\
\textbf{Occupational Similarity based on O*NET} & We compute cosine similarities between occupation $y_{i, t-1}$ and $y_{i,t}$ on eight aspects in the O*NET dataset: ``Abilities'', ``Composite Attributes'', ``Interests'', ``Knowledge'', ``Skills'', ``Work Activities'', ``Work Styles'', and ``Work Values'', separately; then, include the average cosine similarity. \\
\textbf{Similarity between job titles} & Cosine similarity of embeddings for job titles $y_{i, t-1}$ and $y_{i,t}$, generated using the off-the-shelf Llama-2-7B. \\
\textbf{Embedding of career history $\texttemplate(x_{i, \leq t}, y_{i, <t})$} & Embedding of text representation $\texttemplate(x_{i, \leq t}, y_{i, <t})$ generated using the off-the-shelf Llama-2-13B model. The embedding space is reduced from 5,120 to 32 dimensions via PCA for faster GRF estimation.

\\ \bottomrule
\end{tabular}
}
\end{table}

\begin{figure}[t]
\centering
    \caption{Average covariate values within each quintile as defined by the predicted difference in log-likelihood on conditional prediction. \updated{12/04/2025, actual values subject to rounding error.}}
    \label{fig:features-by-CATE-movers}
    \includegraphics[width=\linewidth]{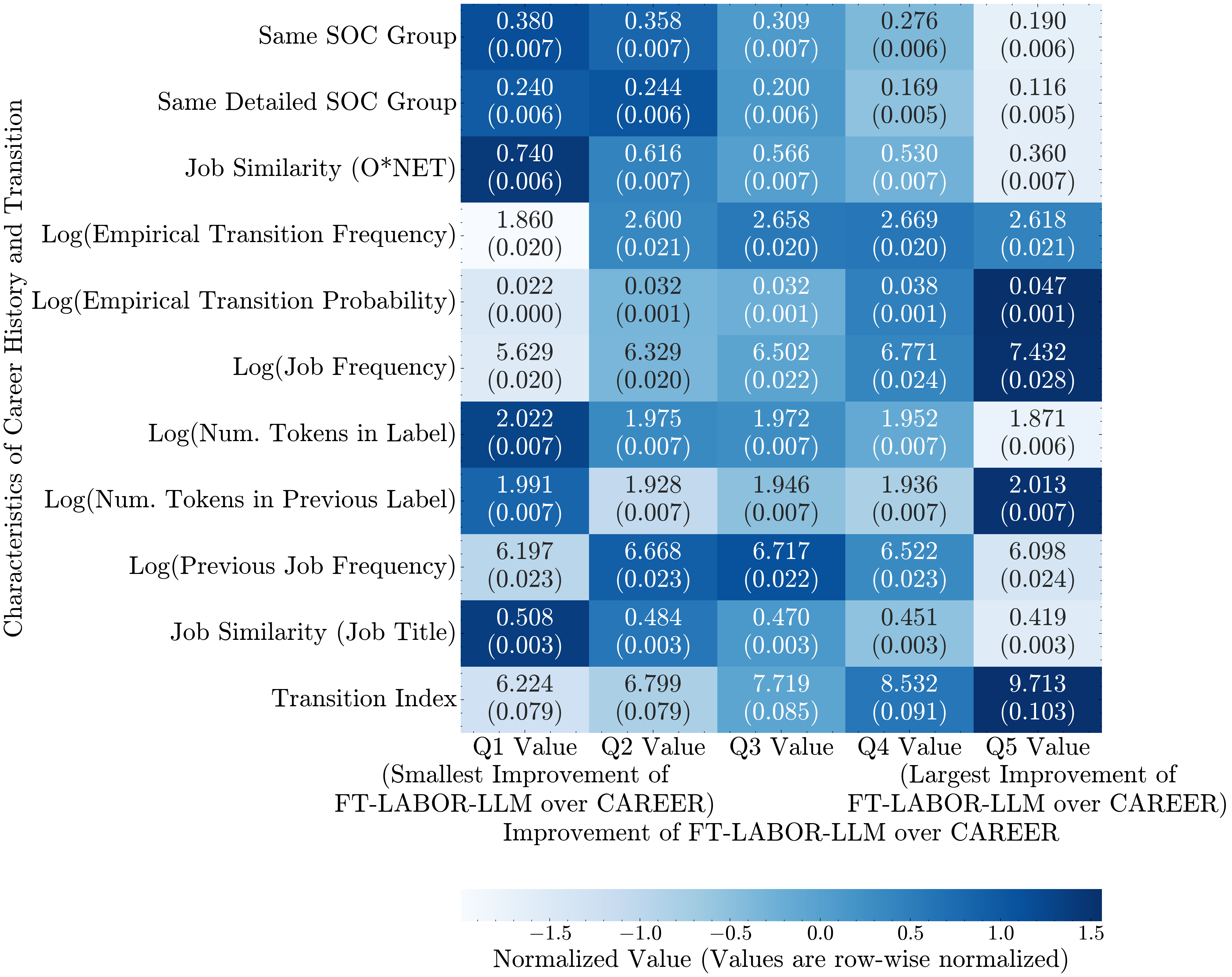}
    \legend{Each cell depicts the corresponding feature's values for each quintile by the estimated difference. Standard errors of feature values are shown in parentheses.}
\end{figure}

Then, we show the values of several features in each quintile, allowing us to understand the factors that vary systematically between higher and lower quintiles.
The corresponding heat map is shown in Figure \ref{fig:features-by-CATE-movers}; for example, Figure \ref{fig:features-by-CATE-movers} shows that fine-tuned Llama-2-13B performs better for movers as the transition index increases and the similarity between $y_{i, t-1}$ and $y_{i, t}$ decreases. This improvement can again be attributed to the attention mechanism and pre-training.

\section{Data Appendix}
\label{sec:appendix-data}

The paper uses three nationally representative survey datasets from the United States to assess the performance of occupation models in predicting career trajectories: the Panel Study of Income Dynamics (PSID81), the National Longitudinal Survey of Youth 1979 (NLSY79), and the National Longitudinal Survey of Youth 1997 (NLSY97). In addition, the paper uses occupational information from O*Net to create an occupation similarity feature in the data. This data appendix details each data source, how it was retrieved, and the data pre-processing steps we took for each dataset. We also provide descriptive statistics on the static variables in this appendix, and describe the process of combining the datasets.

For each survey dataset, we construct a group of static and dynamic variables. Static variables that remain consistent over time are ``personal id,'' ``gender,'' ``birth year,'' ``race/ethnicity,'' and ``region.'' We also construct two dynamic variables for each survey year, ``occupation'' and ``education level,'' that employ two input variables ``education enrollment status'' (for NLSY datasets only) and ``employment status.'' The sections below describe how the listed variables are constructed using each dataset.

\subsection{The Panel Study of Income Dynamics (PSID81)}

The Panel Study of Income Dynamics (PSID) is a longitudinal U.S. household survey tracking families and their individual members (\citet{PSID2024}). The first annual wave from 1968 included approximately 4,800 households. Since then, the PSID has traced all individuals from those households and their descendants, collecting information on individuals and their co-residents on an annual basis through 1997, then biennially starting in 1999. Each member of the original PSID study and their descendants continue to be surveyed, even after leaving the household of origin. This is true for children, other adult members, and ex-spouses forming new family units. The original PSID study was focused on the dynamics of poverty, so the 1968 wave oversampled low-income households and had a relatively large sub-sample of Black respondents. A representative sample of 2,043 Latino households (of Mexican, Cuban and Puerto Rican origin) was added in 1990, but was dropped by the PSID in 1995, so we drop this sample from our final dataset.

To replicate the results in our study, researchers can download the data file that we used from the PSID data center at \href{https://simba.isr.umich.edu/DC/c.aspx}{https://simba.isr.umich.edu/DC/c.aspx}. After creating an account, the researcher can use the ``Previous Cart'' option, search for the email \href{tianyudu@stanford.edu}{tianyudu@stanford.edu}, and select Job ``339649'' The raw data file used for the analysis in this paper was created and downloaded on November $2^\text{nd}$, 2024 at 10:52:52 PM.
If the above dataset cannot be successfully retrieved, our replication notebook also provides a complete list of variables we used and the instruction to obtain these data from the PSID data server at \href{https://simba.isr.umich.edu/DC/l.aspx}{https://simba.isr.umich.edu/DC/l.aspx}.

In this project, we restrict our attention to survey years between 1981 and 2021 (inclusive) because occupation code was originally recorded with only one or two digits in 1979 and 1980, and retrospective updating to three-digit codes was missing for many individuals. We also restrict our sample to individual-year observations that are household heads or spouses because we observe occupation and race/ethnicity information only for these family members. After the pre-processing described below, our resulting final dataset, which we refer to as PSID81, has 31,056 individuals and 313,622 total individual-year observations of occupations.

We use five static covariates for each individual, dropping individuals for whom this information is missing: personal id, gender, race/ethnicity, region, and birth year. We construct each individual's personal id by combining the PSID identifiers for family and individual. We use the main PSID variable for gender, classifying individuals as ``male'' or ``female.'' Race/ethnicity is recorded each survey year by the PSID, with definitions varying slightly from year to year.\footnote{Race/ethnicity for spouse was collected by the PSID starting in 1985.} We collapse all definitions into either ``White'' (consistent category across years), ``Black,'' or ``Other/Unknown.'' Then, we take the first non-Other/Unknown observation of race/ethnicity for our static variable, or classify the individual as Other/Unknown if their race/ethnicity is never classified as White or Black. We base the region variable on the state in which a family lives, which is recorded each survey year by the PSID. First, we construct region as a 4-category variable that takes the values ``northeast,'' ``south,'' ``west,'' and ``northcentral'' based on state. Then, we take the first non-missing observation as our static variable.

We construct birth year based on the age variable recorded each survey year by the PSID. To compute birth year, we take the mode of the difference between the survey year and the individual's age for each individual-year observation. When there is more than one mode, we take the average of the two most frequent birth years. Two modes, which we observe for 1,702 individuals, are likely the result of variation in the timing of a survey within the calendar year. Three and four modes, which we observe for 32 and 3 individuals, respectively, are likely due to measurement error.

We construct two dynamic variables for each individual-year observation in addition to the calendar year of survey: education level and occupation. We construct education level based on the years of education recorded each survey year in the PSID81. We categorize years of education into ``less than high school,'' ``high school,'' ``some college,'' ``college,'' and ``any graduate'' each year, then forward-fill education to replace missing values and impose the restriction that education level be non-decreasing.

We construct our main variable of interest, occupation, using the same pre-processing steps applied by \cite{vafa_career_2024} to facilitate comparisons, combining information from multiple variables recorded each survey year by the PSID81. First, we crosswalk individual-year observations of occupation that are recorded as either 1970 or 2000 census codes to the occ1990dd scheme for uniformity throughout the dataset (\citet{occ1990dd}).

We then collapse the employment status variable into four categories: ``employed,'' ``out of labor force'' (defined as ``Retired,'' ``Permanently disabled,'' or ``Housewife''), ``unemployed'' (defined as ``Only temporarily laid off'' or ``Looking for work, unemployed''), and ``student.'' All other original values that do not fit into these categories are treated as missing for employment status. Lastly, we replace individual-year observations of occupation with employment status when employment status is non-employed (out-of-labor-force, unemployed, or student). So employment status replaces missing values of occupation, but it also replaces valid occupation codes when employment status is one of the three non-employed statuses, meaning that non-employed statuses take priority over occupation.

After constructing our dynamic variables of interest, we filter individuals and individual-year observations with invalid values for these variables. Our data filtering process starts with 35,516 individuals with 360,373 individual-year observations after the 1981 survey (inclusive), when the individual was either the household head or the spouse of the head.

We start with restricting our dataset to individual-year observations that have ``sequence number'' values between 1 and 20, meaning the individual lives in the household, leading to 35,298 individuals and 352,191 individual-year observations.
We then restrict individual-year observations with age between 18 and 80 (inclusive), resulting in 344,682 individual-year observations from 35,068 unique individuals. Then, we drop 2,999 individuals whose occupation status is not in the labor force across all years, resulted in 32,069 unique individuals and 323,420 individual-year observations.
After combining occupation and employment status into our final occupation variable, we drop 5,037 individual-year observations with missing or invalid values for occupation, leading to 31,795 individuals and 318,383 individual-year observations.
We drop 632 individuals with 4,512 individual-year observations with missing educational information even after the forward filling, which corresponds to individuals whom we never observe years of education and individual-year observations that occur before the first non-missing observation of years of education.
The filtering on educational level leads to 31,163 individuals and 313,871 individual-year observations.
Finally, 107 individual (249 individual-year observations) with no observation of family state (for the region variable), resulted in 31,056 individuals and 313,622 individual-year observations.
After the processing above, we have no missing values for personal id or gender, or birth year, and race/ethnicity has no missing values by construction (other/unknown category). The sequential filtering steps lead to the final PSID81 dataset used in this study.

\subsection{National Longitudinal Survey of Youth (NLSY)}

The National Longitudinal Survey of Youth of 1979 (NLSY79) and 1997 (NLSY97) are two cohort-based surveys sponsored by the U.S. Bureau of Labor Statistics that follow individuals born in the United States.

\paragraph*{NLSY79} The NLSY79 includes individuals born between 1957 and 1964 who were between 14 and 22 years old at the time data collection started in 1979. The original cohort contained 12,686 respondents. These individuals were interviewed annually from 1979 through 1994, and biennially thereafter. We use data from surveys conducted 1979 through 2020.

To replicate the results in our study, researchers can download the NLSY79 data file at \href{https://www.nlsinfo.org/investigator/pages/search}{https://www.nlsinfo.org/investigator/pages/search}. After creating an account, the researcher can search and select the variables listed, and download the data file. After the pre-processing described below, our resulting dataset, which we refer to as NLSY79, has 12,479 individuals and 259,778 total individual-year observations of occupations.

As in the PSID81 dataset, we use five static covariates for each individual, dropping individuals for whom this information is missing: personal id, gender, race/ethnicity, region, and birth year. Personal id requires no processing. We use the main NLSY variables for gender, race/ethnicity, and birth year. There are no missing values for these variables and the only processing is descriptive labeling. Gender has two values: ``male'' or ``female.'' Race/ethnicity has three values: ``Hispanic,'' ``Black,'' or ``non-Hispanic/non-Black.'' Birth year has eight values from ``1959'' to ``1964.''

The region variable is recorded each survey year by the NLSY as one of four values: ``northeast,'' ``south,'' ``west,'' and ``northcentral.'' We take the first non-missing observation as our static variable. We drop 2 individuals with no region information in any year.

We construct two dynamic variables for each individual-year observation, dropping observations for which either variable is missing: education level and occupation. We construct education level based on the years of education recorded each survey year in the NLSY through 2016.\footnote{This variable is labeled ``highest degree obtained'' by NLSY, but captures years of education rather than just completed degrees.} For 2018 and 2020, we use the same educational level as in 2016. When we compare the highest degree obtained in 2016 to the highest degree ever obtained, we have a $99.59\%$ match.

We categorize years of education into ``less than high school,'' ``high school,'' ``some college,'' ``college,'' and ``any graduate'' each year, then forward-fill education to replace missing values and impose the restriction that education level be non-decreasing. We drop 12 individual-year observations because of invalid skip and 12 individual-year observations because of non-interview that occur prior to the first valid observation of education for an individual.

We again construct our main variable of interest, occupation, using similar pre-processing steps applied by \cite{vafa_career_2024} to facilitate comparisons, combining information from multiple variables recorded each survey year by the NLSY.  For the occupation variable, we crosswalk individual-year observations, which are recorded as either 1970 or 2000 census codes, to 1990 census codes for consistency across datasets (\cite{occ1990dd}). The educational enrollment status variable requires no processing beyond descriptive labels and has two values: ``yes'' or ``no,'' where yes means the individual is a student that year.

Employment status is recorded on a weekly basis, with retrospective updating. To create employment status at the year level, we take the most frequent informative response (i.e., not the ``no information'' or ``not working'' status, where the latter does not differentiate unemployed from out of labor force, or other missing values). We then collapse the employment status variable into three categories: ``employed'' (defined as ``active miliary service,'' ``associated with employment,'' or any value that corresponds to a ``job number''), ``out of labor force'' (defined as ``not associated with employment'' or ``out of labor force'') and ``unemployed.'' All other original values that do not fit into these categories are treated as missing for employment status.

To combine the occupation, educational enrollment status, and employment status variables into our final processed occupation variable, we do the following for  each individual-year observation: We use ``student'' when educational enrollment status is yes. If not, we use ``out of labor force'' or ``unemployed'' if employment status is one of those values. If the occupation is still undecided, we use occupational code if it is specified. After combining occupation, educational enrollment status and employment status into our final occupation variable, we drop 108,034 individual-year observations with missing or invalid values for occupation.

\paragraph*{NLSY97} The NLSY97 includes individuals born between 1980 and 1984 who were between 12 and 17 years old at the time data collection started in 1997. The original cohort contained 8,984 respondents. These individuals were interviewed annually from 1997 through 2011, and biennially thereafter. We use data from surveys conducted 1997 through 2021.
To replicate the results in our study, researchers can download the NLSY97 data file at \href{https://www.nlsinfo.org/investigator/pages/search}{https://www.nlsinfo.org/investigator/pages/search}. After creating an account, the researcher can search and select the variables listed, and download the data file. One can find official tutorials of accessing NLSY data at \href{https://www.nlsinfo.org/content/getting-started/introduction-to-the-nls/tutorials-and-videos}{https://www.nlsinfo.org/content/getting-started/introduction-to-the-nls/tutorials-and-videos}. After the pre-processing described below, our resulting dataset, which we refer to as NLSY97, has 8,984 individuals and 148,795 total individual-year observations of occupations.

As in the other two datasets, we use five static covariates for each individual, dropping individuals for whom this information is missing: personal id, gender, race/ethnicity, region, and birth year. Personal id requires no processing. We use the main NLSY variables for gender, race/ethnicity, and birth year. There are no missing values for these variables and the only processing is descriptive labeling. Gender has two values: ``male'' or ``female.'' Differing from NLSY79, race/ethnicity has four values: ``Hispanic or Latino,'' ``Black or African American,'' ``mixed race non-Hispanic,'' or ``non-Hispanic/non-Black.'' Birth year has five values from ``1980'' to ``1984.''

As in the NLSY79, the region variable is recorded each survey year as one of four values: ``northeast,'' ``south,'' ``west,'' and ``northcentral;'' however, there are no missing values for the first year 1997, so we download only the variable for 1997 and use it as our static variable.

The construction of the two dynamic variables, education level and occupation, for each individual-year observation also follows our process for NLSY79. Unlike the NLSY79, the education variable we use records highest \textit{degree} achieved each survey year, so we do not need to convert years of education to degree. We do some aggregation to achieve the same levels as other datasets: ``less than high school'' (defined as ``none'' or ``GED''), ``high school,'' ``some college,'' ``college,'' and ``any graduate'' (defined as ``Master's,'' ``PhD,'' or ``Professional Degree''). As in the other datasets, we forward-fill education to replace missing values and impose the restriction that education level be non-decreasing. There are no individual-year observations that occur before the first non-missing observation of years of education and no individuals for whom we never observe years of education.

We again construct our main variable of interest, occupation, using the same pre-processing steps applied by \cite{vafa_career_2024} to facilitate comparisons, combining information from multiple variables recorded each survey year by the NLSY. For the occupation variable, we crosswalk individual-year observations from the 2000 census codes to 1990 census codes for consistency across datasets (\cite{occ1990dd}).\footnote{To have the right number of digits for the crosswalk, we divide each occupation code by ten.} There are many ``non enrolled'' and ``enrolled'' values for the educational enrollment status variables, which we aggregate.

As in the NLSY79, employment status is recorded on a weekly basis, with retrospective updating. To create employment status at the year level, we take the most frequent informative response (i.e., not the ``no information'' or ``not working'' status). We then collapse the employment status variable into three categories: ``employed'' (defined as ``active miliary service,'' ``associated with employment,'' or any value that corresponds to a ``job number''), ``out of labor force'' (defined as ``not associated with employment'' or ``out of labor force'') and ``unemployed.'' All other original values that do not fit into these categories are treated as missing for employment status.

To combine the occupation, educational enrollment status, and employment status variables into our final processed occupation variable, we do the following for  each individual-year observation: We use ``student'' when educational enrollment status is enrolled. If not, we use ``out of labor force'' or ``unemployed'' if employment status is one of those values. If the occupation is still undecided, we use occupational code if it is specified. After combining occupation, educational enrollment status and employment status into our final occupation variable, we drop 30,885 individual-year observations with missing or invalid values for occupation.

\subsection{O*NET}
The O*NET dataset is the main occupational information database in the United States, developed by the U.S. Department of Labor. For each occupation, it includes the following occupational characteristics, encoded as text: Tasks, Technology Skills, Tools Used, Work Activities, Detailed Work Activities, Work Context, Job Zone, Skills, Knowledge, Abilities, Interests, Work Values, Work Styles, Related Occupations. The O*NET data is publicly available and can be accessed at \href{https://www.onetonline.org/link/details/onet_code}{online}.

We match O*NET data for 335 job titles from career trajectories we built on survey data to further train LABOR-LLM models. O*NET variables included in this matching process are Skills, Knowledge, Abilities, Tasks, Interests, Work Styles, Work Activities, Work Values, and Related Job Titles. We use these variables to build textual representations based on the job description (which includes up to five descriptions from the closest matching SOC codes), categorical data (Skills through Work Values, calculating the average importance score for each variable across all matching SOC and selecting the top five), and Related Job Titles (sample up to five specific job titles from the closest matching SOC codes). We generate one text file for each job title in our dataset.

\subsection{Summary Statistics}
\label{sec-appendix-summarystats}

Table \ref{tab:demographic-distributions} provides summary statistics by dataset for the demographic variables we use in our analysis. Recall that the demographics are assigned to be constant within our cleaned dataset even if they changed over time in the original survey data. Note further that the ethnicity encoding across datasets is slightly different.

\begin{table}[t]
    \centering
    \caption{Share of observations with different demographic characteristics. \updated{passed replication check 12/04/2025}}
    \label{tab:demographic-distributions}
    \footnotesize
    \resizebox{\textwidth}{!}{
    \begin{tabular}{lcccccc}
    \toprule
     & \multicolumn{2}{c}{PSID81} & \multicolumn{2}{c}{NLSY79} & \multicolumn{2}{c}{NLSY97} \\
     & Individual & Transition & Individual & Transition & Individual & Transition \\
    \midrule
    \textbf{Gender} &  &  &  &  &  &  \\
    Female & 50.5\% & 54.2\% & 49.7\% & 51.6\% & 48.8\% & 50.3\% \\
    Male & 49.5\% & 45.8\% & 50.3\% & 48.4\% & 51.2\% & 49.7\% \\
    \midrule
    \textbf{Ethnicity} & & & & & & \\
    Black & - & - & 25.1\% & 28.1\% & - & - \\
    Black or African-American & 34.5\% & 32.1\% & - & - & 26.0\% & 26.9\% \\
    Hispanic & - & - & 16.0\% & 17.9\% & - & - \\
    Hispanic or Latino & - & - & - & - & 21.2\% & 21.3\% \\
    Mixed-Race Non-Hispanic & - & - & - & - & 0.9\% & 0.9\% \\
    Non-Black Non-Hispanic & - & - & 58.9\% & 54.1\% & 51.9\% & 50.9\% \\
    Other or Unknown & 6.1\% & 3.1\% & - & - & - & - \\
    White & 59.4\% & 64.7\% & - & - & - & - \\
    \midrule
    \textbf{Region} & & & & & & \\
    Northcentral & 24.2\% & 25.8\% & 23.8\% & 25.2\% & 22.8\% & 22.8\% \\
    Northeast & 13.7\% & 15.4\% & 20.4\% & 19.2\% & 17.6\% & 17.3\% \\
    South & 43.9\% & 41.8\% & 36.7\% & 37.0\% & 37.4\% & 37.8\% \\
    West & 18.2\% & 17.1\% & 19.1\% & 18.6\% & 22.2\% & 22.1\% \\
    \bottomrule
    \end{tabular}
    }
\end{table}

\begin{figure}[!htbp]
    \centering
    \includegraphics[width=\linewidth]{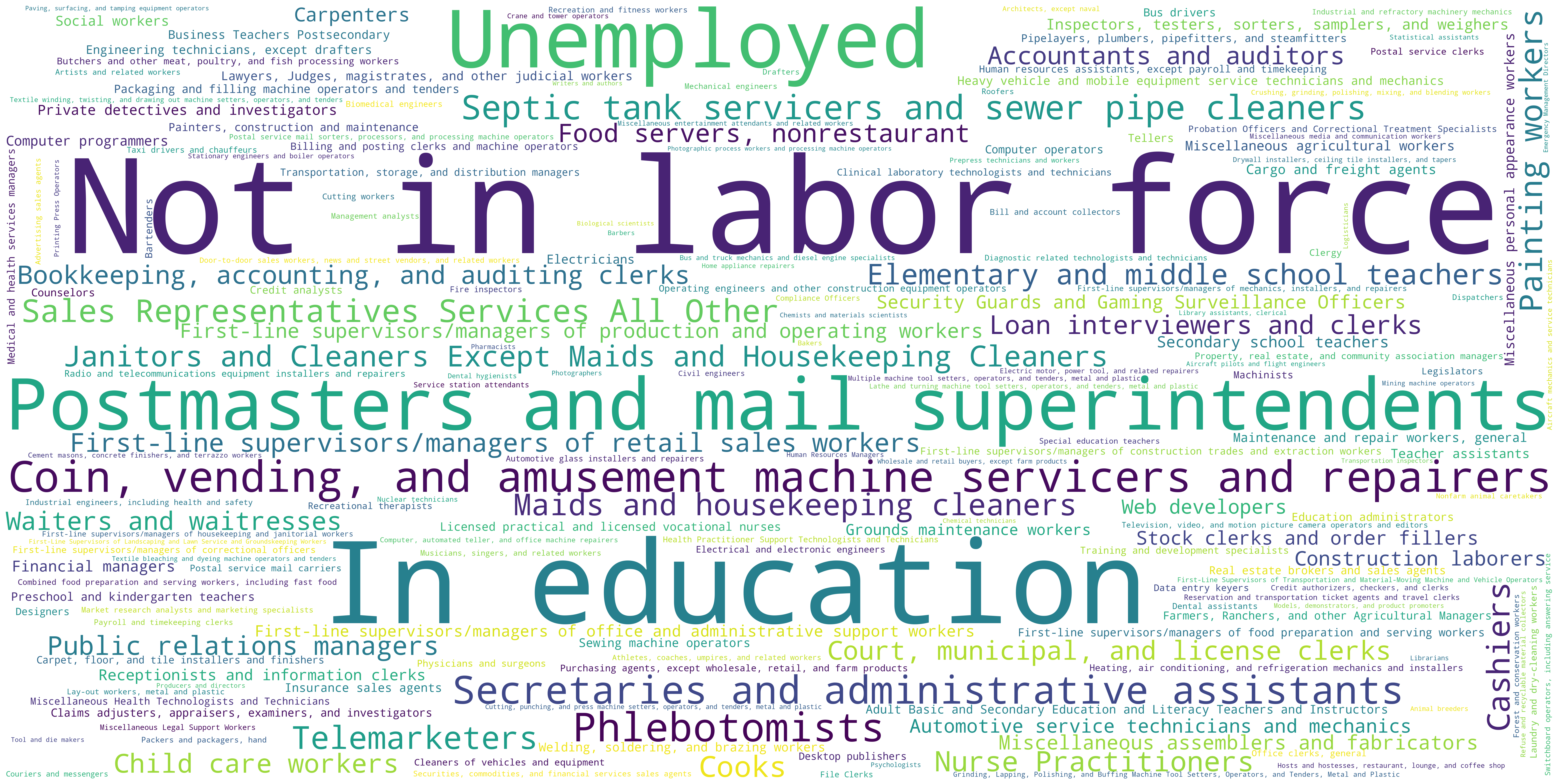}
    \caption{Word cloud of job titles, scaled by title popularity.\updated{12/04/2025}}
    \label{fig:word-cloud-job-titles}
\end{figure}

Figure \ref{fig:word-cloud-job-titles} presents example job titles in a word cloud, weighted by their popularity. Each job title's font size is scaled proportionally to its frequency in the test sets of the three datasets (PSID81, NLSY79, NLSY97) combined, measured by the number of individual-year observations; thus, more prevalent occupations appear larger, highlighting their distribution within our labor market data.

\subsection{Combined data sources}

Once pre-processed, the PSID81, NLSY79, and NLSY97 datasets are used to construct the input files to fine-tune all predictive models included in this project on their career trajectory data and covariates. For this purpose, each dataset is divided into three subdatasets: training, validation and test. The construction of the datasets for this stage follows \cite{vafa_career_2024}. The resumes or sequences of occupations are prepared into individual data files for the split they correspond to. That is, the resume data resulting from PSID81, NLSY79, and NLSY97 is structured as ``train.job,'' ``valid.job,'' and ``test.job.'' In each file, each row corresponds to one individual in the sample, and occupations are designated using a classification code, such as O*NET or occ1990dd. Each covariate included in the dataset follows the same structure, and it should have the same number of rows as the job file associated, corresponding to the same group of individuals.

\end{appendix}

%%%%%%%%%%%%%%%%%%%%%%%%%%%%%%%%%%%%%%%%%%%%%%
%% Bibliography:                            %%
%%%%%%%%%%%%%%%%%%%%%%%%%%%%%%%%%%%%%%%%%%%%%%
%% IMPORTANT: References in the bibliography should be complete,
%% including the first and last names, and date of publication.

%% If your bibliography is in bibtex format, uncomment commands:
\bibliographystyle{qe.bst} % Style BST file
%\bibliography{bibliography}  % Bibliography file (usually '*.bib')
\bibliography{references_qe.bib}

%% Or include bibliography directly:
% \begin{thebibliography}{}
% \bibitem{b1}
% \end{thebibliography}
\newpage
\section*{\Huge \textcolor{red}{Online Appendices}}
% (Online) Test Section \ref{sec:online-appendix-occupation-titles} has a Table \ref{tab:all-job-titles}.
% \textcolor{red}{\emph{Note: Section and table/figure numbers reset.}}

% \renewcommand{\appendixname}{Online}
% \renewcommand{\thefigure}{Online\thesection.\arabic{figure}}
% \renewcommand{\thetable}{Online\thesection.\arabic{table}}
\begin{appendix}
\input{qe_supplementary_0}

\input{qe_supplementary_1}
\end{appendix}

\end{document}

%% file: qe_supplementary_0.tex
\section{Occupation Title Tables}
\label{sec:online-appendix-occupation-titles}
Table \ref{tab:all-job-titles} provides a mapping from the occupation codes in the \texttt{occ1990dd} scheme, as well as the employment statuses ``unemployed'', ``out-of-labor-force'', and ``in education'', to the job titles used in our paper.

\tiny
\begin{longtable}{p{0.2\textwidth} p{0.8\textwidth}}
\caption{List of \texttt{occ1990dd} codes and their corresponding job titles. \label{tab:all-job-titles}} \\
\toprule
occ1990dd & Job Title \\
\midrule
\endfirsthead
\caption[]{List of \texttt{occ1990dd} codes and their corresponding job titles (continued)} \\
\toprule
occ1990dd & Job Title \\
\midrule
\endhead
\bottomrule
\endfoot
103 & Physical therapists \\
104 & Audiologists \\
105 & Recreational therapists \\
106 & Physician assistants \\
13 & Public relations managers \\
14 & Education administrators \\
15 & Medical and health services managers \\
154 & Business Teachers Postsecondary \\
155 & Preschool and kindergarten teachers \\
156 & Elementary and middle school teachers \\
157 & Secondary school teachers \\
158 & Special education teachers \\
159 & Adult Basic and Secondary Education and Literacy Teachers and Instructors \\
163 & Counselors \\
164 & Librarians \\
165 & Archivists, curators, and museum technicians \\
166 & Market research analysts and marketing specialists \\
167 & Psychologists \\
169 & Sociologists \\
173 & Urban and regional planners \\
174 & Social workers \\
176 & Clergy \\
177 & Probation Officers and Correctional Treatment Specialists \\
178 & Lawyers, Judges, magistrates, and other judicial workers \\
18 & Property, real estate, and community association managers \\
183 & Writers and authors \\
184 & Technical writers \\
185 & Designers \\
186 & Musicians, singers, and related workers \\
187 & Producers and directors \\
188 & Artists and related workers \\
189 & Photographers \\
19 & Morticians, Undertakers, and Funeral Directors \\
193 & Dancers and choreographers \\
194 & Miscellaneous media and communication workers \\
195 & Television, video, and motion picture camera operators and editors \\
198 & Announcers \\
199 & Athletes, coaches, umpires, and related workers \\
203 & Clinical laboratory technologists and technicians \\
204 & Dental hygienists \\
205 & Medical records and health information technicians \\
206 & Diagnostic related technologists and technicians \\
207 & Licensed practical and licensed vocational nurses \\
208 & Miscellaneous Health Technologists and Technicians \\
214 & Engineering technicians, except drafters \\
217 & Drafters \\
218 & Surveyors, cartographers, and photogrammetrists \\
22 & Postmasters and mail superintendents \\
223 & Agricultural and food science technicians \\
224 & Chemical technicians \\
225 & Geological and petroleum technicians \\
226 & Aircraft pilots and flight engineers \\
227 & Air traffic controllers and airfield operations specialists \\
228 & Broadcast and sound engineering technicians and radio operators \\
229 & Computer programmers \\
23 & Accountants and auditors \\
233 & Computer control programmers and operators \\
234 & Miscellaneous Legal Support Workers \\
235 & Nuclear technicians \\
24 & Insurance underwriters \\
243 & First-line supervisors/managers of retail sales workers \\
25 & Credit analysts \\
253 & Insurance sales agents \\
254 & Real estate brokers and sales agents \\
255 & Securities, commodities, and financial services sales agents \\
256 & Advertising sales agents \\
258 & Sales engineers \\
26 & Management analysts \\
27 & Training and development specialists \\
274 & Sales Representatives Services All Other \\
275 & Telemarketers \\
276 & Cashiers \\
277 & Door-to-door sales workers, news and street vendors, and related workers \\
28 & Purchasing agents and buyers, farm products \\
283 & Models, demonstrators, and product promoters \\
29 & Wholesale and retail buyers, except farm products \\
303 & First-line supervisors/managers of office and administrative support workers \\
308 & Computer operators \\
313 & Secretaries and administrative assistants \\
315 & Desktop publishers \\
316 & Credit authorizers, checkers, and clerks \\
317 & Hotel, motel, and resort desk clerks \\
318 & Reservation and transportation ticket agents and travel clerks \\
319 & Receptionists and information clerks \\
326 & Correspondence clerks \\
328 & Human resources assistants, except payroll and timekeeping \\
329 & Library assistants, clerical \\
33 & Purchasing agents, except wholesale, retail, and farm products \\
335 & File Clerks \\
336 & Brokerage clerks \\
337 & Bookkeeping, accounting, and auditing clerks \\
338 & Payroll and timekeeping clerks \\
34 & Agents and business managers of artists, performers, and athletes \\
344 & Billing and posting clerks and machine operators \\
346 & Postal service mail sorters, processors, and processing machine operators \\
347 & Office machine operators, except computer \\
348 & Switchboard operators, including answering service \\
349 & Communications Equipment Operators All Other \\
35 & Construction and building inspectors \\
354 & Postal service clerks \\
355 & Postal service mail carriers \\
356 & Mail clerks and mail machine operators, except postal service \\
357 & Couriers and messengers \\
359 & Dispatchers \\
36 & Compliance Officers \\
364 & Cargo and freight agents \\
365 & Stock clerks and order fillers \\
366 & Meter readers, utilities \\
368 & Weighers, measurers, checkers, and samplers, recordkeeping \\
37 & Emergency Management Directors \\
373 & Transportation, storage, and distribution managers \\
375 & Claims adjusters, appraisers, examiners, and investigators \\
376 & Loan interviewers and clerks \\
377 & Eligibility interviewers, government programs \\
378 & Bill and account collectors \\
379 & Office clerks, general \\
383 & Tellers \\
384 & Proofreaders and copy markers \\
385 & Data entry keyers \\
386 & Statistical assistants \\
387 & Teacher assistants \\
389 & Court, municipal, and license clerks \\
4 & Legislators \\
405 & Maids and housekeeping cleaners \\
408 & Laundry and dry-cleaning workers \\
415 & First-Line Supervisors of Protective Service Workers (All Other) \\
417 & Fire inspectors \\
418 & Private detectives and investigators \\
423 & First-line supervisors/managers of correctional officers \\
425 & Crossing guards \\
426 & Security Guards and Gaming Surveillance Officers \\
427 & Animal control workers \\
43 & Architects, except naval \\
433 & First-line supervisors/managers of food preparation and serving workers \\
434 & Bartenders \\
435 & Waiters and waitresses \\
436 & Cooks \\
439 & Combined food preparation and serving workers, including fast food \\
44 & Aerospace engineers \\
444 & Food servers, nonrestaurant \\
445 & Dental assistants \\
447 & Phlebotomists \\
448 & First-line supervisors/managers of housekeeping and janitorial workers \\
45 & Materials engineers \\
450 & First-Line Supervisors of Landscaping and Lawn Service and Groundskeeping Workers \\
451 & Grounds maintenance workers \\
453 & Janitors and Cleaners Except Maids and Housekeeping Cleaners \\
455 & Pest control workers \\
457 & Barbers \\
458 & Miscellaneous personal appearance workers \\
459 & Miscellaneous entertainment attendants and related workers \\
461 & Tour and travel guides \\
462 & Ushers, lobby attendants, and ticket takers \\
464 & Baggage porters, bellhops, and concierges \\
466 & Recreation and fitness workers \\
467 & Motion picture projectionists \\
468 & Child care workers \\
469 & Hosts and hostesses, restaurant, lounge, and coffee shop \\
47 & Petroleum engineers \\
470 & First-line supervisors/managers of personal service workers \\
471 & Transportation inspectors \\
472 & Nonfarm animal caretakers \\
473 & Farmers, Ranchers, and other Agricultural Managers \\
475 & Animal breeders \\
479 & Miscellaneous agricultural workers \\
48 & Chemical engineers \\
488 & Graders and sorters, agricultural products \\
489 & Agricultural inspectors \\
496 & Forest and conservation workers \\
498 & Fishers and related fishing workers \\
503 & First-line supervisors/managers of mechanics, installers, and repairers \\
505 & Automotive service technicians and mechanics \\
507 & Bus and truck mechanics and diesel engine specialists \\
508 & Aircraft mechanics and service technicians \\
509 & Small engine mechanics \\
514 & Automotive glass installers and repairers \\
516 & Heavy vehicle and mobile equipment service technicians and mechanics \\
518 & Industrial and refractory machinery mechanics \\
519 & Maintenance workers, machinery \\
523 & Electronic home entertainment equipment installers and repairers \\
525 & Computer, automated teller, and office machine repairers \\
526 & Home appliance repairers \\
527 & Radio and telecommunications equipment installers and repairers \\
53 & Civil engineers \\
533 & Electronic equipment installers and repairers, motor vehicles \\
534 & Heating, air conditioning, and refrigeration mechanics and installers \\
535 & Precision instrument and equipment repairers \\
536 & Locksmiths and safe repairers \\
539 & Control and valve installers and repairers \\
543 & Elevator installers and repairers \\
544 & Millwrights \\
549 & Maintenance and repair workers, general \\
55 & Electrical and electronic engineers \\
558 & First-line supervisors/managers of construction trades and extraction workers \\
56 & Industrial engineers, including health and safety \\
563 & Carpet, floor, and tile installers and finishers \\
567 & Carpenters \\
57 & Mechanical engineers \\
573 & Drywall installers, ceiling tile installers, and tapers \\
575 & Electricians \\
577 & Electric motor, power tool, and related repairers \\
579 & Painters, construction and maintenance \\
583 & Paperhangers \\
584 & Plasterers and stucco masons \\
585 & Pipelayers, plumbers, pipefitters, and steamfitters \\
588 & Cement masons, concrete finishers, and terrazzo workers \\
589 & Glaziers \\
59 & Biomedical engineers \\
593 & Insulation workers \\
594 & Paving, surfacing, and tamping equipment operators \\
595 & Roofers \\
597 & Structural metal fabricators and fitters \\
598 & Earth drillers, except oil and gas \\
599 & Sheet metal workers \\
614 & Derrick, rotary drill, and service unit operators, oil, gas, and mining \\
615 & Explosives workers, ordnance handling experts, and blasters \\
616 & Mining machine operators \\
617 & Roof bolters, mining \\
628 & First-line supervisors/managers of production and operating workers \\
634 & Tool and die makers \\
637 & Machinists \\
64 & Web developers \\
643 & Boilermakers \\
644 & Tool grinders, filers, and sharpeners \\
645 & Model makers and patternmakers, metal and plastic \\
649 & Etchers and engravers \\
65 & Logisticians \\
653 & Lay-out workers, metal and plastic \\
657 & Cabinetmakers and bench carpenters \\
658 & Furniture finishers \\
66 & Actuaries \\
666 & Tailors, dressmakers, and sewers \\
668 & Upholsterers \\
669 & Shoe and leather workers and repairers \\
675 & Molders, shapers, and casters, except metal and plastic \\
677 & Opticians, dispensing \\
678 & Health Practitioner Support Technologists and Technicians \\
679 & Print Binding and Finishing Workers \\
68 & Statisticians \\
684 & Multiple machine tool setters, operators, and tenders, metal and plastic \\
686 & Butchers and other meat, poultry, and fish processing workers \\
687 & Bakers \\
688 & Food batchmakers \\
69 & Astronomers and physicists \\
694 & Water and liquid waste treatment plant and system operators \\
695 & Power plant operators, distributors, and dispatchers \\
696 & Stationary engineers and boiler operators \\
699 & Miscellaneous plant and system operators \\
7 & Financial managers \\
703 & Lathe and turning machine tool setters, operators, and tenders, metal and plastic \\
706 & Cutting, punching, and press machine setters, operators, and tenders, metal and plastic \\
707 & Rolling machine setters, operators, and tenders, metal and plastic \\
708 & Drilling and boring machine tool setters, operators, and tenders, metal and plastic \\
709 & Grinding, Lapping, Polishing, and Buffing Machine Tool Setters, Operators, and Tenders, Metal and Plastic \\
713 & Forging machine setters, operators, and tenders, metal and plastic \\
719 & Molders and molding machine setters, operators, and tenders, metal and plastic \\
723 & Plating and coating machine setters, operators, and tenders, metal and plastic \\
724 & Heat treating equipment setters, operators, and tenders, metal and plastic \\
727 & Sawing machine setters, operators, and tenders, wood \\
729 & Woodworking machine setters, operators, and tenders, except sawing \\
73 & Chemists and materials scientists \\
733 & Woodworkers (All Other) \\
734 & Printing Press Operators \\
736 & Prepress technicians and workers \\
738 & Textile winding, twisting, and drawing out machine setters, operators, and tenders \\
739 & Textile knitting and weaving machine setters, operators, and tenders \\
74 & Atmospheric and space scientists \\
743 & Textile cutting machine setters, operators, and tenders \\
744 & Sewing machine operators \\
745 & Shoe machine operators and tenders \\
747 & Pressers, textile, garment, and related materials \\
749 & Textile bleaching and dyeing machine operators and tenders \\
75 & Environmental scientists and geoscientists \\
753 & Cementing and gluing machine operators and tenders \\
754 & Packaging and filling machine operators and tenders \\
755 & Extruding and drawing machine setters, operators, and tenders, metal and plastic \\
756 & Crushing, grinding, polishing, mixing, and blending workers \\
757 & Chemical processing machine setters, operators, and tenders \\
76 & Physical scientists, all other \\
763 & Food and tobacco roasting, baking, and drying machine operators and tenders \\
764 & Cleaning, washing, and metal pickling equipment operators and tenders \\
765 & Paper goods machine setters, operators, and tenders \\
766 & Metal furnace and kiln operators and tenders \\
769 & Cutting workers \\
77 & Agricultural and food scientists \\
774 & Photographic process workers and processing machine operators \\
779 & Painting workers \\
78 & Biological scientists \\
783 & Welding, soldering, and brazing workers \\
785 & Miscellaneous assemblers and fabricators \\
789 & Painters, Transportation Equipment \\
79 & Conservation scientists and foresters \\
799 & Inspectors, testers, sorters, samplers, and weighers \\
8 & Human Resources Managers \\
803 & First-Line Supervisors of Transportation and Material-Moving Machine and Vehicle Operators \\
804 & Coin, vending, and amusement machine servicers and repairers \\
808 & Bus drivers \\
809 & Taxi drivers and chauffeurs \\
813 & Parking lot attendants \\
823 & Railroad conductors and yardmasters \\
824 & Locomotive engineers and operators \\
825 & Railroad brake, signal, and switch operators \\
829 & Ship and boat captains and operators \\
83 & Medical scientists \\
834 & Bridge and lock tenders \\
84 & Physicians and surgeons \\
844 & Operating engineers and other construction equipment operators \\
848 & Crane and tower operators \\
85 & Dentists \\
853 & Dredge, excavating, and loading machine operators \\
859 & Shuttle car operators \\
86 & Veterinarians \\
865 & Helpers--installation, maintenance, and repair workers \\
866 & Helpers, construction trades \\
869 & Construction laborers \\
87 & Optometrists \\
873 & Helpers--production workers \\
875 & Refuse and recyclable material collectors \\
878 & Machine feeders and offbearers \\
88 & Podiatrists \\
885 & Service station attendants \\
887 & Cleaners of vehicles and equipment \\
888 & Packers and packagers, hand \\
889 & Septic tank servicers and sewer pipe cleaners \\
89 & Health diagnosing and treating practitioners, all other \\
905 & Air Crew Officers \\
95 & Nurse Practitioners \\
96 & Pharmacists \\
97 & Dietitians and nutritionists \\
98 & Respiratory therapists \\
99 & Occupational therapist assistants and aides \\
999 & Special Code (occ1990dd=999) \\
education & In education \\
not\_in\_labor\_force & Not in labor force \\
unemployed & Unemployed \\
\end{longtable}
\normalsize

We create a set of numeric job titles in this paper to assess the impact of literal job titles on FT-LABOR-LLM's performance. Specifically, we first randomly permute the list of all job titles, then translate each literal job title into a numeric one ranging from \texttt{job\_001} to \texttt{job\_334}. Table \ref{tab:all-numeric-titles} lists the complete mapping from literal job titles to the numeric ones.

\tiny
\begin{longtable}{p{0.8\textwidth} p{0.2\textwidth}}
\caption{List of numeric job titles and their corresponding English language job titles. \label{tab:all-numeric-titles}} \\
\toprule
Job Title & Numerical Job Title \\
\midrule
\endfirsthead
\caption[]{List of numeric job titles and their corresponding English language job titles. (continued)} \\
\toprule
Job Title & Numerical Job Title \\
\midrule
\endhead
\bottomrule
\endfoot
Transportation inspectors & job\_000 \\
In education & job\_001 \\
Bill and account collectors & job\_002 \\
Audiologists & job\_003 \\
Credit analysts & job\_004 \\
Web developers & job\_005 \\
Model makers and patternmakers, metal and plastic & job\_006 \\
Inspectors, testers, sorters, samplers, and weighers & job\_007 \\
Laundry and dry-cleaning workers & job\_008 \\
Radio and telecommunications equipment installers and repairers & job\_009 \\
Dancers and choreographers & job\_010 \\
Plasterers and stucco masons & job\_011 \\
Power plant operators, distributors, and dispatchers & job\_012 \\
Helpers--production workers & job\_013 \\
Electronic home entertainment equipment installers and repairers & job\_014 \\
Secondary school teachers & job\_015 \\
Structural metal fabricators and fitters & job\_016 \\
Civil engineers & job\_017 \\
Tool and die makers & job\_018 \\
Painting workers & job\_019 \\
Welding, soldering, and brazing workers & job\_020 \\
Nurse Practitioners & job\_021 \\
Agricultural and food science technicians & job\_022 \\
Carpenters & job\_023 \\
Air Crew Officers & job\_024 \\
Roof bolters, mining & job\_025 \\
Dental hygienists & job\_026 \\
Maids and housekeeping cleaners & job\_027 \\
Miscellaneous Legal Support Workers & job\_028 \\
Biomedical engineers & job\_029 \\
Communications Equipment Operators All Other & job\_030 \\
Combined food preparation and serving workers, including fast food & job\_031 \\
Textile cutting machine setters, operators, and tenders & job\_032 \\
Chemists and materials scientists & job\_033 \\
Heat treating equipment setters, operators, and tenders, metal and plastic & job\_034 \\
Pipelayers, plumbers, pipefitters, and steamfitters & job\_035 \\
Food and tobacco roasting, baking, and drying machine operators and tenders & job\_036 \\
Physical therapists & job\_037 \\
Pest control workers & job\_038 \\
Brokerage clerks & job\_039 \\
Astronomers and physicists & job\_040 \\
Derrick, rotary drill, and service unit operators, oil, gas, and mining & job\_041 \\
Miscellaneous assemblers and fabricators & job\_042 \\
Private detectives and investigators & job\_043 \\
First-line supervisors/managers of housekeeping and janitorial workers & job\_044 \\
Cashiers & job\_045 \\
Surveyors, cartographers, and photogrammetrists & job\_046 \\
Emergency Management Directors & job\_047 \\
Food servers, nonrestaurant & job\_048 \\
Animal breeders & job\_049 \\
Septic tank servicers and sewer pipe cleaners & job\_050 \\
Receptionists and information clerks & job\_051 \\
Proofreaders and copy markers & job\_052 \\
Recreational therapists & job\_053 \\
Transportation, storage, and distribution managers & job\_054 \\
Agents and business managers of artists, performers, and athletes & job\_055 \\
Couriers and messengers & job\_056 \\
Packaging and filling machine operators and tenders & job\_057 \\
Security Guards and Gaming Surveillance Officers & job\_058 \\
Electronic equipment installers and repairers, motor vehicles & job\_059 \\
Sociologists & job\_060 \\
Miscellaneous plant and system operators & job\_061 \\
Preschool and kindergarten teachers & job\_062 \\
Weighers, measurers, checkers, and samplers, recordkeeping & job\_063 \\
Ship and boat captains and operators & job\_064 \\
Writers and authors & job\_065 \\
Electric motor, power tool, and related repairers & job\_066 \\
Paper goods machine setters, operators, and tenders & job\_067 \\
Cleaning, washing, and metal pickling equipment operators and tenders & job\_068 \\
Payroll and timekeeping clerks & job\_069 \\
Archivists, curators, and museum technicians & job\_070 \\
Machinists & job\_071 \\
Dispatchers & job\_072 \\
Cement masons, concrete finishers, and terrazzo workers & job\_073 \\
Pressers, textile, garment, and related materials & job\_074 \\
Furniture finishers & job\_075 \\
Agricultural and food scientists & job\_076 \\
Purchasing agents, except wholesale, retail, and farm products & job\_077 \\
Billing and posting clerks and machine operators & job\_078 \\
Wholesale and retail buyers, except farm products & job\_079 \\
Locksmiths and safe repairers & job\_080 \\
Bus and truck mechanics and diesel engine specialists & job\_081 \\
Aircraft pilots and flight engineers & job\_082 \\
Human resources assistants, except payroll and timekeeping & job\_083 \\
Helpers, construction trades & job\_084 \\
Credit authorizers, checkers, and clerks & job\_085 \\
First-line supervisors/managers of mechanics, installers, and repairers & job\_086 \\
Medical records and health information technicians & job\_087 \\
Computer control programmers and operators & job\_088 \\
Education administrators & job\_089 \\
Earth drillers, except oil and gas & job\_090 \\
Materials engineers & job\_091 \\
Environmental scientists and geoscientists & job\_092 \\
Sales engineers & job\_093 \\
Prepress technicians and workers & job\_094 \\
Millwrights & job\_095 \\
Woodworkers (All Other) & job\_096 \\
First-Line Supervisors of Transportation and Material-Moving Machine and Vehicle Operators & job\_097 \\
Licensed practical and licensed vocational nurses & job\_098 \\
Pharmacists & job\_099 \\
Operating engineers and other construction equipment operators & job\_100 \\
Shoe machine operators and tenders & job\_101 \\
Counselors & job\_102 \\
Bakers & job\_103 \\
Tour and travel guides & job\_104 \\
Refuse and recyclable material collectors & job\_105 \\
Forging machine setters, operators, and tenders, metal and plastic & job\_106 \\
Chemical processing machine setters, operators, and tenders & job\_107 \\
Bridge and lock tenders & job\_108 \\
Producers and directors & job\_109 \\
Grinding, Lapping, Polishing, and Buffing Machine Tool Setters, Operators, and Tenders, Metal and Plastic & job\_110 \\
Sales Representatives Services All Other & job\_111 \\
Psychologists & job\_112 \\
First-Line Supervisors of Protective Service Workers (All Other) & job\_113 \\
First-line supervisors/managers of personal service workers & job\_114 \\
Atmospheric and space scientists & job\_115 \\
Respiratory therapists & job\_116 \\
Health Practitioner Support Technologists and Technicians & job\_117 \\
Lathe and turning machine tool setters, operators, and tenders, metal and plastic & job\_118 \\
Purchasing agents and buyers, farm products & job\_119 \\
Crushing, grinding, polishing, mixing, and blending workers & job\_120 \\
Insurance underwriters & job\_121 \\
Real estate brokers and sales agents & job\_122 \\
Optometrists & job\_123 \\
Tool grinders, filers, and sharpeners & job\_124 \\
Urban and regional planners & job\_125 \\
Technical writers & job\_126 \\
Heating, air conditioning, and refrigeration mechanics and installers & job\_127 \\
Bus drivers & job\_128 \\
Home appliance repairers & job\_129 \\
Cooks & job\_130 \\
Electrical and electronic engineers & job\_131 \\
Statisticians & job\_132 \\
Door-to-door sales workers, news and street vendors, and related workers & job\_133 \\
Office clerks, general & job\_134 \\
Financial managers & job\_135 \\
Molders and molding machine setters, operators, and tenders, metal and plastic & job\_136 \\
Grounds maintenance workers & job\_137 \\
Engineering technicians, except drafters & job\_138 \\
Maintenance and repair workers, general & job\_139 \\
Advertising sales agents & job\_140 \\
Multiple machine tool setters, operators, and tenders, metal and plastic & job\_141 \\
Precision instrument and equipment repairers & job\_142 \\
Electricians & job\_143 \\
Postmasters and mail superintendents & job\_144 \\
Barbers & job\_145 \\
Computer, automated teller, and office machine repairers & job\_146 \\
Clinical laboratory technologists and technicians & job\_147 \\
Tellers & job\_148 \\
Paperhangers & job\_149 \\
Correspondence clerks & job\_150 \\
Physical scientists, all other & job\_151 \\
Cementing and gluing machine operators and tenders & job\_152 \\
Bartenders & job\_153 \\
Miscellaneous Health Technologists and Technicians & job\_154 \\
Data entry keyers & job\_155 \\
Reservation and transportation ticket agents and travel clerks & job\_156 \\
Mechanical engineers & job\_157 \\
Special education teachers & job\_158 \\
Computer programmers & job\_159 \\
Securities, commodities, and financial services sales agents & job\_160 \\
Coin, vending, and amusement machine servicers and repairers & job\_161 \\
First-line supervisors/managers of production and operating workers & job\_162 \\
Loan interviewers and clerks & job\_163 \\
Automotive service technicians and mechanics & job\_164 \\
Railroad brake, signal, and switch operators & job\_165 \\
Graders and sorters, agricultural products & job\_166 \\
Property, real estate, and community association managers & job\_167 \\
Farmers, Ranchers, and other Agricultural Managers & job\_168 \\
Athletes, coaches, umpires, and related workers & job\_169 \\
Cutting workers & job\_170 \\
Sheet metal workers & job\_171 \\
Forest and conservation workers & job\_172 \\
Lay-out workers, metal and plastic & job\_173 \\
Logisticians & job\_174 \\
Control and valve installers and repairers & job\_175 \\
Animal control workers & job\_176 \\
Medical and health services managers & job\_177 \\
Cutting, punching, and press machine setters, operators, and tenders, metal and plastic & job\_178 \\
Stationary engineers and boiler operators & job\_179 \\
Small engine mechanics & job\_180 \\
Stock clerks and order fillers & job\_181 \\
Industrial and refractory machinery mechanics & job\_182 \\
Conservation scientists and foresters & job\_183 \\
Cargo and freight agents & job\_184 \\
Desktop publishers & job\_185 \\
Photographers & job\_186 \\
Carpet, floor, and tile installers and finishers & job\_187 \\
Physicians and surgeons & job\_188 \\
Public relations managers & job\_189 \\
Print Binding and Finishing Workers & job\_190 \\
Construction laborers & job\_191 \\
Secretaries and administrative assistants & job\_192 \\
Human Resources Managers & job\_193 \\
Postal service mail carriers & job\_194 \\
Crossing guards & job\_195 \\
Heavy vehicle and mobile equipment service technicians and mechanics & job\_196 \\
Maintenance workers, machinery & job\_197 \\
Woodworking machine setters, operators, and tenders, except sawing & job\_198 \\
First-line supervisors/managers of retail sales workers & job\_199 \\
Occupational therapist assistants and aides & job\_200 \\
Computer operators & job\_201 \\
Recreation and fitness workers & job\_202 \\
Probation Officers and Correctional Treatment Specialists & job\_203 \\
Metal furnace and kiln operators and tenders & job\_204 \\
Drywall installers, ceiling tile installers, and tapers & job\_205 \\
Nonfarm animal caretakers & job\_206 \\
Adult Basic and Secondary Education and Literacy Teachers and Instructors & job\_207 \\
Butchers and other meat, poultry, and fish processing workers & job\_208 \\
Miscellaneous agricultural workers & job\_209 \\
Statistical assistants & job\_210 \\
Painters, construction and maintenance & job\_211 \\
Rolling machine setters, operators, and tenders, metal and plastic & job\_212 \\
Office machine operators, except computer & job\_213 \\
First-line supervisors/managers of office and administrative support workers & job\_214 \\
Aerospace engineers & job\_215 \\
Morticians, Undertakers, and Funeral Directors & job\_216 \\
Drilling and boring machine tool setters, operators, and tenders, metal and plastic & job\_217 \\
Parking lot attendants & job\_218 \\
Elevator installers and repairers & job\_219 \\
Bookkeeping, accounting, and auditing clerks & job\_220 \\
Teacher assistants & job\_221 \\
Industrial engineers, including health and safety & job\_222 \\
Explosives workers, ordnance handling experts, and blasters & job\_223 \\
Sawing machine setters, operators, and tenders, wood & job\_224 \\
Waiters and waitresses & job\_225 \\
First-line supervisors/managers of food preparation and serving workers & job\_226 \\
Automotive glass installers and repairers & job\_227 \\
Phlebotomists & job\_228 \\
Crane and tower operators & job\_229 \\
First-line supervisors/managers of construction trades and extraction workers & job\_230 \\
Textile winding, twisting, and drawing out machine setters, operators, and tenders & job\_231 \\
Opticians, dispensing & job\_232 \\
Paving, surfacing, and tamping equipment operators & job\_233 \\
Miscellaneous entertainment attendants and related workers & job\_234 \\
Not in labor force & job\_235 \\
Hotel, motel, and resort desk clerks & job\_236 \\
Medical scientists & job\_237 \\
Baggage porters, bellhops, and concierges & job\_238 \\
Miscellaneous media and communication workers & job\_239 \\
Plating and coating machine setters, operators, and tenders, metal and plastic & job\_240 \\
Painters, Transportation Equipment & job\_241 \\
Printing Press Operators & job\_242 \\
Janitors and Cleaners Except Maids and Housekeeping Cleaners & job\_243 \\
Motion picture projectionists & job\_244 \\
Dietitians and nutritionists & job\_245 \\
Photographic process workers and processing machine operators & job\_246 \\
Television, video, and motion picture camera operators and editors & job\_247 \\
Roofers & job\_248 \\
Artists and related workers & job\_249 \\
Air traffic controllers and airfield operations specialists & job\_250 \\
Insurance sales agents & job\_251 \\
Etchers and engravers & job\_252 \\
Management analysts & job\_253 \\
Social workers & job\_254 \\
Geological and petroleum technicians & job\_255 \\
First-line supervisors/managers of correctional officers & job\_256 \\
Accountants and auditors & job\_257 \\
Textile bleaching and dyeing machine operators and tenders & job\_258 \\
Petroleum engineers & job\_259 \\
Construction and building inspectors & job\_260 \\
Locomotive engineers and operators & job\_261 \\
Cleaners of vehicles and equipment & job\_262 \\
Actuaries & job\_263 \\
Sewing machine operators & job\_264 \\
Hosts and hostesses, restaurant, lounge, and coffee shop & job\_265 \\
Nuclear technicians & job\_266 \\
Extruding and drawing machine setters, operators, and tenders, metal and plastic & job\_267 \\
Packers and packagers, hand & job\_268 \\
Fire inspectors & job\_269 \\
Fishers and related fishing workers & job\_270 \\
Diagnostic related technologists and technicians & job\_271 \\
Librarians & job\_272 \\
Helpers--installation, maintenance, and repair workers & job\_273 \\
Water and liquid waste treatment plant and system operators & job\_274 \\
Podiatrists & job\_275 \\
Market research analysts and marketing specialists & job\_276 \\
Veterinarians & job\_277 \\
Glaziers & job\_278 \\
Service station attendants & job\_279 \\
Machine feeders and offbearers & job\_280 \\
Railroad conductors and yardmasters & job\_281 \\
Boilermakers & job\_282 \\
File Clerks & job\_283 \\
Eligibility interviewers, government programs & job\_284 \\
Dental assistants & job\_285 \\
Announcers & job\_286 \\
Lawyers, Judges, magistrates, and other judicial workers & job\_287 \\
Aircraft mechanics and service technicians & job\_288 \\
Insulation workers & job\_289 \\
Shuttle car operators & job\_290 \\
Compliance Officers & job\_291 \\
Claims adjusters, appraisers, examiners, and investigators & job\_292 \\
Meter readers, utilities & job\_293 \\
Physician assistants & job\_294 \\
Court, municipal, and license clerks & job\_295 \\
Postal service mail sorters, processors, and processing machine operators & job\_296 \\
Ushers, lobby attendants, and ticket takers & job\_297 \\
Unemployed & job\_298 \\
Business Teachers Postsecondary & job\_299 \\
Postal service clerks & job\_300 \\
Dredge, excavating, and loading machine operators & job\_301 \\
Drafters & job\_302 \\
Molders, shapers, and casters, except metal and plastic & job\_303 \\
Child care workers & job\_304 \\
Models, demonstrators, and product promoters & job\_305 \\
Textile knitting and weaving machine setters, operators, and tenders & job\_306 \\
Taxi drivers and chauffeurs & job\_307 \\
Elementary and middle school teachers & job\_308 \\
Chemical engineers & job\_309 \\
Library assistants, clerical & job\_310 \\
Telemarketers & job\_311 \\
Designers & job\_312 \\
Biological scientists & job\_313 \\
Mining machine operators & job\_314 \\
Clergy & job\_315 \\
Agricultural inspectors & job\_316 \\
Architects, except naval & job\_317 \\
Special Code (occ1990dd=999) & job\_318 \\
Legislators & job\_319 \\
Mail clerks and mail machine operators, except postal service & job\_320 \\
Tailors, dressmakers, and sewers & job\_321 \\
Miscellaneous personal appearance workers & job\_322 \\
Cabinetmakers and bench carpenters & job\_323 \\
Switchboard operators, including answering service & job\_324 \\
Training and development specialists & job\_325 \\
Upholsterers & job\_326 \\
Dentists & job\_327 \\
Broadcast and sound engineering technicians and radio operators & job\_328 \\
Musicians, singers, and related workers & job\_329 \\
Chemical technicians & job\_330 \\
First-Line Supervisors of Landscaping and Lawn Service and Groundskeeping Workers & job\_331 \\
Health diagnosing and treating practitioners, all other & job\_332 \\
Shoe and leather workers and repairers & job\_333 \\
Food batchmakers & job\_334 \\
\legend{Job titles are randomly permuted and we assign ``numeric job titles'' from \texttt{job\_000} to \texttt{job\_334} to them following the random order.}
\end{longtable}
\normalsize

\section{Summary of Tokenized Datasets}
\label{sec:online-appendix-summary-tokenized-datasets}
This section provides a summary of tokenized text templates in various datasets. Different tokenizers could parse the same text into different numbers of tokens; all token counts reported in this section are from the Llama-2 tokenizer. We ignore the ``start of sentence'' token that the Llama-2 tokenizer adds to every paragraph.

\subsection{Training and Validation Splits}
We use the training and validation splits of our dataset to fine-tune large language models. To fine-tune the model, we use the text representations of workers' complete career histories $\texttemplate(x_{i, \leq T_i}, y_{i, \leq T_i})$ as the training and validation corpus, where each worker contributes a single paragraph of text in the corpus.
Table \ref{tab:training-and-validation-splits-num-tokens} summarizes the number of individuals, total number of tokens, and average number of tokens per individual in the training and validation splits for each data file in our replication material. Test splits are only used for inference and are reported in the next subsection.

\tiny
\begin{table}[H]
\caption{Number of tokens in the training and validation splits for fine-tuning using the Llama-2 tokenizer. \updated{12/05/2025, removed rows without birth year, no longer used.}}
\label{tab:training-and-validation-splits-num-tokens}
\centering
\resizebox{\textwidth}{!}{
\begin{tabular}{lcccccc}
\toprule
 & \multicolumn{2}{c}{Number of Individuals} & \multicolumn{2}{c}{Total Number of Tokens} & \multicolumn{2}{c}{Avg. Number of Tokens per Individual} \\
\multicolumn{1}{r}{\textbf{Split}}  & Training & Validation & Training & Validation & Training & Validation \\
\midrule
NLSY79 w/numerical job titles & 8,735 & 1,247 & 3,560,493 & 509,219 & 407.61 & 408.36 \\
NLSY79 w/birth years & 8,735 & 1,247 & 3,788,784 & 539,453 & 433.75 & 432.60 \\
NLSY97 w/numerical job titles & 6,288 & 898 & 2,176,948 & 309,988 & 346.21 & 345.20 \\
NLSY97 w/birth years & 6,288 & 898 & 2,189,953 & 313,107 & 348.27 & 348.67 \\
PSID81 w/numerical job titles & 21,739 & 3,105 & 5,132,422 & 737,382 & 236.09 & 237.48 \\
PSID81 w/birth years & 21,739 & 3,105 & 5,540,144 & 797,781 & 254.85 & 256.93 \\
\midrule
Mixture NLSY79, $10\%$ of other datasets & 9,608 & 1,247 & 4,032,323 & 539,453 & 419.68 & 432.60 \\
Mixture NLSY79, $20\%$ of other datasets & 10,482 & 1,247 & 4,273,244 & 539,453 & 407.67 & 432.60 \\
Mixture NLSY79, $30\%$ of other datasets  & 11,355 & 1,247 & 4,511,019 & 539,453 & 397.27 & 432.60 \\
Mixture NLSY79, $40\%$ of other datasets   & 12,229 & 1,247 & 4,759,164 & 539,453 & 389.17 & 432.60 \\
Mixture NLSY79, $50\%$ of other datasets & 13,102 & 1,247 & 4,999,403 & 539,453 & 381.58 & 432.60 \\
Mixture NLSY79, $60\%$ of other datasets & 13,976 & 1,247 & 5,235,267 & 539,453 & 374.59 & 432.60 \\
Mixture NLSY79, $70\%$ of other datasets & 14,849 & 1,247 & 5,475,725 & 539,453 & 368.76 & 432.60 \\
Mixture NLSY97, $10\%$ of other datasets & 6,916 & 898 & 2,388,943 & 313,107 & 345.42 & 348.67 \\
Mixture NLSY97, $20\%$ of other datasets & 7,545 & 898 & 2,582,108 & 313,107 & 342.23 & 348.67 \\
Mixture NLSY97, $30\%$ of other datasets  & 8,174 & 898 & 2,774,250 & 313,107 & 339.40 & 348.67 \\
Mixture NLSY97, $40\%$ of other datasets   & 8,803 & 898 & 2,959,053 & 313,107 & 336.14 & 348.67 \\
Mixture NLSY97, $50\%$ of other datasets    & 9,432 & 898 & 3,151,501 & 313,107 & 334.13 & 348.67 \\
Mixture NLSY97, $60\%$ of other datasets     & 10,060 & 898 & 3,345,202 & 313,107 & 332.53 & 348.67 \\
Mixture NLSY97, $70\%$ of other datasets      & 10,689 & 898 & 3,534,251 & 313,107 & 330.64 & 348.67 \\
Mixture PSID81, $10\%$ of other datasets & 23,912 & 3,105 & 6,397,322 & 797,781 & 267.54 & 256.93 \\
Mixture PSID81, $20\%$ of other datasets & 26,086 & 3,105 & 7,262,723 & 797,781 & 278.41 & 256.93 \\
Mixture PSID81, $30\%$ of other datasets & 28,260 & 3,105 & 8,125,846 & 797,781 & 287.54 & 256.93 \\
Mixture PSID81, $40\%$ of other datasets  & 30,434 & 3,105 & 8,994,806 & 797,781 & 295.55 & 256.93 \\
Mixture PSID81, $50\%$ of other datasets   & 32,608 & 3,105 & 9,867,246 & 797,781 & 302.60 & 256.93 \\
Mixture PSID81, $60\%$ of other datasets    & 34,782 & 3,105 & 10,728,379 & 797,781 & 308.45 & 256.93 \\
Mixture PSID81, $70\%$ of other datasets     & 36,762 & 3,105 & 11,518,881 & 797,781 & 313.34 & 256.93 \\
\midrule
Mixture $10\%$ of each dataset & 3,676 & 525 & 1,163,405 & 159,834 & 316.49 & 304.45 \\
Mixture $20\%$ of each dataset & 7,352 & 1,050 & 2,316,043 & 325,984 & 315.02 & 310.46 \\
Mixture $30\%$ of each dataset  & 11,028 & 1,575 & 3,443,564 & 489,417 & 312.26 & 310.74 \\
Mixture $40\%$ of each dataset   & 14,704 & 2,100 & 4,591,417 & 651,185 & 312.26 & 310.09 \\
Mixture $50\%$ of each dataset    & 18,381 & 2,625 & 5,745,202 & 820,063 & 312.56 & 312.40 \\
Mixture $60\%$ of each dataset     & 22,057 & 3,150 & 6,901,631 & 984,610 & 312.90 & 312.57 \\
Mixture $70\%$ of each dataset      & 25,733 & 3,675 & 8,060,265 & 1,148,317 & 313.23 & 312.47 \\
Mixture $80\%$ of each dataset       & 29,409 & 4,200 & 9,215,066 & 1,313,050 & 313.34 & 312.63 \\
Mixture $90\%$ of each dataset        & 33,085 & 4,725 & 10,357,526 & 1,482,894 & 313.06 & 313.84 \\
Mixture $100\%$ of each dataset         & 36,762 & 5,250 & 11,518,881 & 1,650,341 & 313.34 & 314.35 \\
\bottomrule
\end{tabular}
}
\legend{This table summarizes data files in our replication package used for fine-tuning FT-LABOR-LLM. Each row corresponds to two CSV files for the training and validation splits. For example, the second row reports the summary statistics of the training and validation splits of the NLSY79 dataset with numeric job titles, the file \texttt{NLSY79 w/numerical job titles} includes the training split of this data for fine-tuning FT-LABOR-LLMs.}
\end{table}
\normalsize

\subsection{Test Splits}
The test splits of our dataset are used to evaluate the performance of the fine-tuned language models. Each worker contributes multiple individual-year observations to the test set, where each observation is a data point in the test set. For each data point (i.e., each individual-year observation), Table \ref{tab:test-splits-num-tokens} provides the number of tokens in the test set, the number of workers, and the average number of tokens per worker for each data file. Please note that these numbers of tokens are counted by the Llama-2 tokenizer; using different tokenizers will result in different token counts. These files contain different variants of prompts of career histories, $\texttemplate(x_{i, \leq t}, y_{i, < t})$.

\tiny
\begin{table}[H]
\caption{Summary statistics for the number of tokens in prompts from the test splits by data file. \updated{12/05/2025, removed rows without birth year, no longer used.}}
\label{tab:test-splits-num-tokens}
\resizebox{\textwidth}{!}{
\begin{tabular}{lccccccccc}
\toprule
& \shortstack{Num\\ Transitions} & Total Tokens & \shortstack{Mean\\ Tokens} & \shortstack{Std\\ Tokens} & \shortstack{Min\\ Tokens} & \shortstack{1\%\\ Tokens} & \shortstack{50\%\\ Tokens} & \shortstack{99\%\\ Tokens} & \shortstack{Max\\ Tokens} \\
\midrule
NLSY79 w/numerical job titles & 51,593 & 13,947,376 & 270.33 & 124.12 & 83 & 86 & 255 & 537 & 568 \\
NLSY79 w/job titles & 51,593 & 181,340,841 & 3514.83 & 140.85 & 3,317 & 3,320 & 3,493 & 3,860 & 4,078 \\
NLSY79 w/birth years & 51,593 & 14,489,079 & 280.83 & 140.85 & 83 & 86 & 259 & 626 & 844 \\
NLSY79 w/birth years \& job titles & 51,593 & 181,959,957 & 3526.83 & 140.85 & 3,329 & 3,332 & 3,505 & 3,872 & 4,090 \\
NLSY97 w/numerical job titles & 29,951 & 6,703,708 & 223.82 & 87.60 & 89 & 89 & 219 & 401 & 414 \\
NLSY97 w/job titles & 29,951 & 103,401,303 & 3452.35 & 91.88 & 3,323 & 3,323 & 3,441 & 3,672 & 3,825 \\
NLSY97 w/birth years & 29,951 & 6,539,769 & 218.35 & 91.88 & 89 & 89 & 207 & 438 & 591 \\
NLSY97 w/birth years \& job titles & 29,951 & 103,760,715 & 3464.35 & 91.88 & 3,335 & 3,335 & 3,453 & 3,684 & 3,837 \\
PSID81 w/numerical job titles & 61,772 & 12,645,942 & 204.72 & 106.17 & 80 & 80 & 177 & 497 & 566 \\
PSID81 w/job titles & 61,772 & 213,349,188 & 3453.82 & 122.73 & 3,314 & 3,314 & 3,420 & 3,812 & 4,077 \\
PSID81 w/birth years & 61,772 & 13,578,540 & 219.82 & 122.73 & 80 & 80 & 186 & 578 & 843 \\
PSID81 w/birth years \& job titles & 61,772 & 214,090,452 & 3465.82 & 122.73 & 3,326 & 3,326 & 3,432 & 3,824 & 4,089 \\
\bottomrule
\end{tabular}
}
\legend{This table summarizes data files in our replication package used for evaluating FT-LABOR-LLM, which consists of the test split of each dataset. For example, the first row reports the summary statistics of test split of the NLSY79 dataset with numeric job titles, the file \texttt{NLSY79 w/numerical job titles} includes the test split of this dataset for evaluating language models.}
\end{table}
\normalsize

\section{Dataset Split Details for Fine-tuning}
\label{sec:online-appendix-summary-datasets}

This appendix reports the details of the training, validation, and test splits of each survey dataset used for fine-tuning. The training set consists of 70\% randomly selected individuals, the validation set consists of 20\% randomly selected individuals and the test set consists of 10\% randomly selected individuals. These three splits of individuals are non-overlapping. If an individual is selected into a specific split (e.g., the training set), all individual-year observations of this individual belong to that split.

Table \ref{tab:data-summary-by-split} shows the number of individual and individual-year observations in the training, validation, and test splits of each survey dataset. Table \ref{tab:number-of-transitions-by-transition-type-and-dataset} reports the number of individual-year observations in the training, validation, and test splits of each survey dataset that are the ``First Observation'' of an individual, an observation in which the individual transitions occupation (``Moving''), or an observation in which the individual stays in the same occupation (``Staying'').

\tiny
\begin{table}[H]
    \centering
    \caption{The number of individuals (the top number) and individual-year observations (the bottom number) in training, validation, and test splits of each survey dataset.
    \updated{12/05/2025, no change during revision, passed replication check.}
    }
    \label{tab:data-summary-by-split}
    \begin{tabular}{lccc}
    \toprule
    \multicolumn{1}{r}{\textbf{Dataset}}
 &    PSID81 & NLSY79 & NLSY97 \\
   \midrule
   Train Split (70\%) & 21,739 & 8,735 & 6,288 \\
    & 220,223 & 182,145 & 104,043 \\
   \midrule
   Validation Split (20\%) & 3,105 & 1,247 & 898 \\
    & 31,627 & 26,040 & 14,801 \\
   \midrule
   Test Split (10\%) & 6,212 & 2,497 & 1,798 \\
    & 61,772 & 51,593 & 29,951 \\
   \midrule
   Total
 Individuals    & 31,056 & 12,479 & 8,984 \\
 Transitions    & 313,622 & 259,778 & 148,795 \\
    \bottomrule
   \end{tabular}
\end{table}
\normalsize

\tiny
\begin{table}[H]
    \caption{The number of individual-year observations by transition type in training, validation, and test splits of each survey dataset.
    \updated{12/05/2025, no change during revision, passed replication check.}
    }
    \label{tab:number-of-transitions-by-transition-type-and-dataset}
    \centering
    \resizebox{\textwidth}{!}{
        \begin{tabular}{lccccccccc}
        \toprule
        \multicolumn{1}{r}{\textbf{Dataset}} & \multicolumn{3}{c}{PSID81} & \multicolumn{3}{c}{NLSY79} & \multicolumn{3}{c}{NLSY97} \\
        \multicolumn{1}{r}{\textbf{Split}} & Train & Validation & Test & Train & Validation & Test & Train & Validation & Test \\
        \midrule
        First Observation & 21,739 & 3,105 & 6,212 & 8,735 & 1,247 & 2,497 & 6,288 & 898 & 1,798 \\
        Moving  & 84,627 & 12,172 & 24,030 & 81,072 & 11,636 & 23,023 & 38,509 & 5,597 & 10,960 \\
        Staying & 113,857 & 16,350 & 31,530 & 92,338 & 13,157 & 26,073 & 59,246 & 8,306 & 17,193 \\
        \bottomrule
        \end{tabular}
    }
\end{table}
\normalsize

\section{Examples of Career Trajectories Generated by Off-the-Shelf LLMs}
\label{sec:online-appendix-llm-trajectory-examples}
LLMs gain knowledge about the labor market and hierarchy among different occupations from the pre-training data.
Therefore, off-the-shelf LLMs can generate plausible sequences of future occupations conditional on one's job history via appropriate prompt engineering.
These sections demonstrate a few examples of career trajectories generated by the off-the-shelf pre-trained Llama-2-7B model without fine-tuning.
In each example, we provide the model with the worker's career history and specify the desired data format through a prompt.
We use the Llama-2 model with full precision inference and \texttt{top\_k = 50, top\_p=0.6, temperature=1.0} to generate texts; we limit the maximum number of new tokens generated to be 50.

\subsection{Examples on an Individual with Low Education Level}
We use the following text as the prompt summarizing the individual's career history.
{
\footnotesize
\begin {Verbatim}[breaklines=true]
<A worker from the PSID dataset>
The following information is available about the work history of a female black or african american US worker residing in the south region.
The worker was born in 1963.
The worker has the following records of work experience, one entry per line, including year, education level, and the job title:
1979 (high school diploma): Cashiers
1980 (high school diploma): Not in labor force
1981 (high school diploma): Food servers, nonrestaurant
1982 (high school diploma): Food servers, nonrestaurant
1983 (high school diploma): Food servers, nonrestaurant
1984 (high school diploma):
\end{Verbatim}
}
The following shows three examples of completions given by the LLM. Note that the last line of generation is incomplete because we limit the number of new tokens generated to be 50.
{
\footnotesize
\begin {Verbatim}[breaklines=true]
===============================================================
LLM Generation 0
===============================================================
<A worker from the PSID dataset>
The following information is available about the work history of a female black or african american US worker residing in the south region.
The worker was born in 1963.
The worker has the following records of work experience, one entry per line, including year, education level, and the job title:
1979 (high school diploma): Cashiers
1980 (high school diploma): Not in labor force
1981 (high school diploma): Food servers, nonrestaurant
1982 (high school diploma): Food servers, nonrestaurant
1983 (high school diploma): Food servers, nonrestaurant
1984 (high school diploma): Food servers, nonrestaurant
1985 (high school diploma): Food servers, nonrestaurant
1986 (high school diploma): Food servers, nonrestaurant
1987
===============================================================
LLM Generation 1
===============================================================
<A worker from the PSID dataset>
The following information is available about the work history of a female black or african american US worker residing in the south region.
The worker was born in 1963.
The worker has the following records of work experience, one entry per line, including year, education level, and the job title:
1979 (high school diploma): Cashiers
1980 (high school diploma): Not in labor force
1981 (high school diploma): Food servers, nonrestaurant
1982 (high school diploma): Food servers, nonrestaurant
1983 (high school diploma): Food servers, nonrestaurant
1984 (high school diploma): Food servers, nonrestaurant
1985 (high school diploma): Food servers, nonrestaurant
1986 (high school diploma): Cashiers
1987 (high school di
===============================================================
LLM Generation 2
===============================================================
<A worker from the PSID dataset>
The following information is available about the work history of a female black or african american US worker residing in the south region.
The worker was born in 1963.
The worker has the following records of work experience, one entry per line, including year, education level, and the job title:
1979 (high school diploma): Cashiers
1980 (high school diploma): Not in labor force
1981 (high school diploma): Food servers, nonrestaurant
1982 (high school diploma): Food servers, nonrestaurant
1983 (high school diploma): Food servers, nonrestaurant
1984 (high school diploma): Cashiers
1985 (high school diploma): Food servers, nonrestaurant
1986 (high school diploma): Food servers, nonrestaurant
1987 (high school di
\end{Verbatim}
}

\subsection{Examples on an Individual with High Education Level}
The second example illustrates the model's capability on the high-education group. We use the following prompt text in this example:

{
\footnotesize
\begin {Verbatim}[breaklines=true]
<A worker from the PSID dataset>
The following information is available about the work history of a female black or african american US worker residing in the south region.
The worker was born in 1963.
The worker has the following records of work experience, one entry per line, including year, education level, and the job title:
1984 (college): Business Teachers Postsecondary
1985 (college): Postmasters and mail superintendents
1986 (graduate degree): Sales Representatives Services All Other
1987 (graduate degree): Wholesale and retail buyers, except farm products
1988 (graduate degree): Elementary and middle school teachers
1989 (graduate degree): Elementary and middle school teachers
1990 (graduate degree):
\end{Verbatim}
}
The following shows three completions generated by the LLM following the prompt above. Note that the last line of generation is incomplete because we limit the number of new tokens generated to be 50.
{
\footnotesize
\begin {Verbatim}[breaklines=true]
===============================================================
LLM Generation 0
===============================================================
<A worker from the PSID dataset>
The following information is available about the work history of a female black or african american US worker residing in the south region.
The worker was born in 1963.
The worker has the following records of work experience, one entry per line, including year, education level, and the job title:
1984 (college): Business Teachers Postsecondary
1985 (college): Postmasters and mail superintendents
1986 (graduate degree): Sales Representatives Services All Other
1987 (graduate degree): Wholesale and retail buyers, except farm products
1988 (graduate degree): Elementary and middle school teachers
1989 (graduate degree): Elementary and middle school teachers
1990 (graduate degree): Postsecondary teachers
1991 (graduate degree): Computer and information systems managers
1992 (graduate degree): Postsecondary teachers
1993 (graduate degree): First-line supervisors
===============================================================
LLM Generation 1
===============================================================
<A worker from the PSID dataset>
The following information is available about the work history of a female black or african american US worker residing in the south region.
The worker was born in 1963.
The worker has the following records of work experience, one entry per line, including year, education level, and the job title:
1984 (college): Business Teachers Postsecondary
1985 (college): Postmasters and mail superintendents
1986 (graduate degree): Sales Representatives Services All Other
1987 (graduate degree): Wholesale and retail buyers, except farm products
1988 (graduate degree): Elementary and middle school teachers
1989 (graduate degree): Elementary and middle school teachers
1990 (graduate degree): Elementary and middle school teachers
1991 (graduate degree): Secondary school teachers
1992 (graduate degree): Elementary and middle school teachers
1993 (graduate degree): Elementary and middle
===============================================================
LLM Generation 2
===============================================================
<A worker from the PSID dataset>
The following information is available about the work history of a female black or african american US worker residing in the south region.
The worker was born in 1963.
The worker has the following records of work experience, one entry per line, including year, education level, and the job title:
1984 (college): Business Teachers Postsecondary
1985 (college): Postmasters and mail superintendents
1986 (graduate degree): Sales Representatives Services All Other
1987 (graduate degree): Wholesale and retail buyers, except farm products
1988 (graduate degree): Elementary and middle school teachers
1989 (graduate degree): Elementary and middle school teachers
1990 (graduate degree): Secondary school teachers
1991 (graduate degree): Secondary school teachers
1992 (graduate degree): Postsecondary teachers
1993 (graduate degree): Business Teachers Postsecondary
\end{Verbatim}
}

\section{Example of CAREER Text Template Construction}
\label{sec:online-appendix-text-template-example}
This section provides an example of how we construct the text templates from the CAREER format dataset. For example, the following is the information of a sample individual in the PSID dataset.
{
\footnotesize \begin {Verbatim}[breaklines=true]
{'education': ['college', 'college', 'college', 'college', 'college', 'college',
               'college'],
 'ethnicity': ['white'],
 'gender': ['male'],
 'state': ['west'],
 'tok': ['56', '57', '274', 'unemployed', '57', '376', '376'],
 'year': ['2009', '2011', '2013', '2015', '2017', '2019', '2021'],
 'year_of_birth': ['1984']}
\end{Verbatim}
}
The individual is a male, white, born in 1984, and living in the west region. The individual holds a college degree throughout the period of observation.

The example individual above has seven years of observations (i.e., transitions) in 2009, 2011, 2013, 2015, 2017, 2019, and 2021, respectively.
The individual started with occupation \texttt{56} (corresponding to ``Industrial engineers, including health and safety'') with a college degree in 2009.

The individual switched to occupation \texttt{57} (corresponding to ``Mechanical engineers'') in 2011.
The individual switched to occupation \texttt{274} (``Sales Representatives Services All Other'') in 2013. Then, the individual became unemployed in 2015.

The text template for this individual is constructed as follows, where the first line indicates the survey dataset the individual came from.
The second line shows the demographic information of this individual.
The third line indicates the birth year of the individual.
The following seven lines show the work experience and education level of the individual, one for each transition.
Finally, the text representation concludes with \texttt{<END OF DATA>}.
{
\footnotesize \begin {Verbatim}[breaklines=true]
<A worker from the PSID dataset>
The following information is available about the work history of a male white US worker residing in the west region.
The worker was born in 1984.
The worker has the following records of work experience, one entry per line, including year, education level, and the job title:
2009 (college): Industrial engineers, including health and safety
2011 (college): Mechanical engineers
2013 (college): Sales Representatives Services All Other
2015 (college): Unemployed
2017 (college): Mechanical engineers
2019 (college): Loan interviewers and clerks
2021 (college): Loan interviewers and clerks
<END OF DATA>
\end{Verbatim}
}
If the individual belongs to the training split, we will use the text representation above to fine-tune the language model.
Otherwise, if the individual is in the test split, the subject contributes five individual-year observations (i.e., predictions) to the test set.
The following is the prompt for each individual-year observation and the ground label for the prediction.
{
\footnotesize
\begin {Verbatim}[breaklines=true]
>>>>>>>>>> Prompt of Transition 1 (i.e., individual-year observation):
<A worker from the PSID dataset>
The following information is available about the work history of a male white US worker residing in the west region.
The worker was born in 1984.
The worker has the following records of work experience, one entry per line, including year, education level, and the job title:
2009 (college):
>>>>>>>>>> Label of Transition 1 (i.e., individual-year observation):
Industrial engineers, including health and safety
>>>>>>>>>> Prompt of Transition 2 (i.e., individual-year observation):
<A worker from the PSID dataset>
The following information is available about the work history of a male white US worker residing in the west region.
The worker was born in 1984.
The worker has the following records of work experience, one entry per line, including year, education level, and the job title:
2009 (college): Industrial engineers, including health and safety
2011 (college):
>>>>>>>>>> Label of Transition 2 (i.e., individual-year observation):
Mechanical engineers
>>>>>>>>>> Prompt of Transition 3 (i.e., individual-year observation):
<A worker from the PSID dataset>
The following information is available about the work history of a male white US worker residing in the west region.
The worker was born in 1984.
The worker has the following records of work experience, one entry per line, including year, education level, and the job title:
2009 (college): Industrial engineers, including health and safety
2011 (college): Mechanical engineers
2013 (college):
>>>>>>>>>> Label of Transition 3 (i.e., individual-year observation):
Sales Representatives Services All Other
>>>>>>>>>> Prompt of Transition 4 (i.e., individual-year observation):
<A worker from the PSID dataset>
The following information is available about the work history of a male white US worker residing in the west region.
The worker was born in 1984.
The worker has the following records of work experience, one entry per line, including year, education level, and the job title:
2009 (college): Industrial engineers, including health and safety
2011 (college): Mechanical engineers
2013 (college): Sales Representatives Services All Other
2015 (college):
>>>>>>>>>> Label of Transition 4 (i.e., individual-year observation):
Unemployed
>>>>>>>>>> Prompt of Transition 5 (i.e., individual-year observation):
<A worker from the PSID dataset>
The following information is available about the work history of a male white US worker residing in the west region.
The worker was born in 1984.
The worker has the following records of work experience, one entry per line, including year, education level, and the job title:
2009 (college): Industrial engineers, including health and safety
2011 (college): Mechanical engineers
2013 (college): Sales Representatives Services All Other
2015 (college): Unemployed
2017 (college):
>>>>>>>>>> Label of Transition 5 (i.e., individual-year observation):
Mechanical engineers
>>>>>>>>>> Prompt of Transition 6 (i.e., individual-year observation):
<A worker from the PSID dataset>
The following information is available about the work history of a male white US worker residing in the west region.
The worker was born in 1984.
The worker has the following records of work experience, one entry per line, including year, education level, and the job title:
2009 (college): Industrial engineers, including health and safety
2011 (college): Mechanical engineers
2013 (college): Sales Representatives Services All Other
2015 (college): Unemployed
2017 (college): Mechanical engineers
2019 (college):
>>>>>>>>>> Label of Transition 6 (i.e., individual-year observation):
Loan interviewers and clerks
>>>>>>>>>> Prompt of Transition 7 (i.e., individual-year observation):
<A worker from the PSID dataset>
The following information is available about the work history of a male white US worker residing in the west region.
The worker was born in 1984.
The worker has the following records of work experience, one entry per line, including year, education level, and the job title:
2009 (college): Industrial engineers, including health and safety
2011 (college): Mechanical engineers
2013 (college): Sales Representatives Services All Other
2015 (college): Unemployed
2017 (college): Mechanical engineers
2019 (college): Loan interviewers and clerks
2021 (college):
>>>>>>>>>> Label of Transition 7 (i.e., individual-year observation):
Loan interviewers and clerks
\end{Verbatim}
}

Readers can refer to our \texttt{summary\_statistics.ipynb} notebook in the replication package to explore more examples.

\subsection{Template with Numerical Job Titles}
\label{sec:online-appendix-numerical-job-titles}

In the main paper, we use a version of the text template that represents career trajectories with numerical job titles.
Instead of using the actual job title such as \texttt{Cashiers}, the numerical template uses job titles like \texttt{job\_144}. Here is an example:
{
\footnotesize
\begin{Verbatim}[breaklines=true]
<A worker from the PSID dataset>
The following information is available about the work history of a female white US worker residing in the west region.
The worker was born in 1985.
The worker has the following records of work experience, one entry per line, including year, education level, and the job title:
2007 (college): job_144
2009 (college): job_169
2011 (college): job_089
2013 (college): job_304
2015 (college): job_304
2017 (college): job_304
2021 (college): job_169
<END OF DATA>
\end{Verbatim}
}

\section{Details of the Transformers and CAREER Model}
\label{sec:online-appendix-career}

\subsection{Details of the (Decoder-Only) Transformers}
There are several types of transformer models, including encoder-only transformers (e.g., BERT for embedding text), decoder-only transformers (e.g., GPT for generating text), and encoder-decoder transformers (e.g., the original transformer architecture for machine translation).
We focus on the decoder-only transformer in this paper because it fits our application of predicting future information in sequences using only past information in the same sequence.
In the main text, we view the decoder-only transformer as a high-dimensional multinomial logit model for predicting the next token. Given a sequence of tokens $v_{1:J} = (v_1, \ldots, v_J)$ with each token $v_j$ in the vocabulary $\mathcal{V}$, to predict the token at position $v_{j+1}$, the model first maps the history $v_{1:j}$ into an embedding
\[
z_j = \mathcal{E}_\text{tsf}(v_{1:j}) \in \mathbb{R}^{d_\text{tsf}}.
\]
This embedding is a learned summary of the entire past sequence.
Conditional on $z_j$, the probability that the next token equals $v \in \mathcal{V}$ is
\begin{equation}
\hat{P}^{\mathcal{V}}_{\text{Transformer}}(v_{j+1}=v \mid v_{1:j})
= \frac{\exp\!\big(z_j^\top \hat{\beta}_v\big)}
       {\sum_{v' \in \mathcal{V}} \exp\!\big(z_j^\top \hat{\beta}_{v'}\big)},
\label{eq:tsf-logit-main}
\end{equation}
where $\hat{\beta}_v \in \mathbb{R}^{d_\text{tsf}}$ is the coefficient vector associated with token $v$. This has the same form as a multinomial logit
with covariates $z_j$ and alternative-specific coefficients $\{\hat{\beta}_v\}_{v \in \mathcal{V}}$.
This multinomial logit model is often referred to as the ``language model prediction head'' in a transformer model.
The key difference is that the covariates $z_j$ are learned flexible functions of the entire history $v_{1:j}$ rather than hand-specified lagged variables like in traditional multinomial logistic regression models.

We now spell out how the embedding $\mathcal{E}_\text{tsf}(v_{1:j})$ is constructed using a standard transformer architecture. Consider again a
token sequence $v_{1:J} = (v_1, \ldots, v_J)$ with $v_j \in \mathcal{V}$.
Each token is first mapped to a embedding $h^{(0)}_j \in \mathbb{R}^{d_\text{tsf}}$ by a token embedding together with a positional
encoding that captures the position of the token in the sequence, called ``positional encoding'' in the transformer literature.
A stack of $L$ transformer layers then iteratively refines these embeddings.
In layer $\ell \in \{0,\ldots,L-1\}$, self-attention computes,
for each position $j$, attention weights on all earlier positions $j' \le j$ to form a weighted average of embeddings of past tokens,
\begin{equation}
\pi_{j,j'}^{(\ell)}
= \frac{\exp\!\big((h_j^{(\ell)})^\top W^{(\ell)} h_{j'}^{(\ell)}\big)}
       {\sum_{k=1}^{j} \exp\!\big((h_j^{(\ell)})^\top W^{(\ell)} h_{k}^{(\ell)}\big)},
\label{eq:attn-weights}
\end{equation}
where $W^{(\ell)} \in \mathbb{R}^{d_\text{tsf} \times d_\text{tsf}}$ is a trainable weight matrix.
Note that this is a simplified formulation of the original multi-head attention mechanism in \citet{vaswani_attention_2017}, which uses several sets of different projection matrices for queries, keys, and values.
Using these attention weights, the layer forms a weighted average of past embeddings,
\begin{equation}
\tilde{h}_j^{(\ell)}
= h_j^{(\ell)} + \sum_{j'=1}^{j} \pi_{j,j'}^{(\ell)}\, h_{j'}^{(\ell)},
\label{eq:attn-agg}
\end{equation}
and then applies a position-wise feed-forward network to produce the refined embedding for the next layer $\ell+1$,
\begin{equation}
h_j^{(\ell+1)} = \operatorname{FFN}^{(\ell)}\!\big(\tilde{h}_j^{(\ell)}\big).
\label{eq:ffn}
\end{equation}
After $L$ layers, we obtain the final representation $h_j^{(L)}$, and we identify this with the embedding $\mathcal{E}_\text{tsf}(v_{1:j})$ used in the main text,
\begin{equation}
\mathcal{E}_\text{tsf}(v_{1:j}) = h_j^{(L)} \in \mathbb{R}^{d_\text{tsf}}.
\label{eq:embedding-def}
\end{equation}

A linear language-model head and softmax then map $z_j = \mathcal{E}_\text{tsf}(v_{1:j})$ to a predictive probability distribution over the vocabulary $\mathcal{V}$. Let
$w_{\mathrm{LM}} \in \mathbb{R}^{|\mathcal{V}| \times d_\text{tsf}}$ and $b \in \mathbb{R}^{|\mathcal{V}|}$ denote the parameters of this output layer. The model-implied probability that the next token equals $v \in \mathcal{V}$ is
\begin{equation}
\hat{P}^{\mathcal{V}}_{\text{Transformer}}(v_{j+1}=v \mid v_{1:j})
= \frac{\exp\!\big( (\hat{w}_{\mathrm{LM}} z_j + \hat{b})_v \big)}
       {\sum_{v' \in \mathcal{V}} \exp\!\big( (\hat{w}_{\mathrm{LM}} z_j + \hat{b})_{v'} \big)}.
\label{eq:softmax-matrix}
\end{equation}

The parameters of the transformer layers $\{W^{(\ell)}, \operatorname{FFN}^{(\ell)}\}_{\ell=0}^{L-1}$ and the output coefficients $\{\hat{\beta}_v\}_{v \in \mathcal{V}}$ (equivalently, $\hat{w}_{\mathrm{LM}}$ and $\hat{b}$) are estimated jointly by maximizing the log-likelihood of the observed sequences under this next-token prediction model.

\subsection{Details of the CAREER Model}
The CAREER model leverages a large-scale dataset of online resumes, which covers 24 million workers. The data is passively collected from online resume platforms, ensuring a broad and diverse representation of career paths across various industries and job roles.

\paragraph*{Resume Dataset for Pre-training} We use the exact same data processing as \citet{vafa_career_2024} to construct the resume dataset for pre-training CAREER.\footnote{Readers can refer to Appendix F in \citet{vafa_career_2024} for additional details on dataset construction.}
First, we convert each resume in the dataset into a chronological sequence of entries with the occupation (Standard Occupational Code, or SOC), starting year, and ending year. If an individual worked in multiple occupations in a single year (i.e., there are overlapping records on the resume), we select the one in which they spent the most time; in cases of equal time spent, we choose the occupation that started earlier in their career. We convert the SOC codes to \texttt{occ1990dd} codes using a crosswalk from \href{https://forum.ipums.org/t/occ1990-to-soc-crosswalk/2098}{Destin Royer} to match the occupation codes in our survey datasets. The survey datasets also distinguish between non-employed categories (unemployed, out of labor force, or student), but these categories were absent in the resumes data. When the year associated with an occupation was missing, we exclude it from the dataset as we cannot determine its position in an individual's career timeline. We link each occupation to the individual's most recent educational degree, categorized into one of eight groups: high school diploma, some college, bachelor's degree, graduate degree, certificate, license, and diploma. The original CAREER model was not designed to handle birth year informaiton, we futher modify the CAREER model to handle birth year information so that we can compare our models to CAREER with the same input features.

In addition to the dynamic variables, we use two static variables imputed based off other data in
the resume: gender and location.
Locations are classified into the 50 U.S. states, Puerto Rico, Washington D.C., and an ``unknown'' category for cases where the location could not be imputed; however, we grouped states into four regions (northeast, north central, south, west), with a fifth ``other'' region for Puerto Rico and missing states, to match the survey datasets. We replaced any missing values for these static variables with a special ``missing'' token.

 This pre-processing results in a dataset containing 24 million resumes and 245 million individual-year observations (i.e., transitions).

\paragraph*{Survey Datasets for Fine-Tuning}
We construct our own copies of the survey datasets to fine-tune CAREER models in this paper. Appendix \ref{sec:appendix-data} provides the details on how we process our survey datasets.

\paragraph*{Model Estimation}
The estimation procedure of CAREER consists of two stages: (1) pre-training using resume datasets, and (2) fine-tuning using survey datasets.
We use \href{https://github.com/keyonvafa/career-code}{CAREER's official repository} for model estimation; a copy of the repository has been included in our replication material.

The pre-training uses the Adam optimizer with $\beta$ parameters (0.9, 0.98), weight decay of 0.01, and no gradient clipping. The learning rate starts at $10^{-7}$ with a scheduler that follows an inverse square root decay, warming up over 4,000 updates to a peak of 0.0005. Training samples have a maximum token length of 512, using end-of-sequence (EOS) tokens to define breaks. Each batch contains up to 16,000 tokens, with updates performed every batch, targeting 85,000 updates in total. Model checkpoints are saved every 1,000 updates to a specified directory, and the model checkpoint with the best validation loss is recorded.
We fine-tune the pre-trained model checkpoints with the lowest validation loss using a survey dataset. The Adam optimizer is used with $\beta$ parameters (0.9, 0.98), a weight decay of 0.01, and no gradient clipping. The learning rate starts at $10^{-7}$ and warms up over 500 updates to 0.0001, following an inverse square root decay scheduler. Each sample contains a maximum of 512 tokens, with end-of-sequence (EOS) tokens used for defining breaks, and each batch can include up to 16,000 tokens. When fine-tuning the survey dataset, we train the model until it overfits according to the validation loss. Finally, we evaluate the model performance using the checkpoint with the best validation performance.
Both the pre-training and fine-tuning use mixed precision (FP16) for computational efficiency.

The model estimation pipeline is performed for each survey dataset separately with the random seed fixed.

\section{Additional Results for Improving Off-the-Shelf LLMs using Prompt Engineering}
\label{sec:online-appendix-detailed-ots}

\subsection{Improving Off-the-Shelf LLM Perplexity using Prompting Strategies}
Table \ref{tab:main-text-baseline-llm-performances} shows that off-the-shelf pre-trained LLMs perform worse at predicting next occupations compared to the state-of-the-art CAREER model. In this section, we show that we can improve their performance by adding additional information into the prompt to facilitate in-context learning.
We explore two types of information: (1) the list of job titles and (2) additional resume examples from other workers.  A limiting factor in our ability to use such prompting strategies is the maximum context length of the models. For most models, we cannot include both the full list of job titles and example resumes.

\paragraph*{Job Titles in the Prompt} We prepend the list of all 335 job titles, one per line, to the text representation of career history $\texttemplate(x_{i, \leq t},y_{i, < t})$, which informs the off-the-shelf model about the prediction space.
With this modification, the prompt passed into the LLM becomes $\text{[List of Job Titles]} \oplus \texttemplate(x_{i, \leq t},y_{i, < t})$, where $\oplus$ denotes string concatenation.

\paragraph*{Example Resumes in the Prompt} We prepend example resumes randomly sampled (without replacement) from workers in the training set to the text representation of career history $\texttemplate(x_{i, \leq t},y_{i, < t})$, which informs the off-the-shelf model about our data structure.
The prompt fed into the model becomes $\texttemplate(x_{j_1, \leq T_{j_1}},y_{j_1, \leq T_{j_1}}) \oplus \cdots \oplus \texttemplate(x_{j_K, \leq T_{j_K}},y_{j_K, \leq T_{j_K}}) \oplus \texttemplate(x_{i, \leq t},y_{i, < t})$ if we add $K$ individuals $j_1,\cdots,j_K$ where $\texttemplate(x_{j, \leq T_j},y_{j, \leq T_j})$ means the complete resume for individual $j$.

Since the main models we study in this paper, Llama-2-7B and Llama-2-13B, only have enough context length for either job titles or a few examples resumes (both have 4k context length; in our dataset, one resume is up to 900 tokens, and the list of job titles is more than 3,200 tokens long, so we cannot include even one resume in combination with all job titles. Online Appendix \ref{sec:online-appendix-summary-tokenized-datasets} provides more details on the token counts of prompts in our datasets.), we study the open-sourced \href{https://huggingface.co/togethercomputer/LLaMA-2-7B-32K}{Llama-2-7B-32k} model provided by Together AI, the Llama-3.1-8B model (with a 128k context window), and the Llama-3.2-1B/3B model (with a 128k context window) to assess the benefits of combining the two prompting approaches. These models with longer context windows allow us to fit significantly more example resumes in our prompt.
The average length of prompts in our experiments is much longer than the $\texttemplate$ representation of career history we use in the previous section, leading to a significant increase in the computational cost of processing each prompt. As a result, we randomly sample 10\% of workers from the test set of each survey dataset in this exercise.

Table \ref{tab:icl-off-the-shelf-perplexity} presents the model perplexities when adding one, three, five, or ten example resumes with and without adding the list of job titles.
When we use ten example resumes and job titles simultaneously, the best-performing model reduces perplexity by a factor of 10 to 20, depending on the dataset. However, this approach to occupation modeling is still substantially worse than that of CAREER. We also observe that adding ten example resumes to the prompt reduces perplexity more than adding job titles for all models in Table \ref{tab:icl-off-the-shelf-perplexity}. Moreover, the inclusion of job titles in the prompt performs as well or better than the inclusion of up to three to five resumes for all models.

\tiny
\begin{longtable}{lccccc}
    \caption{Test-set perplexity for off-the-shelf models with in-context learning examples and/or job titles.
    \updated{12/05/2025, no change during revision, passed replication check.}
    }
    \label{tab:icl-off-the-shelf-perplexity} \\
    \toprule
    \multicolumn{2}{r}{\textbf{Evaluation Dataset}} & PSID81 & NLSY79 & NLSY97 \\
    \multicolumn{2}{r}{\textbf{Number of Test Set Transitions}} & 6,177 & 5,159 & 2,995 \\
    \midrule
    \endfirsthead
    \caption[]{(continued)} \\
    \toprule
    \multicolumn{2}{r}{\textbf{Evaluation Dataset}} & PSID81 & NLSY79 & NLSY97 \\
    \multicolumn{2}{r}{\textbf{Number of Test Set Transitions}} & 6,177 & 5,159 & 2,995 \\
    \midrule
    \endhead
    \midrule
    \multicolumn{5}{r}{\textit{Continued on next page}} \\
    \endfoot
    \bottomrule
    \multicolumn{5}{p{0.9\textwidth}}{\footnotesize Perplexity on a 10\% random sample of the test set, with test-set-bootstrap standard errors in parentheses.} \\
    \endlastfoot
    \textbf{Models Without Job Titles in Prompt} & \textbf{\# Resumes} & & & \\
    \midrule
    OTS Llama-2-13B & 0 & 137.685 (10.699) & 122.732 (9.676) & 107.080 (10.788) \\
    OTS Llama-2-13B & 1 & 49.297 (3.430) & 44.805 (3.352) & 25.168 (2.480) \\
    OTS Llama-2-13B & 3 & 35.988 (2.267) & 27.697 (1.785) & 18.907 (1.772) \\
    OTS Llama-2-7B-32k & 0 & 241.044 (22.812) & 182.748 (16.373) & 173.942 (22.880) \\
    OTS Llama-2-7B-32k & 1 & 81.780 (6.048) & 65.448 (4.990) & 34.827 (3.844) \\
    OTS Llama-2-7B-32k & 3 & 53.498 (3.561) & 38.249 (2.539) & 24.857 (2.634) \\
    OTS Llama-2-7B-32k & 5 & 45.269 (2.753) & 31.642 (1.993) & 21.876 (2.153) \\
    OTS Llama-2-7B-32k & 10 & 36.533 (2.131) & 26.203 (1.495) & 17.518 (1.510) \\
    OTS Llama-2-7B & 0 & 356.329 (27.380) & 293.284 (21.387) & 252.700 (27.979) \\
    OTS Llama-2-7B & 1 & 60.961 (4.290) & 48.847 (3.467) & 28.134 (2.924) \\
    OTS Llama-2-7B & 3 & 40.197 (2.532) & 29.365 (1.818) & 20.178 (2.016) \\
    OTS Llama-3.1-8B & 0 & 127.788 (10.564) & 110.871 (8.973) & 99.156 (11.408) \\
    OTS Llama-3.1-8B & 1 & 53.386 (3.744) & 43.275 (3.013) & 25.438 (2.346) \\
    OTS Llama-3.1-8B & 3 & 35.287 (2.173) & 26.445 (1.567) & 17.935 (1.569) \\
    OTS Llama-3.1-8B & 5 & 30.073 (1.725) & 22.435 (1.246) & 16.107 (1.324) \\
    OTS Llama-3.1-8B & 10 & 25.083 (1.385) & 19.409 (1.009) & 13.675 (1.034) \\
    OTS Llama-3.2-1B & 0 & 456.088 (51.012) & 371.333 (38.769) & 277.735 (40.961) \\
    OTS Llama-3.2-1B & 1 & 165.556 (15.246) & 133.295 (12.842) & 72.933 (10.011) \\
    OTS Llama-3.2-1B & 3 & 92.494 (7.515) & 62.271 (5.065) & 41.714 (5.218) \\
    OTS Llama-3.2-1B & 5 & 71.802 (5.350) & 47.563 (3.620) & 34.383 (4.023) \\
    OTS Llama-3.2-1B & 10 & 52.904 (3.740) & 36.043 (2.409) & 24.988 (2.631) \\
    OTS Llama-3.2-3B & 0 & 165.109 (14.493) & 134.391 (11.186) & 122.578 (14.671) \\
    OTS Llama-3.2-3B & 1 & 64.356 (4.575) & 54.938 (3.970) & 31.223 (3.152) \\
    OTS Llama-3.2-3B & 3 & 44.058 (2.808) & 33.635 (2.156) & 22.268 (2.125) \\
    OTS Llama-3.2-3B & 5 & 37.316 (2.236) & 28.049 (1.729) & 19.891 (1.815) \\
    OTS Llama-3.2-3B & 10 & 29.925 (1.726) & 22.953 (1.306) & 16.213 (1.334) \\
    \midrule
    \textbf{Models With Job Titles in Prompt} & \textbf{\# Resumes}  & & & \\
    \midrule
    OTS Llama-2-13B & 0 & 33.354 (1.913) & 33.779 (1.935) & 28.345 (1.987) \\
    OTS Llama-2-7B-32k & 0 & 42.005 (2.522) & 45.718 (2.678) & 47.952 (4.127) \\
    OTS Llama-2-7B-32k & 1 & 28.276 (1.459) & 26.037 (1.225) & 16.247 (1.118) \\
    OTS Llama-2-7B-32k & 3 & 24.001 (1.128) & 20.784 (0.868) & 13.522 (0.897) \\
    OTS Llama-2-7B-32k & 5 & 22.570 (1.046) & 19.579 (0.822) & 12.737 (0.839) \\
    OTS Llama-2-7B-32k & 10 & 20.735 (0.918) & 18.038 (0.732) & 11.741 (0.736) \\
    OTS Llama-2-7B & 0 & 36.909 (2.135) & 33.138 (1.760) & 31.456 (2.400) \\
    OTS Llama-3.1-8B & 0 & 30.850 (1.633) & 26.984 (1.309) & 21.910 (1.394) \\
    OTS Llama-3.1-8B & 1 & 22.120 (1.102) & 20.426 (0.921) & 13.897 (0.912) \\
    OTS Llama-3.1-8B & 3 & 19.167 (0.912) & 16.954 (0.726) & 11.858 (0.769) \\
    OTS Llama-3.1-8B & 5 & 17.863 (0.828) & 16.022 (0.676) & 11.349 (0.742) \\
    OTS Llama-3.1-8B & 10 & 16.445 (0.763) & 15.202 (0.631) & 10.494 (0.672) \\
    OTS Llama-3.2-1B & 0 & 62.234 (3.885) & 53.313 (3.068) & 45.252 (3.518) \\
    OTS Llama-3.2-1B & 1 & 35.440 (1.880) & 31.672 (1.663) & 20.850 (1.630) \\
    OTS Llama-3.2-1B & 3 & 28.715 (1.431) & 24.561 (1.163) & 17.001 (1.248) \\
    OTS Llama-3.2-1B & 5 & 26.034 (1.280) & 22.695 (1.057) & 15.980 (1.155) \\
    OTS Llama-3.2-1B & 10 & 22.947 (1.130) & 20.254 (0.913) & 14.019 (0.990) \\
    OTS Llama-3.2-3B & 0 & 39.811 (2.199) & 39.236 (2.227) & 35.443 (2.700) \\
    OTS Llama-3.2-3B & 1 & 24.777 (1.204) & 23.276 (1.091) & 14.840 (0.987) \\
    OTS Llama-3.2-3B & 3 & 20.814 (0.983) & 18.662 (0.810) & 12.880 (0.856) \\
    OTS Llama-3.2-3B & 5 & 19.561 (0.926) & 17.400 (0.750) & 12.380 (0.827) \\
    OTS Llama-3.2-3B & 10 & 17.806 (0.824) & 16.391 (0.683) & 11.516 (0.749) \\
\end{longtable}
\normalsize

\subsection{Likelihood of Generating Valid Job Titles}
\label{sec:validjobtitles}
\begin{figure}[H]
    \centering
    \includegraphics[width=\linewidth]{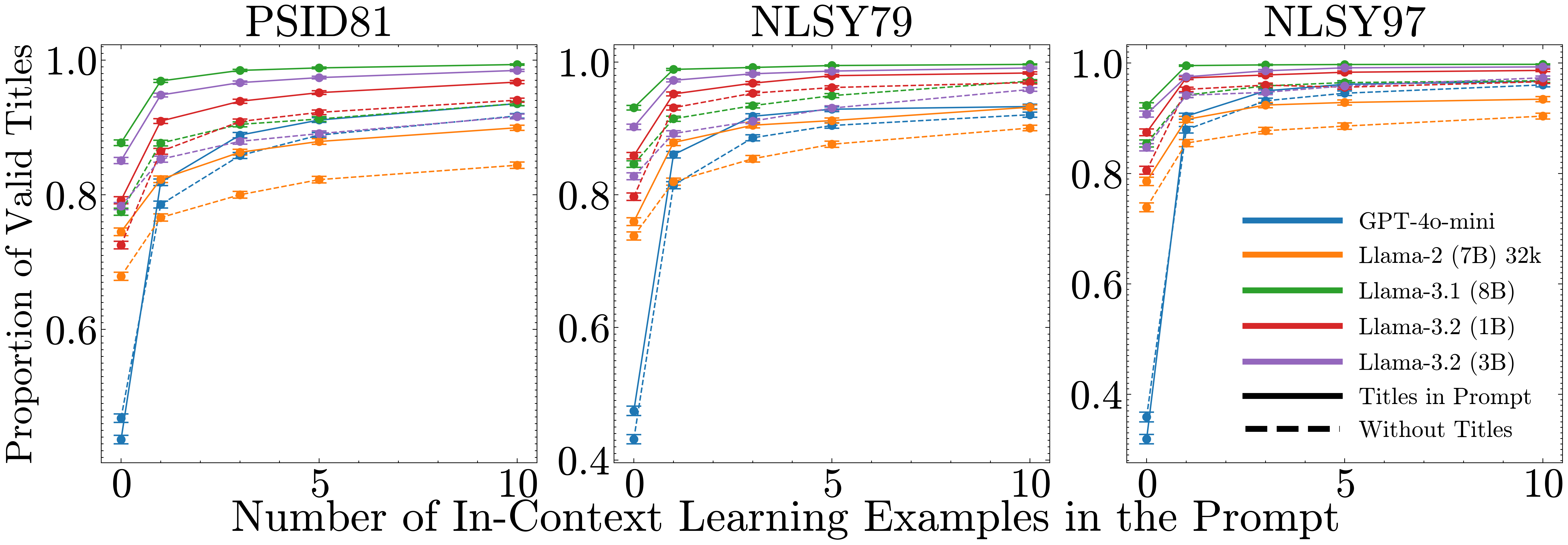}
    \caption{Likelihoods of generating valid job titles given different numbers of in-context learning examples (resumes) and job titles in the prompt for off-the-shelf LLMs. \updated{12/05/2025, no change during revision, passed replication check.}}
    \label{fig:prop-valid-titles}
\end{figure}

As mentioned in Section \ref{sec:llms}, we can feed a LLM with a prompt and repeatedly sample from the LLM's output distribution to generate a sequence of tokens as the continuation of the prompt.
Specifically, in the settings considered in the last subsection, we assess whether the model generates a continuation that starts with a valid job title:
\begin{align}
    \exists y \in \mathcal{Y} \text{ s.t., } \text{LLM.generate(prompt)}.\text{startswith}(\jobtitle(y))
    \label{eq:valid-title}
\end{align}
Figure \ref{fig:prop-valid-titles} summarizes the empirical probability that off-the-shelf Llama models generate valid job titles (i.e., the event in Equation (\ref{eq:valid-title}) occurs) on a 10\% sub-sample of the test set, where the figure illustrates how the results vary with different prompting strategies. We find that the probabilities range from 0.68 to greater than 0.99, the latter performance obtained from combining job titles and example resumes in the prompt.

We then conduct the same exercise using the \texttt{gpt-4o-mini-2024-07-18} model provided by OpenAI.\footnote{We use OpenAI's chat completion batch API to generate those continuations. We set the temperature to be 1, the seed to be 42, the maximum number of generated tokens to be 20, and the stop word to be the new line symbol (i.e., ``$\backslash$\texttt{n}'') for these generations.}  As illustrated in Figure \ref{fig:prop-valid-titles}, the OpenAI patterns and results are similar to those of the Llama models, with slightly larger probabilities of correct job titles and a maximum of 0.97 on the NLSY97 dataset with both job titles and ten sample resumes included in the prompt.

\section{Details on Full-Precision versus Quantized Models}
\label{sec:online-appendix-full-prec-quant}
Model quantization is a technique for improving models' computational efficiency and decreasing memory usage by reducing the numerical precision of model parameters (e.g., from 32-bit to 8-bit or 4-bit).
Existing research has shown that LLMs with quantization can achieve similar performance to full-precision models \citet{dettmers_qlora_2023}.
We fine-tune the Llama-2-7B model under full precision using Together AI's platform, but we quantize model weights to 8-bit before conducting experiments for LLM inference in the main paper to save computational resources.

In this appendix, we compare the performance of the full-precision and 8-bit quantization versions of the FT-7B.
Specifically, we take the FT-7B that was fine-tuned under full precision; then, we query predicted probabilities of future job titles using the two variants of the fine-tuned model, one in full precision and the other quantized to 8-bit.
Table \ref{tab:full-precision-and-quantized-llama-2-7b} compares models' performance on different datasets. These results suggest no significant difference between the full-precision and quantized models in terms of predictive performance.

It is extremely challenging for an individual researcher to obtain the hardware for full-precision fine-tuning (e.g., >112GiB of GPU memory for 7B). Fine-tuning on quantized models would require additional tricks like LoRA because one cannot run backpropagation on quantized parameters directly. Different LoRA techniques lead to different model performance, but exploration of these techniques is beyond the scope of this paper. We highly recommend that researchers outsource model fine-tuning to a third party due to the engineering complexity (e.g., training on multiple GPUs). We quantize the model during inference to speed up the inference and save GPU memory.

\tiny
\begin{table}[H]
\centering
\caption{Test-set perplexity of full-precision versus quantized (8-bit) FT-7B. \updated{12/05/2025, no change during revision, passed replication check.}}
\label{tab:full-precision-and-quantized-llama-2-7b}
\begin{tabular}{llll}
\toprule
\multicolumn{1}{r}{\textbf{Evaluation Dataset}} & PSID81 & NLSY79 & NLSY97 \\
\multicolumn{1}{r}{\textbf{Number of Test Set Transitions}} &  61,772 & 51,593 & 29,951 \\
\midrule
FT-7B 8-bit Quantized Inference & 8.184 (0.126) & 8.329 (0.147) & 6.350 (0.101) \\
FT-7B Full Precision Inference & 8.161 (0.126) & 8.309 (0.147) & 6.337 (0.100) \\
\bottomrule
\end{tabular}
\legend{FT-7B was fine-tuned using full precision. Test-set-bootstrap standard errors are in parentheses. Prompts of LLMs include birth year information in this table.}
\end{table}
\normalsize

\section{Model Performance by Different Education Groups}
\label{sec:online-appendix-edu-group}

In this appendix, we explore how models perform on different subgroups defined by educational backgrounds to evaluate whether the main results of our paper are consistent across subpopulations. First, Table \ref{tab:ppl-diff-edu-group} presents the perplexity differences between FT-7B, FT-13B, and CAREER on different subgroups and datasets. Specifically, we group individual-year observations $(i, t)$ based on education level, then compare perplexities of FT-LABOR-LLM and CAREER on these subsets of observations separately. Note that education level can change throughout an individual's career history so different observations of the same individual can belong to different education subgroups.
Table \ref{tab:ppl-diff-edu-group} indicates that our language-based model consistently outperforms the previous state-of-the-art model on sub-groups with sufficient observations.

\tiny
\begin{table}[hbtp]
    \centering
    \caption{Test-set perplexity by different education groups.\updated{12/05/2025, passed replication check.}}
    \label{tab:ppl-diff-edu-group}
    \resizebox{\textwidth}{!}{
    \begin{tabular}{lccc}
    \toprule
    Dataset & PSID81 & NLSY79 & NLSY97 \\
    \midrule
    \textbf{Subgroup with College Degree} & & & \\
    \midrule
    \textbf{Number of Test Set Transitions} & 30,920 & 19,204 & 5,898 \\
    \midrule
    PPL(CAREER)-PPL(FT-7B) & 0.388 (0.028) & 0.298 (0.043) & -0.227 (0.076) \\
    PPL(CAREER)-PPL(FT-13B) & 0.411 (0.030) & 0.390 (0.043) & -0.181 (0.073) \\
    \midrule
    \textbf{Subgroup without College Degree} & & & \\
    \midrule
    \textbf{Number of Test Set Transitions} & 30,852 & 32,389 & 24,053 \\
    \midrule
    PPL(CAREER)-PPL(FT-7B) & 0.393 (0.030) & 0.180 (0.026) & 0.035 (0.017) \\
    PPL(CAREER)-PPL(FT-13B) & 0.458 (0.031) & 0.204 (0.023) & 0.056 (0.016) \\
    \bottomrule
    \end{tabular}
    }
    \legend{Test-set-bootstrap standard errors are in parentheses.}
\end{table}
\normalsize

\begin{figure}[hbtp]
    \centering
    \includegraphics[width=0.9\linewidth]{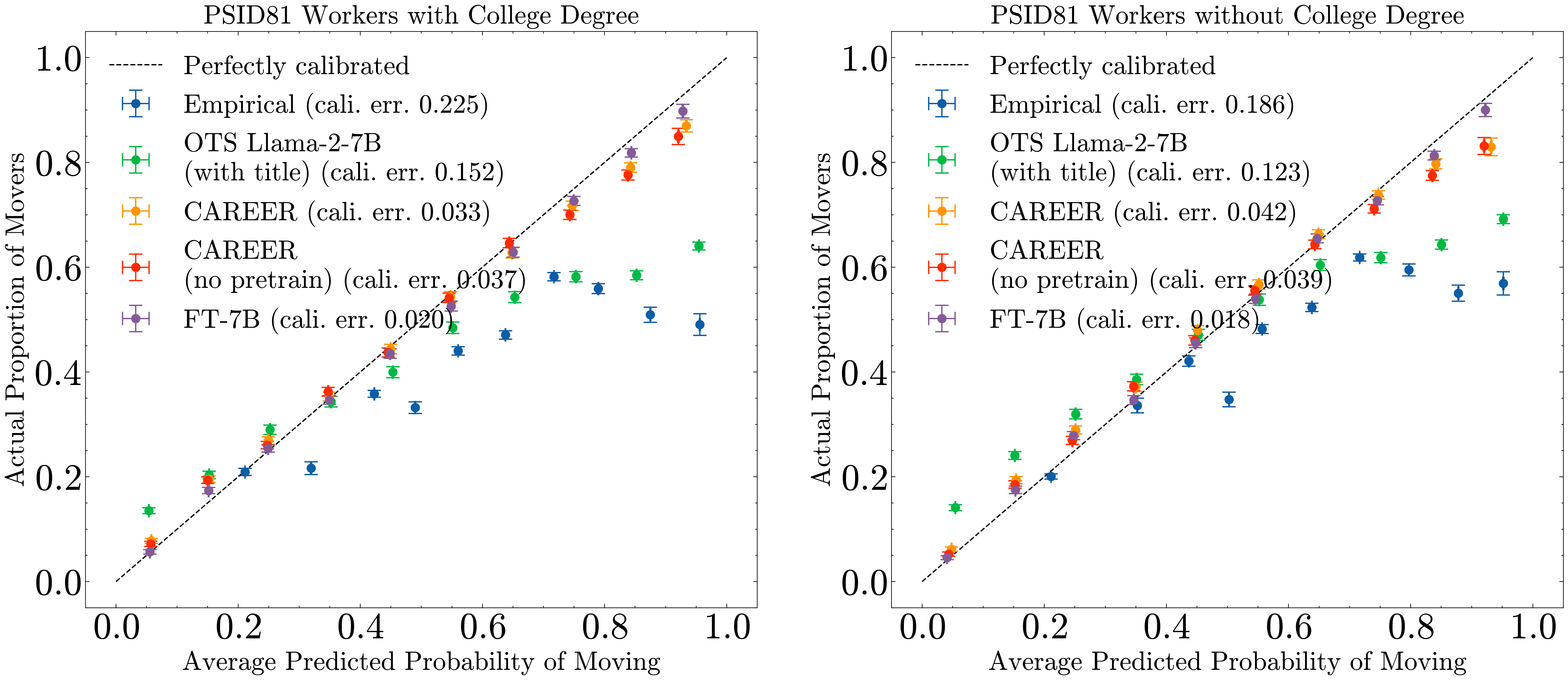}
    \includegraphics[width=0.9\linewidth]{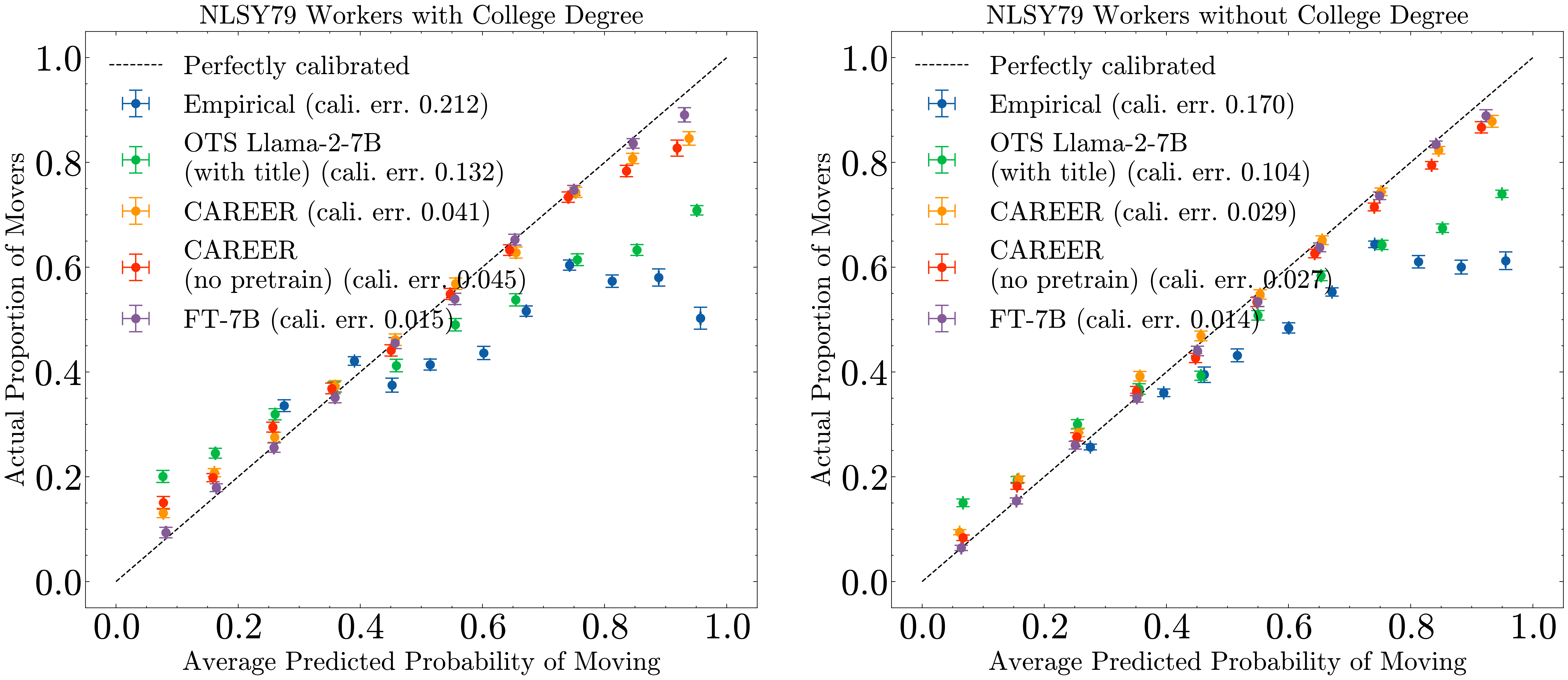}
    \includegraphics[width=0.9\linewidth]{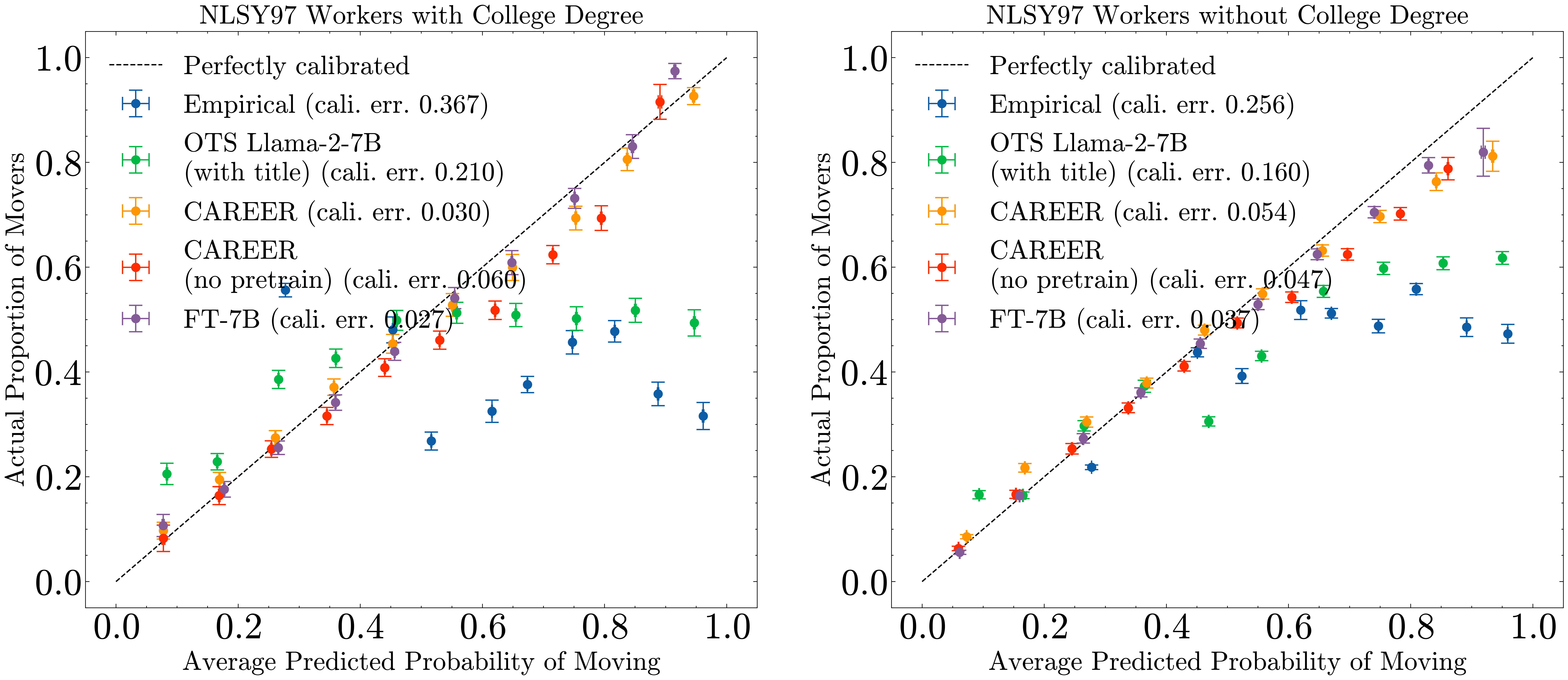}
    \caption{Calibration plots for predicting moving by different education groups.\updated{12/05/2025, passed replication check.}}
    \label{fig:education-group-calibration-separate-datasets}
\end{figure}

Next, we consider measures of performance based on the problem of predicting whether a worker changes occupations. Figure \ref{fig:education-group-calibration-separate-datasets} depicts the calibration plots for FT-7B, OTS-7B, CAREER, and empirical transition probability of predicting moving from different education subgroups and datasets. Our experiment results indicate that FT-LABOR-LLM is consistently better calibrated than CAREER across subpopulations.

Finally, Table \ref{tab:roc-edu-group-separate-datasets} presents the AUC-ROC performance metric for the empirical transitions frequency model, off-the-shelf Llama-2-7B with job titles included in the prompt, FT-7B model, and CAREER model from predicting moving in different education subgroups and datasets. Again, our results indicate that FT-LABOR-LLM consistently outperforms or achieves comparable performance to CAREER across subpopulations.

\tiny
\begin{table}[hbtp]
    \caption{Area under the ROC Curve (AUC-ROC) by different education groups.\updated{12/05/2025, passed replication check.}}
    \label{tab:roc-edu-group-separate-datasets}
    \centering
    \begin{tabular}{lrrrrrrrr}
    \toprule
     & \multicolumn{2}{c}{PSID81} & \multicolumn{2}{c}{NLSY79} & \multicolumn{2}{c}{NLSY97} & \multicolumn{2}{c}{Aggregated} \\
    Has College Degree & Yes & No & Yes & No & Yes & No & Yes & No \\
    \midrule
    Empirical & 0.640 & 0.667 & 0.588 & 0.662 & 0.441 & 0.648 & 0.599 & 0.663 \\
    OTS Llama-2-7B (with job titles) & 0.710 & 0.721 & 0.691 & 0.731 & 0.613 & 0.699 & 0.694 & 0.720\\
    CAREER & 0.777 & 0.778 & 0.763 & 0.787 & 0.748 & 0.766 & 0.770 & 0.780 \\
    CAREER
    (without pre-training) & 0.765 & 0.769 & 0.742 & 0.773 & 0.694 & 0.750 & 0.750 & 0.768 \\
    FT-7B & 0.788 & 0.789 & 0.772 & 0.796 & 0.744 & 0.769 & 0.778 & 0.788 \\
    \bottomrule
    \end{tabular}
\end{table}
\normalsize

%% file: qe_supplementary_1.tex
\newpage
\section{Alternative Evaluation Tasks}
\label{sec:online-appendix-alt-eval-tasks}
This appendix describes our experiments on evaluating different models' performance on four economically meaningful tasks alternative to the main task of predicting the next occupation. Our results here demonstrate that the LABOR-LLM models perform well on these tasks, even though it was not directly fine-tuned on these tasks, compared to baseline models trained specifically on these tasks.

\subsection{Data and Features} Following the main text, we analyze longitudinal work-history sequences from PSID81, NLSY79, and NLSY97, with individual-year transitions $(i,t)$ as the unit of observation. Data are split into train/validation/test at the individual level and the same data split is used for all models. For each observation $(i, t)$, we collect the following features: (1) structural features including ethnicity, region, education, and the calendar year, and (2) the embedding vector of the text template of career history. We use $X_{i, t}^{\mathcal{E}}=\mathcal{E}_\text{LLM}(\texttemplate(x_{i,\le t},y_{i,<t}))$, $X_{i, t}^{\text{struct}}$, and $X_{i, t}^{\text{both}}$ to denote the embedding vector, structural features, and the concatenation of the first two feature vectors, respectively.

\subsection{Four Alternative Prediction Tasks} We assess model performance across four prediction tasks that complement the primary occupation prediction objective. Let $y_{i, t}$ denote the occupation at time $t$ for individual $i$ (i.e., the prediction target in the main paper). The prediction target $\tilde{y}_{i, t}$ for four alternative tasks is derived from $y_{i, t}$ as follows: \textbf{(1) Unemployment}, which is a binary classification task with the target $\tilde{y}_{i,t} = \mathbf{1}\{y_{i,t}=\text{Unemployed}\}$; \textbf{(2) Unemployment Plus}, which is a binary classification task with the target $\tilde{y}_{i,t} = \mathbf{1}\{y_{i,t}\in\{\text{Unemployed},\ \text{Not in labor force},\ \text{In education}\}\}$; \textbf{(3) Occupation Change}, which is a binary classification task with the target $\tilde{y}_{i,t} = \mathbf{1}\{y_{i,t} \ne y_{i,t-1}\}$ (the first transition is dropped); and \textbf{(4) Coarse Occupation (multiclass)}, which is a multiclass classification task with the target $\tilde{y}_{i,t} = \texttt{coarse-label}(y_{i,t})$, where $\texttt{coarse-label}$ is a fixed mapping from fine labels to $7$ coarse occupation categories: (1) \emph{management/professional/technical/financial sales/public security}; (2) \emph{administrative support and retail sales}; (3) \emph{low-skill services}; (4) \emph{precision production \& craft}; (5) \emph{machine operators, assemblers, inspectors}; (6) \emph{transportation/construction/mechanics/mining/agricultural}; and (7) \emph{other categories}.
The detailed mapping from the fine labels to the coarse categories is presented in Table \ref{tab:occ-to-coarse-categories}.

\tiny
\begin{longtable}{p{0.2\textwidth} p{0.4\textwidth} p{0.2\textwidth}}
    \caption{Mapping from \texttt{occ1990dd} codes to coarse categories.} \label{tab:occ-to-coarse-categories} \\
    \toprule
    \shortstack{OCC\\1990DD} & \shortstack{Job \\ Title} & \shortstack{Coarse \\ Category} \\
    \midrule
    \endfirsthead
    \caption[]{Occupation Code Mapping to Coarse Categories} \\
    \toprule
    \shortstack{OCC\\1990DD} & \shortstack{Job \\ Title} & \shortstack{Coarse \\ Category} \\
    \midrule
    \endhead
    \midrule
    \multicolumn{3}{r}{Continued on next page} \\
    \midrule
    \endfoot
    \bottomrule
    \endlastfoot
    4 & Legislators &  Managproftech \\
    7 & Financial managers &  Managproftech \\
    8 & Human Resources Managers &  Managproftech \\
    13 & Public relations managers &  Managproftech \\
    14 & Education administrators &  Managproftech \\
    15 & Medical and health services managers &  Managproftech \\
    18 & Property, real estate, and community association managers &  Managproftech \\
    19 & Morticians, Undertakers, and Funeral Directors &  Managproftech \\
    22 & Postmasters and mail superintendents &  Managproftech \\
    23 & Accountants and auditors &  Managproftech \\
    24 & Insurance underwriters &  Managproftech \\
    25 & Credit analysts &  Managproftech \\
    26 & Management analysts &  Managproftech \\
    27 & Training and development specialists &  Managproftech \\
    28 & Purchasing agents and buyers, farm products &  Managproftech \\
    29 & Wholesale and retail buyers, except farm products &  Managproftech \\
    33 & Purchasing agents, except wholesale, retail, and farm products &  Managproftech \\
    34 & Agents and business managers of artists, performers, and athletes &  Managproftech \\
    35 & Construction and building inspectors &  Managproftech \\
    36 & Compliance Officers &  Managproftech \\
    37 & Emergency Management Directors &  Managproftech \\
    43 & Architects, except naval &  Managproftech \\
    44 & Aerospace engineers &  Managproftech \\
    45 & Materials engineers &  Managproftech \\
    47 & Petroleum engineers &  Managproftech \\
    48 & Chemical engineers &  Managproftech \\
    53 & Civil engineers &  Managproftech \\
    55 & Electrical and electronic engineers &  Managproftech \\
    56 & Industrial engineers, including health and safety &  Managproftech \\
    57 & Mechanical engineers &  Managproftech \\
    59 & Biomedical engineers &  Managproftech \\
    64 & Web developers &  Managproftech \\
    65 & Logisticians &  Managproftech \\
    66 & Actuaries &  Managproftech \\
    68 & Statisticians &  Managproftech \\
    69 & Astronomers and physicists &  Managproftech \\
    73 & Chemists and materials scientists &  Managproftech \\
    74 & Atmospheric and space scientists &  Managproftech \\
    75 & Environmental scientists and geoscientists &  Managproftech \\
    76 & Physical scientists, all other &  Managproftech \\
    77 & Agricultural and food scientists &  Managproftech \\
    78 & Biological scientists &  Managproftech \\
    79 & Conservation scientists and foresters &  Managproftech \\
    83 & Medical scientists &  Managproftech \\
    84 & Physicians and surgeons &  Managproftech \\
    85 & Dentists &  Managproftech \\
    86 & Veterinarians &  Managproftech \\
    87 & Optometrists &  Managproftech \\
    88 & Podiatrists &  Managproftech \\
    89 & Health diagnosing and treating practitioners, all other &  Managproftech \\
    95 & Nurse Practitioners &  Managproftech \\
    96 & Pharmacists &  Managproftech \\
    97 & Dietitians and nutritionists &  Managproftech \\
    98 & Respiratory therapists &  Managproftech \\
    99 & Occupational therapist assistants and aides &  Managproftech \\
    103 & Physical therapists &  Managproftech \\
    104 & Audiologists &  Managproftech \\
    105 & Recreational therapists &  Managproftech \\
    106 & Physician assistants &  Managproftech \\
    154 & Business Teachers Postsecondary &  Managproftech \\
    155 & Preschool and kindergarten teachers &  Managproftech \\
    156 & Elementary and middle school teachers &  Managproftech \\
    157 & Secondary school teachers &  Managproftech \\
    158 & Special education teachers &  Managproftech \\
    159 & Adult Basic and Secondary Education and Literacy Teachers and Instructors &  Managproftech \\
    163 & Counselors &  Managproftech \\
    164 & Librarians &  Managproftech \\
    165 & Archivists, curators, and museum technicians &  Managproftech \\
    166 & Market research analysts and marketing specialists &  Managproftech \\
    167 & Psychologists &  Managproftech \\
    169 & Sociologists &  Managproftech \\
    173 & Urban and regional planners &  Managproftech \\
    174 & Social workers &  Managproftech \\
    176 & Clergy &  Managproftech \\
    177 & Probation Officers and Correctional Treatment Specialists &  Managproftech \\
    178 & Lawyers, Judges, magistrates, and other judicial workers &  Managproftech \\
    183 & Writers and authors &  Managproftech \\
    184 & Technical writers &  Managproftech \\
    185 & Designers &  Managproftech \\
    186 & Musicians, singers, and related workers &  Managproftech \\
    187 & Producers and directors &  Managproftech \\
    188 & Artists and related workers &  Managproftech \\
    189 & Photographers &  Managproftech \\
    193 & Dancers and choreographers &  Managproftech \\
    194 & Miscellaneous media and communication workers &  Managproftech \\
    195 & Television, video, and motion picture camera operators and editors &  Managproftech \\
    198 & Announcers &  Managproftech \\
    199 & Athletes, coaches, umpires, and related workers &  Managproftech \\
    203 & Clinical laboratory technologists and technicians &  Managproftech \\
    204 & Dental hygienists &  Managproftech \\
    205 & Medical records and health information technicians &  Managproftech \\
    206 & Diagnostic related technologists and technicians &  Managproftech \\
    207 & Licensed practical and licensed vocational nurses &  Managproftech \\
    208 & Miscellaneous Health Technologists and Technicians &  Managproftech \\
    214 & Engineering technicians, except drafters &  Managproftech \\
    217 & Drafters &  Managproftech \\
    218 & Surveyors, cartographers, and photogrammetrists &  Managproftech \\
    223 & Agricultural and food science technicians &  Managproftech \\
    224 & Chemical technicians &  Managproftech \\
    225 & Geological and petroleum technicians &  Managproftech \\
    226 & Aircraft pilots and flight engineers &  Managproftech \\
    227 & Air traffic controllers and airfield operations specialists &  Managproftech \\
    228 & Broadcast and sound engineering technicians and radio operators &  Managproftech \\
    229 & Computer programmers &  Managproftech \\
    233 & Computer control programmers and operators &  Managproftech \\
    234 & Miscellaneous Legal Support Workers &  Managproftech \\
    235 & Nuclear technicians &  Managproftech \\
    243 & First-line supervisors/managers of retail sales workers &  Managproftech \\
    253 & Insurance sales agents &  Managproftech \\
    254 & Real estate brokers and sales agents &  Managproftech \\
    255 & Securities, commodities, and financial services sales agents &  Managproftech \\
    256 & Advertising sales agents &  Managproftech \\
    258 & Sales engineers &  Managproftech \\
    274 & Sales Representatives Services All Other &  Clericretail \\
    275 & Telemarketers &  Clericretail \\
    276 & Cashiers &  Clericretail \\
    277 & Door-to-door sales workers, news and street vendors, and related workers &  Clericretail \\
    283 & Models, demonstrators, and product promoters &  Clericretail \\
    303 & First-line supervisors/managers of office and administrative support workers &  Clericretail \\
    308 & Computer operators &  Clericretail \\
    313 & Secretaries and administrative assistants &  Clericretail \\
    315 & Desktop publishers &  Clericretail \\
    316 & Credit authorizers, checkers, and clerks &  Clericretail \\
    317 & Hotel, motel, and resort desk clerks &  Clericretail \\
    318 & Reservation and transportation ticket agents and travel clerks &  Clericretail \\
    319 & Receptionists and information clerks &  Clericretail \\
    326 & Correspondence clerks &  Clericretail \\
    328 & Human resources assistants, except payroll and timekeeping &  Clericretail \\
    329 & Library assistants, clerical &  Clericretail \\
    335 & File Clerks &  Clericretail \\
    336 & Brokerage clerks &  Clericretail \\
    337 & Bookkeeping, accounting, and auditing clerks &  Clericretail \\
    338 & Payroll and timekeeping clerks &  Clericretail \\
    344 & Billing and posting clerks and machine operators &  Clericretail \\
    346 & Postal service mail sorters, processors, and processing machine operators &  Clericretail \\
    347 & Office machine operators, except computer &  Clericretail \\
    348 & Switchboard operators, including answering service &  Clericretail \\
    349 & Communications Equipment Operators All Other &  Clericretail \\
    354 & Postal service clerks &  Clericretail \\
    355 & Postal service mail carriers &  Clericretail \\
    356 & Mail clerks and mail machine operators, except postal service &  Clericretail \\
    357 & Couriers and messengers &  Clericretail \\
    359 & Dispatchers &  Clericretail \\
    364 & Cargo and freight agents &  Clericretail \\
    365 & Stock clerks and order fillers &  Clericretail \\
    366 & Meter readers, utilities &  Clericretail \\
    368 & Weighers, measurers, checkers, and samplers, recordkeeping &  Clericretail \\
    373 & Transportation, storage, and distribution managers &  Clericretail \\
    375 & Claims adjusters, appraisers, examiners, and investigators &  Clericretail \\
    376 & Loan interviewers and clerks &  Clericretail \\
    377 & Eligibility interviewers, government programs &  Clericretail \\
    378 & Bill and account collectors &  Clericretail \\
    379 & Office clerks, general &  Clericretail \\
    383 & Tellers &  Clericretail \\
    384 & Proofreaders and copy markers &  Clericretail \\
    385 & Data entry keyers &  Clericretail \\
    386 & Statistical assistants &  Clericretail \\
    387 & Teacher assistants &  Clericretail \\
    389 & Court, municipal, and license clerks &  Clericretail \\
    405 & Maids and housekeeping cleaners &  Service \\
    408 & Laundry and dry-cleaning workers &  Service \\
    415 & First-Line Supervisors of Protective Service Workers (All Other) &  Service \\
    417 & Fire inspectors &  Managproftech \\
    418 & Private detectives and investigators &  Managproftech \\
    423 & First-line supervisors/managers of correctional officers &  Managproftech \\
    425 & Crossing guards &  Service \\
    426 & Security Guards and Gaming Surveillance Officers &  Service \\
    427 & Animal control workers &  Service \\
    433 & First-line supervisors/managers of food preparation and serving workers &  Service \\
    434 & Bartenders &  Service \\
    435 & Waiters and waitresses &  Service \\
    436 & Cooks &  Service \\
    439 & Combined food preparation and serving workers, including fast food &  Service \\
    444 & Food servers, nonrestaurant &  Service \\
    445 & Dental assistants &  Service \\
    447 & Phlebotomists &  Service \\
    448 & First-line supervisors/managers of housekeeping and janitorial workers &  Service \\
    450 & First-Line Supervisors of Landscaping and Lawn Service and Groundskeeping Workers &  Service \\
    451 & Grounds maintenance workers &  Service \\
    453 & Janitors and Cleaners Except Maids and Housekeeping Cleaners &  Service \\
    455 & Pest control workers &  Service \\
    457 & Barbers &  Service \\
    458 & Miscellaneous personal appearance workers &  Service \\
    459 & Miscellaneous entertainment attendants and related workers &  Service \\
    461 & Tour and travel guides &  Service \\
    462 & Ushers, lobby attendants, and ticket takers &  Service \\
    464 & Baggage porters, bellhops, and concierges &  Service \\
    466 & Recreation and fitness workers &  Service \\
    467 & Motion picture projectionists &  Service \\
    468 & Child care workers &  Service \\
    469 & Hosts and hostesses, restaurant, lounge, and coffee shop &  Service \\
    470 & First-line supervisors/managers of personal service workers &  Service \\
    471 & Transportation inspectors &  Service \\
    472 & Nonfarm animal caretakers &  Service \\
    473 & Farmers, Ranchers, and other Agricultural Managers &  Transmechcraft \\
    475 & Animal breeders &  Transmechcraft \\
    479 & Miscellaneous agricultural workers &  Transmechcraft \\
    488 & Graders and sorters, agricultural products &  Transmechcraft \\
    489 & Agricultural inspectors &  Transmechcraft \\
    496 & Forest and conservation workers &  Transmechcraft \\
    498 & Fishers and related fishing workers &  Transmechcraft \\
    503 & First-line supervisors/managers of mechanics, installers, and repairers &  Transmechcraft \\
    505 & Automotive service technicians and mechanics &  Transmechcraft \\
    507 & Bus and truck mechanics and diesel engine specialists &  Transmechcraft \\
    508 & Aircraft mechanics and service technicians &  Transmechcraft \\
    509 & Small engine mechanics &  Transmechcraft \\
    514 & Automotive glass installers and repairers &  Transmechcraft \\
    516 & Heavy vehicle and mobile equipment service technicians and mechanics &  Transmechcraft \\
    518 & Industrial and refractory machinery mechanics &  Transmechcraft \\
    519 & Maintenance workers, machinery &  Transmechcraft \\
    523 & Electronic home entertainment equipment installers and repairers &  Transmechcraft \\
    525 & Computer, automated teller, and office machine repairers &  Transmechcraft \\
    526 & Home appliance repairers &  Transmechcraft \\
    527 & Radio and telecommunications equipment installers and repairers &  Transmechcraft \\
    533 & Electronic equipment installers and repairers, motor vehicles &  Transmechcraft \\
    534 & Heating, air conditioning, and refrigeration mechanics and installers &  Transmechcraft \\
    535 & Precision instrument and equipment repairers &  Transmechcraft \\
    536 & Locksmiths and safe repairers &  Transmechcraft \\
    539 & Control and valve installers and repairers &  Transmechcraft \\
    543 & Elevator installers and repairers &  Transmechcraft \\
    544 & Millwrights &  Transmechcraft \\
    549 & Maintenance and repair workers, general &  Transmechcraft \\
    558 & First-line supervisors/managers of construction trades and extraction workers &  Transmechcraft \\
    563 & Carpet, floor, and tile installers and finishers &  Transmechcraft \\
    567 & Carpenters &  Transmechcraft \\
    573 & Drywall installers, ceiling tile installers, and tapers &  Transmechcraft \\
    575 & Electricians &  Transmechcraft \\
    577 & Electric motor, power tool, and related repairers &  Transmechcraft \\
    579 & Painters, construction and maintenance &  Transmechcraft \\
    583 & Paperhangers &  Transmechcraft \\
    584 & Plasterers and stucco masons &  Transmechcraft \\
    585 & Pipelayers, plumbers, pipefitters, and steamfitters &  Transmechcraft \\
    588 & Cement masons, concrete finishers, and terrazzo workers &  Transmechcraft \\
    589 & Glaziers &  Transmechcraft \\
    593 & Insulation workers &  Transmechcraft \\
    594 & Paving, surfacing, and tamping equipment operators &  Transmechcraft \\
    595 & Roofers &  Transmechcraft \\
    597 & Structural metal fabricators and fitters &  Transmechcraft \\
    598 & Earth drillers, except oil and gas &  Transmechcraft \\
    599 & Sheet metal workers &  Transmechcraft \\
    614 & Derrick, rotary drill, and service unit operators, oil, gas, and mining &  Transmechcraft \\
    615 & Explosives workers, ordnance handling experts, and blasters &  Transmechcraft \\
    616 & Mining machine operators &  Transmechcraft \\
    617 & Roof bolters, mining &  Transmechcraft \\
    628 & First-line supervisors/managers of production and operating workers &  Product \\
    634 & Tool and die makers &  Product \\
    637 & Machinists &  Product \\
    643 & Boilermakers &  Product \\
    644 & Tool grinders, filers, and sharpeners &  Product \\
    645 & Model makers and patternmakers, metal and plastic &  Product \\
    649 & Etchers and engravers &  Product \\
    653 & Lay-out workers, metal and plastic &  Product \\
    657 & Cabinetmakers and bench carpenters &  Product \\
    658 & Furniture finishers &  Product \\
    666 & Tailors, dressmakers, and sewers &  Product \\
    668 & Upholsterers &  Product \\
    669 & Shoe and leather workers and repairers &  Product \\
    675 & Molders, shapers, and casters, except metal and plastic &  Product \\
    677 & Opticians, dispensing &  Product \\
    678 & Health Practitioner Support Technologists and Technicians &  Product \\
    679 & Print Binding and Finishing Workers &  Product \\
    684 & Multiple machine tool setters, operators, and tenders, metal and plastic &  Product \\
    686 & Butchers and other meat, poultry, and fish processing workers &  Product \\
    687 & Bakers &  Product \\
    688 & Food batchmakers &  Product \\
    694 & Water and liquid waste treatment plant and system operators &  Product \\
    695 & Power plant operators, distributors, and dispatchers &  Product \\
    696 & Stationary engineers and boiler operators &  Product \\
    699 & Miscellaneous plant and system operators &  Product \\
    703 & Lathe and turning machine tool setters, operators, and tenders, metal and plastic &  Operator \\
    706 & Cutting, punching, and press machine setters, operators, and tenders, metal and plastic &  Operator \\
    707 & Rolling machine setters, operators, and tenders, metal and plastic &  Operator \\
    708 & Drilling and boring machine tool setters, operators, and tenders, metal and plastic &  Operator \\
    709 & Grinding, Lapping, Polishing, and Buffing Machine Tool Setters, Operators, and Tenders, Metal and Plastic &  Operator \\
    713 & Forging machine setters, operators, and tenders, metal and plastic &  Operator \\
    719 & Molders and molding machine setters, operators, and tenders, metal and plastic &  Operator \\
    723 & Plating and coating machine setters, operators, and tenders, metal and plastic &  Operator \\
    724 & Heat treating equipment setters, operators, and tenders, metal and plastic &  Operator \\
    727 & Sawing machine setters, operators, and tenders, wood &  Operator \\
    729 & Woodworking machine setters, operators, and tenders, except sawing &  Operator \\
    733 & Woodworkers (All Other) &  Operator \\
    734 & Printing Press Operators &  Operator \\
    736 & Prepress technicians and workers &  Operator \\
    738 & Textile winding, twisting, and drawing out machine setters, operators, and tenders &  Operator \\
    739 & Textile knitting and weaving machine setters, operators, and tenders &  Operator \\
    743 & Textile cutting machine setters, operators, and tenders &  Operator \\
    744 & Sewing machine operators &  Operator \\
    745 & Shoe machine operators and tenders &  Operator \\
    747 & Pressers, textile, garment, and related materials &  Operator \\
    749 & Textile bleaching and dyeing machine operators and tenders &  Operator \\
    753 & Cementing and gluing machine operators and tenders &  Operator \\
    754 & Packaging and filling machine operators and tenders &  Operator \\
    755 & Extruding and drawing machine setters, operators, and tenders, metal and plastic &  Operator \\
    756 & Crushing, grinding, polishing, mixing, and blending workers &  Operator \\
    757 & Chemical processing machine setters, operators, and tenders &  Operator \\
    763 & Food and tobacco roasting, baking, and drying machine operators and tenders &  Operator \\
    764 & Cleaning, washing, and metal pickling equipment operators and tenders &  Operator \\
    765 & Paper goods machine setters, operators, and tenders &  Operator \\
    766 & Metal furnace and kiln operators and tenders &  Operator \\
    769 & Cutting workers &  Operator \\
    774 & Photographic process workers and processing machine operators &  Operator \\
    779 & Painting workers &  Operator \\
    783 & Welding, soldering, and brazing workers &  Operator \\
    785 & Miscellaneous assemblers and fabricators &  Operator \\
    789 & Painters, Transportation Equipment &  Operator \\
    799 & Inspectors, testers, sorters, samplers, and weighers &  Operator \\
    803 & First-Line Supervisors of Transportation and Material-Moving Machine and Vehicle Operators &  Transmechcraft \\
    804 & Coin, vending, and amusement machine servicers and repairers &  Transmechcraft \\
    808 & Bus drivers &  Transmechcraft \\
    809 & Taxi drivers and chauffeurs &  Transmechcraft \\
    813 & Parking lot attendants &  Transmechcraft \\
    823 & Railroad conductors and yardmasters &  Transmechcraft \\
    824 & Locomotive engineers and operators &  Transmechcraft \\
    825 & Railroad brake, signal, and switch operators &  Transmechcraft \\
    829 & Ship and boat captains and operators &  Transmechcraft \\
    834 & Bridge and lock tenders &  Transmechcraft \\
    844 & Operating engineers and other construction equipment operators &  Transmechcraft \\
    848 & Crane and tower operators &  Transmechcraft \\
    853 & Dredge, excavating, and loading machine operators &  Transmechcraft \\
    859 & Shuttle car operators &  Transmechcraft \\
    865 & Helpers--installation, maintenance, and repair workers &  Transmechcraft \\
    866 & Helpers, construction trades &  Transmechcraft \\
    869 & Construction laborers &  Transmechcraft \\
    873 & Helpers--production workers &  Transmechcraft \\
    875 & Refuse and recyclable material collectors &  Transmechcraft \\
    878 & Machine feeders and offbearers &  Transmechcraft \\
    885 & Service station attendants &  Transmechcraft \\
    887 & Cleaners of vehicles and equipment &  Transmechcraft \\
    888 & Packers and packagers, hand &  Transmechcraft \\
    889 & Septic tank servicers and sewer pipe cleaners &  Transmechcraft \\
    905 & Air Crew Officers & Other Category \\
    999 & Special Code (occ1990dd=999) & Other Category \\
    education & In education & Other Category \\
    not\_in\_labor\_force & Not in labor force & Other Category \\
    unemployed & Unemployed & Other Category \\
\end{longtable}
\normalsize

\subsection{Models} For LABOR-LLM and CAREER, we utilize the models fine-tuned on the \emph{primary occupation prediction objective} described in the main paper. To adapt these models to the alternative tasks, we aggregate the predicted probabilities over the relevant occupation categories.
Specifically, for the Coarse Occupation task, we sum the probabilities of all fine-grained occupations mapping to a given coarse category. Similarly, for the binary tasks (Unemployment, Unemployment Plus, and Occupation Change), we aggregate probabilities over the labels corresponding to the positive class.

\subsection{Baseline Models and Training} We evaluate several baseline models trained directly on the alternative tasks. The model families include Logistic Regression with L2 regularization, Multilayer Perceptrons (MLP), Gradient Boosting, and Random Forest.
For each alternative prediction task and model family, we perform a hyperparameter search over 100 configurations sampled from the grid described in Table \ref{tab:alternative-evaluation-tasks-hyper-parameter-search-space}, with more than 98.3\% of the sampled configurations successfully completed within the 48-hour time limit.
We then train a model on the training data using the best hyperparameters (selected via validation set log-loss) and evaluate it on the test data.
Unlike the LABOR-LLM and CAREER models, which are fine-tuned on the primary next occupation prediction task and then adapted to these alternative tasks by aggregating their predicted probabilities, these baseline models are trained directly on the specific target $\tilde{y}_{i, t}$ for each alternative task. This direct supervision should provide a stronger signal for the specific prediction tasks we consider in this section.

For each model family, we also evaluate probability calibration. We fit calibrators on validation predictions only: Platt scaling (logistic regression on scores) and isotonic regression; we choose the method with the lower validation Brier score. For multiclass tasks, calibration is applied in a one-vs-rest manner and the calibrated classwise probabilities are renormalized to sum to one. The selected calibrator is then applied to test predictions; no test information is used for training or selection. We compare the performance of the calibrated models with the uncalibrated models and report the better performance between the two.

\tiny
\begin{table}[H]
  \centering
    \caption{Hyperparameter search space. For each model family and feature set combination, we sample 100 random configurations. Each configuration is trained with a distinct random seed to ensure robustness. PCA applies only to embeddings-bearing feature sets; $0$ denotes no PCA (i.e., full features are used).}
  \label{tab:alternative-evaluation-tasks-hyper-parameter-search-space}
  \setlength{\tabcolsep}{6pt}
  \renewcommand{\arraystretch}{1.05}
  \begin{tabular}{@{}p{0.3\linewidth}p{0.28\linewidth}p{0.32\linewidth}@{}}
    \toprule
    Model family & Hyperparameter & Search space \\
    \midrule
    \multicolumn{3}{@{}l}{\textbf{Logistic Regression (scikit-learn)}} \\
     & \texttt{pca\_emb\_dim} & $\{0, 128, 256\}$ (embeddings only) \\
     & \texttt{C} & $\{0.001, 0.003, 0.01, 0.03, 0.1, 0.3, 1, 3, 10, 30\}$ \\
     & \texttt{class\_weight} & $\{\texttt{None}, \texttt{balanced}\}$ \\
    \midrule
    \multicolumn{3}{@{}l}{\shortstack[l]{\textbf{MLP}\\\textbf{(PyTorch)}}} \\
     & \texttt{pca\_emb\_dim} & $\{0, 128, 256\}$ (embeddings only) \\
     & \texttt{depth} & $\{2, 3\}$ \\
     & \texttt{width} & $\{512, 1024, 2048, 4096\}$ \\
     & \texttt{activation} & $\{\texttt{ReLU}, \texttt{GELU}\}$ \\
     & \texttt{dropout} & $\{0.0, 0.1, 0.2, 0.3\}$ \\
     & \texttt{weight\_decay} & $\{0, 10^{-6}, 10^{-5}, 10^{-4}, 10^{-3}\}$ \\
     & \texttt{lr} & $\{10^{-4}, 3\cdot10^{-4}, 10^{-3}\}$ \\
     & \texttt{batch\_size} & $\{512, 1024, 2048, 4096\}$ \\
     & \texttt{scheduler} & $\{\texttt{none}, \texttt{plateau}\}$ \\
     & \texttt{plateau\_factor} & $\{0.3, 0.5, 0.7\}$ \\
     & \texttt{plateau\_patience} & $\{3, 5, 8\}$ \\
     & \texttt{max\_epochs} & $1000$ \\
     & \texttt{patience} & $20$ \\
    \midrule
    \multicolumn{3}{@{}l}{\shortstack[l]{\textbf{Gradient Boosting}\\\textbf{(scikit-learn HGBT)}}} \\
     & \texttt{pca\_emb\_dim} & $\{0, 128, 256\}$ (embeddings only) \\
     & \texttt{learning\_rate} & $\{0.03, 0.05, 0.1\}$ \\
     & \texttt{max\_depth} & $\{\texttt{None}, 8, 10, 12\}$ \\
     & \texttt{max\_iter} & $\{500, 800, 1200\}$ \\
     & \texttt{l2\_regularization} & $\{0, 10^{-4}, 10^{-3}, 10^{-2}, 10^{-1}\}$ \\
     & \texttt{max\_leaf\_nodes} & $\{63, 127, 255\}$ \\
    \midrule
    \multicolumn{3}{@{}l}{\shortstack[l]{\textbf{Random Forest}\\\textbf{(scikit-learn)}}} \\
     & \texttt{pca\_emb\_dim} & $\{0, 128, 256\}$ (embeddings only) \\
     & \texttt{n\_estimators} & $\{500, 800, 1200, 1600\}$ \\
     & \texttt{max\_depth} & $\{15, 25, 35\}$ \\
     & \texttt{min\_samples\_leaf} & $\{1, 2, 5, 10\}$ \\
     & \texttt{max\_features} & $\{\texttt{sqrt}, \texttt{log2}, 0.4, 0.6, 0.8, 1.0\}$ \\
     & \texttt{class\_weight} & $\{\texttt{None}, \texttt{balanced}\}$ \\
    \bottomrule
  \end{tabular}
\end{table}
\normalsize

\subsection{Metrics for Evaluation}
After model training and the optional calibration step, we compute a set of comprehensive metrics on the test set to compare the model's performance. \textbf{(1) Log-loss and log-likelihood}: We define the average log-likelihood as $\ell = \frac{1}{N}\sum_{i}\sum_{t=1}^{T_i} \log \hat{p}_{i,t}$, where $\hat{p}_{i,t}$ is the model's predicted probability for the true label $\tilde{y}_{i,t}$ in each of the four tasks.
We report both log-loss, defined as $-\ell$ (lower is better), and log-likelihood $\ell$ (higher is better).
\textbf{(2) Brier score.} The Brier score captures the mean square error between the predicted probability and the true label, defined as $\text{Brier} = \frac{1}{N}\sum_{i,t}\big(\hat{p}_{i,t}-\tilde{y}_{i,t}\big)^2$, where $\hat{p}_{i,t}$ is the model's predicted probability for the true label $\tilde{y}_{i,t}$ in the three binary prediction tasks. For the multiclass prediction task, we use the mean per-class squared error: $\text{Brier} = \frac{1}{N}\sum_{i,t}\frac{1}{K}\big\|\hat{\mathbf{P}}_{i,t}-\mathbf{e}(\tilde{y}_{i,t})\big\|_2^2$, where $\hat{\mathbf{P}}_{i,t}$ denotes the vector of predicted class probabilities, $\mathbf{e}(\tilde{y}_{i,t})$ is the one-hot vector for the true label, and $K$ is the number of classes. A lower Brier score indicates better performance.
\textbf{(3) AUC-ROC and AUPRC.} For binary tasks, we compute Area Under the Receiver Operating Characteristic Curve (AUC-ROC) and Area Under the Precision-Recall Curve (AUPRC) using the pairs $\{(\tilde{y}_{i,t},\hat{p}_{i,t})\}$. For multiclass, we use macro one-vs-rest (average across classes) as in scikit-learn. A higher AUC-ROC and AUPRC indicate better performance.
\textbf{(4) Expected Calibration Error (ECE).} Using $M=15$ equal-width bins over confidence, let the confidence be $c_{i,t}=\max\{\hat{p}_{i,t},1-\hat{p}_{i,t}\}$ in binary and $c_{i,t}=\max_k \hat{P}(k\mid x_{i,\le t}, y_{i,<t})$ in multiclass with $\hat{y}_{i,t}=\arg\max_k \hat{P}(k\mid \cdot)$. Let $B_b$ be the set of instances whose confidence falls in bin $b$, with $\text{acc}_b=\frac{1}{|B_b|}\sum_{(i,t)\in B_b}\mathbf{1}\{\hat{y}_{i,t}=\tilde{y}_{i,t}\}$ and $\text{conf}_b=\frac{1}{|B_b|}\sum_{(i,t)\in B_b} c_{i,t}$. Then $\text{ECE} = \sum_{b=1}^{M}\frac{|B_b|}{N}\,\big|\text{acc}_b-\text{conf}_b\big|$. A lower ECE indicates better performance.
\textbf{(5) Accuracy.} For binary tasks, we use the $0.5$ decision rule; for multiclass, we report top-1 accuracy (i.e., $\mathbf{1}\{\hat{y}_{i,t}=\tilde{y}_{i,t}\}$ averaged over the test set). A higher accuracy indicates better performance.

\subsection{Uncertainty Quantification}
As mentioned in the main text, there are two sources of data uncertainty in our evaluation: (1) variation in the training data and (2) variation in the test data. We use the bootstrap to quantify the uncertainty of the model's performance. We bootstrap the training data at the individual level and train a model on each bootstrap sample, which provides an estimate of the uncertainty in the model's performance due to the variation in the training data. We use $B_{\text{train}}=12$ samples for the training-set bootstrapping due to the computational cost. We then evaluate the model on the test data and compute the standard deviation of the performance metrics when we bootstrap the test data. We use $B_{\text{test}} = 100$ resamples for the test-set bootstrap, as reported in Table \ref{tab:alternative-metrics-performance-results-with-se}. We report these two types of uncertainties as bootstrap standard errors alongside point estimates.

\subsection{Results}
Table \ref{tab:alternative-tasks-best-hyperparameters} reports the best hyperparameters selected via cross-validation for each model configuration.

\tiny
\begin{landscape}
\begin{longtable}{llllp{6cm}}
    \caption{Best hyperparameters selected via cross-validation for each model configuration. Hyperparameters were tuned separately for each combination of task, model type, feature set, and calibration setting.} \label{tab:alternative-tasks-best-hyperparameters} \\
    \toprule
    Prediction Task & Model & Feature Set & Calibrated & Best Hyperparameters \\
    \midrule
    \endfirsthead
    \caption[]{Best hyperparameters selected via cross-validation for each model configuration. Hyperparameters were tuned separately for each combination of task, model type, feature set, and calibration setting.\updated{12/05/2025, passed replication check.}} \\
    \toprule
    Prediction Task & Model & Feature Set & Calibrated & Best Hyperparameters \\
    \midrule
    \endhead
    \midrule
    \multicolumn{5}{r}{Continued on next page} \\
    \midrule
    \endfoot
    \bottomrule
    \endlastfoot
    Coarse Occupation & Gradient Boosting & $X_{i, t}^{\text{both}}$ & True & l2\_regularization=0, learning\_rate=0.03, max\_depth=None, max\_iter=800, max\_leaf\_nodes=63, pca\_emb\_dim=256 \\
    Coarse Occupation & Gradient Boosting & $X_{i, t}^{\mathcal{E}}$ & True & l2\_regularization=0.01, learning\_rate=0.03, max\_depth=8, max\_iter=1200, max\_leaf\_nodes=63, pca\_emb\_dim=256 \\
    Coarse Occupation & Gradient Boosting & $X_{i, t}^{\text{struct}}$ & True & l2\_regularization=0, learning\_rate=0.03, max\_depth=8, max\_iter=500, max\_leaf\_nodes=63, pca\_emb\_dim=0 \\
    Coarse Occupation & Logistic Regression & $X_{i, t}^{\text{both}}$ & True & C=0.01, class\_weight=None, pca\_emb\_dim=0 \\
    Coarse Occupation & Logistic Regression & $X_{i, t}^{\mathcal{E}}$ & True & C=0.03, class\_weight=None, pca\_emb\_dim=0 \\
    Coarse Occupation & Logistic Regression & $X_{i, t}^{\text{struct}}$ & True & C=0.001, class\_weight=None, pca\_emb\_dim=0 \\
    Coarse Occupation & MLP & $X_{i, t}^{\text{both}}$ & True & activation=gelu, batch\_size=1024, dropout=0.3, hidden=[512, 512], lr=0.0001, max\_epochs=1000, patience=20, pca\_emb\_dim=0, plateau\_factor=0.7, plateau\_patience=8, scheduler=none, weight\_decay=1e-05 \\
    Coarse Occupation & MLP & $X_{i, t}^{\mathcal{E}}$ & True & activation=relu, batch\_size=2048, dropout=0.3, hidden=[1024, 1024, 1024], lr=0.0001, max\_epochs=1000, patience=20, pca\_emb\_dim=0, plateau\_factor=0.7, plateau\_patience=8, scheduler=plateau, weight\_decay=0.0001 \\
    Coarse Occupation & MLP & $X_{i, t}^{\text{struct}}$ & True & activation=gelu, batch\_size=2048, dropout=0.3, hidden=[1024, 1024, 1024], lr=0.0001, max\_epochs=1000, patience=20, pca\_emb\_dim=0, plateau\_factor=0.5, plateau\_patience=3, scheduler=plateau, weight\_decay=0 \\
    Coarse Occupation & Random Forest & $X_{i, t}^{\text{both}}$ & True & class\_weight=None, max\_depth=25, max\_features=log2, min\_samples\_leaf=5, n\_estimators=1200, pca\_emb\_dim=256 \\
    Coarse Occupation & Random Forest & $X_{i, t}^{\mathcal{E}}$ & True & class\_weight=None, max\_depth=25, max\_features=log2, min\_samples\_leaf=10, n\_estimators=500, pca\_emb\_dim=256 \\
    Coarse Occupation & Random Forest & $X_{i, t}^{\text{struct}}$ & True & class\_weight=None, max\_depth=15, max\_features=log2, min\_samples\_leaf=10, n\_estimators=800, pca\_emb\_dim=0 \\
    \midrule
    Occupation Change & Gradient Boosting & $X_{i, t}^{\text{both}}$ & True & l2\_regularization=0.001, learning\_rate=0.03, max\_depth=None, max\_iter=1200, max\_leaf\_nodes=127, pca\_emb\_dim=0 \\
    Occupation Change & Gradient Boosting & $X_{i, t}^{\mathcal{E}}$ & True & l2\_regularization=0.0001, learning\_rate=0.03, max\_depth=10, max\_iter=1200, max\_leaf\_nodes=63, pca\_emb\_dim=0 \\
    Occupation Change & Gradient Boosting & $X_{i, t}^{\text{struct}}$ & True & l2\_regularization=0, learning\_rate=0.03, max\_depth=8, max\_iter=500, max\_leaf\_nodes=63, pca\_emb\_dim=0 \\
    Occupation Change & Logistic Regression & $X_{i, t}^{\text{both}}$ & True & C=0.01, class\_weight=balanced, pca\_emb\_dim=0 \\
    Occupation Change & Logistic Regression & $X_{i, t}^{\mathcal{E}}$ & True & C=0.01, class\_weight=balanced, pca\_emb\_dim=0 \\
    Occupation Change & Logistic Regression & $X_{i, t}^{\text{struct}}$ & True & C=0.001, class\_weight=balanced, pca\_emb\_dim=0 \\
    Occupation Change & MLP & $X_{i, t}^{\text{both}}$ & True & activation=relu, batch\_size=512, dropout=0.3, hidden=[1024, 1024, 1024], lr=0.0001, max\_epochs=1000, patience=20, pca\_emb\_dim=0, plateau\_factor=0.5, plateau\_patience=5, scheduler=plateau, weight\_decay=1e-06 \\
    Occupation Change & MLP & $X_{i, t}^{\mathcal{E}}$ & True & activation=relu, batch\_size=2048, dropout=0.2, hidden=[512, 512, 512], lr=0.001, max\_epochs=1000, patience=20, pca\_emb\_dim=0, plateau\_factor=0.7, plateau\_patience=3, scheduler=none, weight\_decay=1e-06 \\
    Occupation Change & MLP & $X_{i, t}^{\text{struct}}$ & True & activation=relu, batch\_size=4096, dropout=0.3, hidden=[512, 512, 512], lr=0.001, max\_epochs=1000, patience=20, pca\_emb\_dim=0, plateau\_factor=0.5, plateau\_patience=5, scheduler=none, weight\_decay=1e-05 \\
    Occupation Change & Random Forest & $X_{i, t}^{\text{both}}$ & True & class\_weight=balanced, max\_depth=35, max\_features=0.4, min\_samples\_leaf=10, n\_estimators=800, pca\_emb\_dim=256 \\
    Occupation Change & Random Forest & $X_{i, t}^{\mathcal{E}}$ & True & class\_weight=balanced, max\_depth=15, max\_features=0.4, min\_samples\_leaf=10, n\_estimators=800, pca\_emb\_dim=256 \\
    Occupation Change & Random Forest & $X_{i, t}^{\text{struct}}$ & True & class\_weight=balanced, max\_depth=15, max\_features=0.4, min\_samples\_leaf=10, n\_estimators=800, pca\_emb\_dim=0 \\
    \midrule
    Unemployment & Gradient Boosting & $X_{i, t}^{\text{both}}$ & True & l2\_regularization=0, learning\_rate=0.03, max\_depth=None, max\_iter=800, max\_leaf\_nodes=63, pca\_emb\_dim=256 \\
    Unemployment & Gradient Boosting & $X_{i, t}^{\mathcal{E}}$ & True & l2\_regularization=0.001, learning\_rate=0.03, max\_depth=None, max\_iter=800, max\_leaf\_nodes=63, pca\_emb\_dim=256 \\
    Unemployment & Gradient Boosting & $X_{i, t}^{\text{struct}}$ & True & l2\_regularization=0.001, learning\_rate=0.03, max\_depth=12, max\_iter=500, max\_leaf\_nodes=63, pca\_emb\_dim=0 \\
    Unemployment & Logistic Regression & $X_{i, t}^{\text{both}}$ & True & C=0.003, class\_weight=None, pca\_emb\_dim=0 \\
    Unemployment & Logistic Regression & $X_{i, t}^{\mathcal{E}}$ & True & C=0.01, class\_weight=None, pca\_emb\_dim=0 \\
    Unemployment & Logistic Regression & $X_{i, t}^{\text{struct}}$ & True & C=30, class\_weight=balanced, pca\_emb\_dim=0 \\
    Unemployment & MLP & $X_{i, t}^{\text{both}}$ & True & activation=gelu, batch\_size=2048, dropout=0.3, hidden=[1024, 1024, 1024], lr=0.0001, max\_epochs=1000, patience=20, pca\_emb\_dim=0, plateau\_factor=0.5, plateau\_patience=3, scheduler=plateau, weight\_decay=0 \\
    Unemployment & MLP & $X_{i, t}^{\mathcal{E}}$ & True & activation=gelu, batch\_size=1024, dropout=0.2, hidden=[512, 512, 512], lr=0.0001, max\_epochs=1000, patience=20, pca\_emb\_dim=0, plateau\_factor=0.3, plateau\_patience=5, scheduler=none, weight\_decay=0.0001 \\
    Unemployment & MLP & $X_{i, t}^{\text{struct}}$ & True & activation=relu, batch\_size=2048, dropout=0.3, hidden=[4096, 4096, 4096], lr=0.0003, max\_epochs=1000, patience=20, pca\_emb\_dim=0, plateau\_factor=0.7, plateau\_patience=5, scheduler=none, weight\_decay=0.001 \\
    Unemployment & Random Forest & $X_{i, t}^{\text{both}}$ & True & class\_weight=None, max\_depth=15, max\_features=log2, min\_samples\_leaf=5, n\_estimators=1200, pca\_emb\_dim=128 \\
    Unemployment & Random Forest & $X_{i, t}^{\mathcal{E}}$ & True & class\_weight=None, max\_depth=15, max\_features=log2, min\_samples\_leaf=10, n\_estimators=800, pca\_emb\_dim=128 \\
    Unemployment & Random Forest & $X_{i, t}^{\text{struct}}$ & True & class\_weight=balanced, max\_depth=15, max\_features=log2, min\_samples\_leaf=10, n\_estimators=1600, pca\_emb\_dim=0 \\
    \midrule
    Unemployment Plus & Gradient Boosting & $X_{i, t}^{\text{both}}$ & True & l2\_regularization=0, learning\_rate=0.05, max\_depth=8, max\_iter=800, max\_leaf\_nodes=63, pca\_emb\_dim=0 \\
    Unemployment Plus & Gradient Boosting & $X_{i, t}^{\mathcal{E}}$ & True & l2\_regularization=0.0001, learning\_rate=0.03, max\_depth=10, max\_iter=1200, max\_leaf\_nodes=63, pca\_emb\_dim=0 \\
    Unemployment Plus & Gradient Boosting & $X_{i, t}^{\text{struct}}$ & True & l2\_regularization=0, learning\_rate=0.03, max\_depth=8, max\_iter=500, max\_leaf\_nodes=63, pca\_emb\_dim=0 \\
    Unemployment Plus & Logistic Regression & $X_{i, t}^{\text{both}}$ & True & C=0.03, class\_weight=None, pca\_emb\_dim=0 \\
    Unemployment Plus & Logistic Regression & $X_{i, t}^{\mathcal{E}}$ & True & C=0.01, class\_weight=None, pca\_emb\_dim=0 \\
    Unemployment Plus & Logistic Regression & $X_{i, t}^{\text{struct}}$ & True & C=0.3, class\_weight=None, pca\_emb\_dim=0 \\
    Unemployment Plus & MLP & $X_{i, t}^{\text{both}}$ & True & activation=relu, batch\_size=4096, dropout=0.1, hidden=[4096, 4096, 4096], lr=0.001, max\_epochs=1000, patience=20, pca\_emb\_dim=0, plateau\_factor=0.3, plateau\_patience=3, scheduler=plateau, weight\_decay=1e-06 \\
    Unemployment Plus & MLP & $X_{i, t}^{\mathcal{E}}$ & True & activation=gelu, batch\_size=1024, dropout=0.3, hidden=[512, 512, 512], lr=0.0003, max\_epochs=1000, patience=20, pca\_emb\_dim=0, plateau\_factor=0.3, plateau\_patience=5, scheduler=none, weight\_decay=0.0001 \\
    Unemployment Plus & MLP & $X_{i, t}^{\text{struct}}$ & True & activation=gelu, batch\_size=4096, dropout=0, hidden=[1024, 1024, 1024], lr=0.001, max\_epochs=1000, patience=20, pca\_emb\_dim=0, plateau\_factor=0.3, plateau\_patience=3, scheduler=plateau, weight\_decay=0.001 \\
    Unemployment Plus & Random Forest & $X_{i, t}^{\text{both}}$ & True & class\_weight=balanced, max\_depth=25, max\_features=0.4, min\_samples\_leaf=10, n\_estimators=800, pca\_emb\_dim=256 \\
    Unemployment Plus & Random Forest & $X_{i, t}^{\mathcal{E}}$ & True & class\_weight=balanced, max\_depth=25, max\_features=0.4, min\_samples\_leaf=10, n\_estimators=800, pca\_emb\_dim=0 \\
    Unemployment Plus & Random Forest & $X_{i, t}^{\text{struct}}$ & True & class\_weight=balanced, max\_depth=15, max\_features=log2, min\_samples\_leaf=10, n\_estimators=1600, pca\_emb\_dim=0 \\
    \end{longtable}
\end{landscape}
\normalsize

Table \ref{tab:alternative-metrics-performance-results-with-se} reports performance across four tasks: Coarse Occupation, Occupation Change, Unemployment, and Unemployment Plus, sorted by log-likelihood within each task.
LABOR-LLM attains the best log-likelihood in every task and consistently yields lower Brier scores.
Classification quality improves as well: AUC-ROC rises relative to the best baseline in each task, with corresponding gains in AUPRC and small but systematic improvements in accuracy.
Despite not being explicitly calibrated during training, LABOR-LLM achieves calibration that is competitive with the strongest baseline models with explicit calibration and markedly better than CAREER across all tasks.
For uncertainty quantification, we report test-set bootstrap standard errors with $B_{\text{test}}=100$ resamples for all metrics. In addition, for LABOR-LLM and the main baselines on the three binary prediction tasks, we report training-set bootstrap standard errors with $B_{\text{train}}=12$ resamples for all metrics (log-likelihood, Brier score, ECE, AUC-ROC, AUPRC, and accuracy) to capture variability from model re-fitting.
Overall, these results indicate that LABOR-LLM not only delivers superior predictions of the next occupation but also performs strongly on standard binary outcomes, such as job change and unemployment status, of broad interest in economics and the social sciences.

\tiny
\begin{landscape}
% ════════════════════════════════════════════════════════════════════════════════
% 📄 LaTeX Table (All Tasks Combined) [DEV_MODE=False]
% ════════════════════════════════════════════════════════════════════════════════
\begin{longtable}{lclllllll}
    \caption{Performance Evaluation Results Across All Tasks (Sorted by Log Likelihood Within Each Task)} \label{tab:alternative-metrics-performance-results-with-se} \\
    \toprule
    Prediction Task & Feature Set & Model & Log-likelihood ↑ & Brier Score ↓ & ECE ↓ & AUC-ROC ↑ & AUPRC ↑ & Accuracy ↑ \\
    \midrule
    \endfirsthead
    \caption[]{Performance Evaluation Results Across All Tasks (Sorted by Log Likelihood Within Each Task)} \\
    \toprule
    Prediction Task & Feature Set & Model & Log-likelihood ↑ & Brier Score ↓ & ECE ↓ & AUC-ROC ↑ & AUPRC ↑ & Accuracy ↑ \\
    \midrule
    \endhead
    \midrule
    \multicolumn{9}{r}{Continued on next page} \\
    \midrule
    \endfoot
    \bottomrule
    \endlastfoot
    Coarse Occupation & -- & LABOR-LLM & \shortstack{-0.8668 \\ (TestSE=0.0046)} & \shortstack{0.0592 \\ (TestSE=0.0003)} & \shortstack{0.0052 \\ (TestSE=0.0011)} & \shortstack{0.9164 \\ (TestSE=0.0009)} & \shortstack{0.6767 \\ (TestSE=0.0031)} & \shortstack{0.7117 \\ (TestSE=0.0019)} \\
    Coarse Occupation & -- & CAREER & \shortstack{-0.8877 \\ (TestSE=0.0050)} & \shortstack{0.0603 \\ (TestSE=0.0003)} & \shortstack{0.0114 \\ (TestSE=0.0010)} & \shortstack{0.9122 \\ (TestSE=0.0010)} & \shortstack{0.6667 \\ (TestSE=0.0036)} & \shortstack{0.7079 \\ (TestSE=0.0019)} \\
    Coarse Occupation & $X_{i, t}^{\text{both}}$ & MLP & \shortstack{-0.9247 \\ (TestSE=0.0049)} & \shortstack{0.0633 \\ (TestSE=0.0003)} & \shortstack{0.0040 \\ (TestSE=0.0010)} & \shortstack{0.9057 \\ (TestSE=0.0010)} & \shortstack{0.6338 \\ (TestSE=0.0035)} & \shortstack{0.6842 \\ (TestSE=0.0019)} \\
    Coarse Occupation & $X_{i, t}^{\mathcal{E}}$ & MLP & \shortstack{-0.9264 \\ (TestSE=0.0049)} & \shortstack{0.0634 \\ (TestSE=0.0003)} & \shortstack{0.0070 \\ (TestSE=0.0012)} & \shortstack{0.9053 \\ (TestSE=0.0010)} & \shortstack{0.6339 \\ (TestSE=0.0037)} & \shortstack{0.6845 \\ (TestSE=0.0019)} \\
    Coarse Occupation & $X_{i, t}^{\text{both}}$ & Logistic Regression & \shortstack{-0.9353 \\ (TestSE=0.0049)} & \shortstack{0.0636 \\ (TestSE=0.0003)} & \shortstack{0.0062 \\ (TestSE=0.0009)} & \shortstack{0.9018 \\ (TestSE=0.0011)} & \shortstack{0.6261 \\ (TestSE=0.0036)} & \shortstack{0.6878 \\ (TestSE=0.0019)} \\
    Coarse Occupation & $X_{i, t}^{\mathcal{E}}$ & Logistic Regression & \shortstack{-0.9363 \\ (TestSE=0.0050)} & \shortstack{0.0637 \\ (TestSE=0.0003)} & \shortstack{0.0060 \\ (TestSE=0.0010)} & \shortstack{0.9016 \\ (TestSE=0.0011)} & \shortstack{0.6257 \\ (TestSE=0.0036)} & \shortstack{0.6874 \\ (TestSE=0.0019)} \\
    Coarse Occupation & $X_{i, t}^{\text{both}}$ & Gradient Boosting & \shortstack{-1.0058 \\ (TestSE=0.0055)} & \shortstack{0.0691 \\ (TestSE=0.0004)} & \shortstack{0.0039 \\ (TestSE=0.0010)} & \shortstack{0.8873 \\ (TestSE=0.0013)} & \shortstack{0.5772 \\ (TestSE=0.0038)} & \shortstack{0.6463 \\ (TestSE=0.0022)} \\
    Coarse Occupation & $X_{i, t}^{\mathcal{E}}$ & Gradient Boosting & \shortstack{-1.0062 \\ (TestSE=0.0055)} & \shortstack{0.0691 \\ (TestSE=0.0004)} & \shortstack{0.0049 \\ (TestSE=0.0010)} & \shortstack{0.8871 \\ (TestSE=0.0013)} & \shortstack{0.5761 \\ (TestSE=0.0037)} & \shortstack{0.6475 \\ (TestSE=0.0022)} \\
    Coarse Occupation & $X_{i, t}^{\mathcal{E}}$ & Random Forest & \shortstack{-1.0765 \\ (TestSE=0.0057)} & \shortstack{0.0739 \\ (TestSE=0.0004)} & \shortstack{0.0203 \\ (TestSE=0.0018)} & \shortstack{0.8692 \\ (TestSE=0.0014)} & \shortstack{0.5379 \\ (TestSE=0.0041)} & \shortstack{0.6189 \\ (TestSE=0.0027)} \\
    Coarse Occupation & $X_{i, t}^{\text{both}}$ & Random Forest & \shortstack{-1.0796 \\ (TestSE=0.0058)} & \shortstack{0.0741 \\ (TestSE=0.0004)} & \shortstack{0.0177 \\ (TestSE=0.0018)} & \shortstack{0.8682 \\ (TestSE=0.0015)} & \shortstack{0.5354 \\ (TestSE=0.0041)} & \shortstack{0.6159 \\ (TestSE=0.0027)} \\
    Coarse Occupation & $X_{i, t}^{\text{struct}}$ & MLP & \shortstack{-1.4865 \\ (TestSE=0.0066)} & \shortstack{0.0983 \\ (TestSE=0.0003)} & \shortstack{0.0034 \\ (TestSE=0.0017)} & \shortstack{0.6959 \\ (TestSE=0.0024)} & \shortstack{0.2718 \\ (TestSE=0.0018)} & \shortstack{0.4423 \\ (TestSE=0.0033)} \\
    Coarse Occupation & $X_{i, t}^{\text{struct}}$ & Gradient Boosting & \shortstack{-1.4896 \\ (TestSE=0.0065)} & \shortstack{0.0983 \\ (TestSE=0.0003)} & \shortstack{0.0058 \\ (TestSE=0.0019)} & \shortstack{0.6934 \\ (TestSE=0.0024)} & \shortstack{0.2710 \\ (TestSE=0.0018)} & \shortstack{0.4428 \\ (TestSE=0.0034)} \\
    Coarse Occupation & $X_{i, t}^{\text{struct}}$ & Random Forest & \shortstack{-1.4948 \\ (TestSE=0.0067)} & \shortstack{0.0984 \\ (TestSE=0.0003)} & \shortstack{0.0062 \\ (TestSE=0.0018)} & \shortstack{0.6915 \\ (TestSE=0.0023)} & \shortstack{0.2702 \\ (TestSE=0.0018)} & \shortstack{0.4422 \\ (TestSE=0.0034)} \\
    Coarse Occupation & $X_{i, t}^{\text{struct}}$ & Logistic Regression & \shortstack{-1.5178 \\ (TestSE=0.0065)} & \shortstack{0.1006 \\ (TestSE=0.0003)} & \shortstack{0.0128 \\ (TestSE=0.0023)} & \shortstack{0.6767 \\ (TestSE=0.0029)} & \shortstack{0.2533 \\ (TestSE=0.0017)} & \shortstack{0.4367 \\ (TestSE=0.0034)} \\
    \midrule
    Occupation Change & -- & LABOR-LLM & \shortstack{-0.5512 \\ (TestSE=0.0016) \\ (TrainSE=0.0013)} & \shortstack{0.1865 \\ (TestSE=0.0007) \\ (TrainSE=0.0005)} & \shortstack{0.0100 \\ (TestSE=0.0012) \\ (TrainSE=0.0051)} & \shortstack{0.7862 \\ (TestSE=0.0015) \\ (TrainSE=0.0009)} & \shortstack{0.7284 \\ (TestSE=0.0023) \\ (TrainSE=0.0012)} & \shortstack{0.7171 \\ (TestSE=0.0015) \\ (TrainSE=0.0011)} \\
    Occupation Change & -- & CAREER & \shortstack{-0.5645 \\ (TestSE=0.0017)} & \shortstack{0.1912 \\ (TestSE=0.0007)} & \shortstack{0.0223 \\ (TestSE=0.0012)} & \shortstack{0.7762 \\ (TestSE=0.0015)} & \shortstack{0.7116 \\ (TestSE=0.0024)} & \shortstack{0.7095 \\ (TestSE=0.0015)} \\
    Occupation Change & $X_{i, t}^{\text{both}}$ & MLP & \shortstack{-0.5747 \\ (TestSE=0.0017) \\ (TrainSE=0.0032)} & \shortstack{0.1961 \\ (TestSE=0.0007) \\ (TrainSE=0.0013)} & \shortstack{0.0040 \\ (TestSE=0.0010) \\ (TrainSE=0.0013)} & \shortstack{0.7618 \\ (TestSE=0.0016) \\ (TrainSE=0.0041)} & \shortstack{0.6845 \\ (TestSE=0.0025) \\ (TrainSE=0.0054)} & \shortstack{0.6967 \\ (TestSE=0.0016) \\ (TrainSE=0.0034)} \\
    Occupation Change & $X_{i, t}^{\mathcal{E}}$ & MLP & \shortstack{-0.5769 \\ (TestSE=0.0017) \\ (TrainSE=0.0027)} & \shortstack{0.1965 \\ (TestSE=0.0007) \\ (TrainSE=0.0012)} & \shortstack{0.0041 \\ (TestSE=0.0011) \\ (TrainSE=0.0013)} & \shortstack{0.7609 \\ (TestSE=0.0016) \\ (TrainSE=0.0039)} & \shortstack{0.6824 \\ (TestSE=0.0025) \\ (TrainSE=0.0057)} & \shortstack{0.6962 \\ (TestSE=0.0016) \\ (TrainSE=0.0031)} \\
    Occupation Change & $X_{i, t}^{\text{both}}$ & Logistic Regression & \shortstack{-0.5832 \\ (TestSE=0.0017) \\ (TrainSE=0.0014)} & \shortstack{0.1992 \\ (TestSE=0.0006) \\ (TrainSE=0.0002)} & \shortstack{0.0042 \\ (TestSE=0.0011) \\ (TrainSE=0.0015)} & \shortstack{0.7539 \\ (TestSE=0.0016) \\ (TrainSE=0.0005)} & \shortstack{0.6729 \\ (TestSE=0.0025) \\ (TrainSE=0.0009)} & \shortstack{0.6906 \\ (TestSE=0.0015) \\ (TrainSE=0.0010)} \\
    Occupation Change & $X_{i, t}^{\mathcal{E}}$ & Logistic Regression & \shortstack{-0.5835 \\ (TestSE=0.0016) \\ (TrainSE=0.0013)} & \shortstack{0.1992 \\ (TestSE=0.0006) \\ (TrainSE=0.0002)} & \shortstack{0.0034 \\ (TestSE=0.0009) \\ (TrainSE=0.0013)} & \shortstack{0.7538 \\ (TestSE=0.0016) \\ (TrainSE=0.0005)} & \shortstack{0.6729 \\ (TestSE=0.0025) \\ (TrainSE=0.0010)} & \shortstack{0.6903 \\ (TestSE=0.0016) \\ (TrainSE=0.0010)} \\
    Occupation Change & $X_{i, t}^{\text{both}}$ & Gradient Boosting & \shortstack{-0.5939 \\ (TestSE=0.0015) \\ (TrainSE=0.0008)} & \shortstack{0.2045 \\ (TestSE=0.0006) \\ (TrainSE=0.0002)} & \shortstack{0.0045 \\ (TestSE=0.0013) \\ (TrainSE=0.0012)} & \shortstack{0.7381 \\ (TestSE=0.0017) \\ (TrainSE=0.0006)} & \shortstack{0.6558 \\ (TestSE=0.0027) \\ (TrainSE=0.0013)} & \shortstack{0.6752 \\ (TestSE=0.0016) \\ (TrainSE=0.0011)} \\
    Occupation Change & $X_{i, t}^{\mathcal{E}}$ & Gradient Boosting & \shortstack{-0.5949 \\ (TestSE=0.0016) \\ (TrainSE=0.0018)} & \shortstack{0.2045 \\ (TestSE=0.0006) \\ (TrainSE=0.0002)} & \shortstack{0.0043 \\ (TestSE=0.0012) \\ (TrainSE=0.0010)} & \shortstack{0.7382 \\ (TestSE=0.0017) \\ (TrainSE=0.0006)} & \shortstack{0.6563 \\ (TestSE=0.0028) \\ (TrainSE=0.0012)} & \shortstack{0.6760 \\ (TestSE=0.0016) \\ (TrainSE=0.0009)} \\
    Occupation Change & $X_{i, t}^{\text{both}}$ & Random Forest & \shortstack{-0.6064 \\ (TestSE=0.0014) \\ (TrainSE=0.0012)} & \shortstack{0.2101 \\ (TestSE=0.0006) \\ (TrainSE=0.0001)} & \shortstack{0.0033 \\ (TestSE=0.0010) \\ (TrainSE=0.0009)} & \shortstack{0.7215 \\ (TestSE=0.0019) \\ (TrainSE=0.0006)} & \shortstack{0.6325 \\ (TestSE=0.0028) \\ (TrainSE=0.0009)} & \shortstack{0.6640 \\ (TestSE=0.0017) \\ (TrainSE=0.0006)} \\
    Occupation Change & $X_{i, t}^{\mathcal{E}}$ & Random Forest & \shortstack{-0.6087 \\ (TestSE=0.0014) \\ (TrainSE=0.0010)} & \shortstack{0.2111 \\ (TestSE=0.0006) \\ (TrainSE=0.0002)} & \shortstack{0.0029 \\ (TestSE=0.0010) \\ (TrainSE=0.0018)} & \shortstack{0.7180 \\ (TestSE=0.0018) \\ (TrainSE=0.0009)} & \shortstack{0.6308 \\ (TestSE=0.0028) \\ (TrainSE=0.0011)} & \shortstack{0.6614 \\ (TestSE=0.0017) \\ (TrainSE=0.0009)} \\
    Occupation Change & $X_{i, t}^{\text{struct}}$ & Gradient Boosting & \shortstack{-0.6632 \\ (TestSE=0.0010) \\ (TrainSE=0.0012)} & \shortstack{0.2356 \\ (TestSE=0.0005) \\ (TrainSE=0.0001)} & \shortstack{0.0022 \\ (TestSE=0.0012) \\ (TrainSE=0.0014)} & \shortstack{0.6154 \\ (TestSE=0.0023) \\ (TrainSE=0.0008)} & \shortstack{0.5302 \\ (TestSE=0.0032) \\ (TrainSE=0.0011)} & \shortstack{0.5904 \\ (TestSE=0.0021) \\ (TrainSE=0.0008)} \\
    Occupation Change & $X_{i, t}^{\text{struct}}$ & MLP & \shortstack{-0.6634 \\ (TestSE=0.0009) \\ (TrainSE=0.0011)} & \shortstack{0.2357 \\ (TestSE=0.0004) \\ (TrainSE=0.0005)} & \shortstack{0.0020 \\ (TestSE=0.0012) \\ (TrainSE=0.0022)} & \shortstack{0.6142 \\ (TestSE=0.0022) \\ (TrainSE=0.0041)} & \shortstack{0.5297 \\ (TestSE=0.0031) \\ (TrainSE=0.0041)} & \shortstack{0.5906 \\ (TestSE=0.0021) \\ (TrainSE=0.0029)} \\
    Occupation Change & $X_{i, t}^{\text{struct}}$ & Random Forest & \shortstack{-0.6641 \\ (TestSE=0.0009) \\ (TrainSE=0.0008)} & \shortstack{0.2360 \\ (TestSE=0.0004) \\ (TrainSE=0.0001)} & \shortstack{0.0036 \\ (TestSE=0.0014) \\ (TrainSE=0.0009)} & \shortstack{0.6133 \\ (TestSE=0.0022) \\ (TrainSE=0.0007)} & \shortstack{0.5290 \\ (TestSE=0.0032) \\ (TrainSE=0.0012)} & \shortstack{0.5888 \\ (TestSE=0.0019) \\ (TrainSE=0.0010)} \\
    Occupation Change & $X_{i, t}^{\text{struct}}$ & Logistic Regression & \shortstack{-0.6757 \\ (TestSE=0.0010) \\ (TrainSE=0.0004)} & \shortstack{0.2410 \\ (TestSE=0.0004) \\ (TrainSE=0.0001)} & \shortstack{0.0085 \\ (TestSE=0.0019) \\ (TrainSE=0.0017)} & \shortstack{0.5799 \\ (TestSE=0.0024) \\ (TrainSE=0.0006)} & \shortstack{0.5006 \\ (TestSE=0.0030) \\ (TrainSE=0.0012)} & \shortstack{0.5823 \\ (TestSE=0.0022) \\ (TrainSE=0.0009)} \\
    \midrule
    Unemployment & -- & LABOR-LLM & \shortstack{-0.1654 \\ (TestSE=0.0022) \\ (TrainSE=0.0005)} & \shortstack{0.0430 \\ (TestSE=0.0007) \\ (TrainSE=0.0001)} & \shortstack{0.0016 \\ (TestSE=0.0004) \\ (TrainSE=0.0024)} & \shortstack{0.8110 \\ (TestSE=0.0036) \\ (TrainSE=0.0008)} & \shortstack{0.2445 \\ (TestSE=0.0087) \\ (TrainSE=0.0015)} & \shortstack{0.9494 \\ (TestSE=0.0010) \\ (TrainSE=0.0001)} \\
    Unemployment & $X_{i, t}^{\text{both}}$ & MLP & \shortstack{-0.1692 \\ (TestSE=0.0024) \\ (TrainSE=0.0007)} & \shortstack{0.0437 \\ (TestSE=0.0008) \\ (TrainSE=0.0000)} & \shortstack{0.0007 \\ (TestSE=0.0004) \\ (TrainSE=0.0004)} & \shortstack{0.8004 \\ (TestSE=0.0038) \\ (TrainSE=0.0017)} & \shortstack{0.2123 \\ (TestSE=0.0076) \\ (TrainSE=0.0022)} & \shortstack{0.9493 \\ (TestSE=0.0010) \\ (TrainSE=0.0001)} \\
    Unemployment & $X_{i, t}^{\mathcal{E}}$ & MLP & \shortstack{-0.1697 \\ (TestSE=0.0024) \\ (TrainSE=0.0007)} & \shortstack{0.0438 \\ (TestSE=0.0007) \\ (TrainSE=0.0001)} & \shortstack{0.0017 \\ (TestSE=0.0005) \\ (TrainSE=0.0004)} & \shortstack{0.7997 \\ (TestSE=0.0039) \\ (TrainSE=0.0013)} & \shortstack{0.2124 \\ (TestSE=0.0078) \\ (TrainSE=0.0019)} & \shortstack{0.9492 \\ (TestSE=0.0010) \\ (TrainSE=0.0001)} \\
    Unemployment & $X_{i, t}^{\text{both}}$ & Logistic Regression & \shortstack{-0.1698 \\ (TestSE=0.0023) \\ (TrainSE=0.0005)} & \shortstack{0.0437 \\ (TestSE=0.0008) \\ (TrainSE=0.0001)} & \shortstack{0.0015 \\ (TestSE=0.0004) \\ (TrainSE=0.0003)} & \shortstack{0.7986 \\ (TestSE=0.0039) \\ (TrainSE=0.0008)} & \shortstack{0.2149 \\ (TestSE=0.0078) \\ (TrainSE=0.0014)} & \shortstack{0.9492 \\ (TestSE=0.0010) \\ (TrainSE=0.0001)} \\
    Unemployment & -- & CAREER & \shortstack{-0.1701 \\ (TestSE=0.0025)} & \shortstack{0.0437 \\ (TestSE=0.0007)} & \shortstack{0.0051 \\ (TestSE=0.0006)} & \shortstack{0.7946 \\ (TestSE=0.0040)} & \shortstack{0.2245 \\ (TestSE=0.0073)} & \shortstack{0.9492 \\ (TestSE=0.0009)} \\
    Unemployment & $X_{i, t}^{\mathcal{E}}$ & Logistic Regression & \shortstack{-0.1707 \\ (TestSE=0.0023) \\ (TrainSE=0.0004)} & \shortstack{0.0438 \\ (TestSE=0.0008) \\ (TrainSE=0.0001)} & \shortstack{0.0019 \\ (TestSE=0.0005) \\ (TrainSE=0.0003)} & \shortstack{0.7970 \\ (TestSE=0.0039) \\ (TrainSE=0.0012)} & \shortstack{0.2109 \\ (TestSE=0.0077) \\ (TrainSE=0.0021)} & \shortstack{0.9492 \\ (TestSE=0.0010) \\ (TrainSE=0.0001)} \\
    Unemployment & $X_{i, t}^{\text{both}}$ & Gradient Boosting & \shortstack{-0.1742 \\ (TestSE=0.0024) \\ (TrainSE=0.0005)} & \shortstack{0.0444 \\ (TestSE=0.0008) \\ (TrainSE=0.0001)} & \shortstack{0.0015 \\ (TestSE=0.0005) \\ (TrainSE=0.0004)} & \shortstack{0.7857 \\ (TestSE=0.0039) \\ (TrainSE=0.0008)} & \shortstack{0.1943 \\ (TestSE=0.0067) \\ (TrainSE=0.0022)} & \shortstack{0.9491 \\ (TestSE=0.0011) \\ (TrainSE=0.0001)} \\
    Unemployment & $X_{i, t}^{\mathcal{E}}$ & Gradient Boosting & \shortstack{-0.1750 \\ (TestSE=0.0023) \\ (TrainSE=0.0006)} & \shortstack{0.0444 \\ (TestSE=0.0008) \\ (TrainSE=0.0001)} & \shortstack{0.0022 \\ (TestSE=0.0005) \\ (TrainSE=0.0005)} & \shortstack{0.7851 \\ (TestSE=0.0039) \\ (TrainSE=0.0015)} & \shortstack{0.1937 \\ (TestSE=0.0067) \\ (TrainSE=0.0013)} & \shortstack{0.9492 \\ (TestSE=0.0010) \\ (TrainSE=0.0001)} \\
    Unemployment & $X_{i, t}^{\mathcal{E}}$ & Random Forest & \shortstack{-0.1754 \\ (TestSE=0.0024) \\ (TrainSE=0.0004)} & \shortstack{0.0447 \\ (TestSE=0.0008) \\ (TrainSE=0.0000)} & \shortstack{0.0019 \\ (TestSE=0.0005) \\ (TrainSE=0.0005)} & \shortstack{0.7736 \\ (TestSE=0.0040) \\ (TrainSE=0.0013)} & \shortstack{0.1842 \\ (TestSE=0.0067) \\ (TrainSE=0.0013)} & \shortstack{0.9490 \\ (TestSE=0.0010) \\ (TrainSE=0.0001)} \\
    Unemployment & $X_{i, t}^{\text{both}}$ & Random Forest & \shortstack{-0.1760 \\ (TestSE=0.0024) \\ (TrainSE=0.0004)} & \shortstack{0.0448 \\ (TestSE=0.0008) \\ (TrainSE=0.0000)} & \shortstack{0.0014 \\ (TestSE=0.0005) \\ (TrainSE=0.0003)} & \shortstack{0.7717 \\ (TestSE=0.0039) \\ (TrainSE=0.0010)} & \shortstack{0.1815 \\ (TestSE=0.0064) \\ (TrainSE=0.0016)} & \shortstack{0.9490 \\ (TestSE=0.0010) \\ (TrainSE=0.0000)} \\
    Unemployment & $X_{i, t}^{\text{struct}}$ & Gradient Boosting & \shortstack{-0.1885 \\ (TestSE=0.0028) \\ (TrainSE=0.0002)} & \shortstack{0.0470 \\ (TestSE=0.0009) \\ (TrainSE=0.0000)} & \shortstack{0.0016 \\ (TestSE=0.0006) \\ (TrainSE=0.0007)} & \shortstack{0.7033 \\ (TestSE=0.0044) \\ (TrainSE=0.0014)} & \shortstack{0.1059 \\ (TestSE=0.0032) \\ (TrainSE=0.0006)} & \shortstack{0.9491 \\ (TestSE=0.0010) \\ (TrainSE=0.0000)} \\
    Unemployment & $X_{i, t}^{\text{struct}}$ & MLP & \shortstack{-0.1890 \\ (TestSE=0.0029) \\ (TrainSE=0.0004)} & \shortstack{0.0470 \\ (TestSE=0.0009) \\ (TrainSE=0.0000)} & \shortstack{0.0010 \\ (TestSE=0.0006) \\ (TrainSE=0.0003)} & \shortstack{0.7044 \\ (TestSE=0.0043) \\ (TrainSE=0.0012)} & \shortstack{0.1069 \\ (TestSE=0.0032) \\ (TrainSE=0.0012)} & \shortstack{0.9491 \\ (TestSE=0.0010) \\ (TrainSE=0.0000)} \\
    Unemployment & $X_{i, t}^{\text{struct}}$ & Logistic Regression & \shortstack{-0.1904 \\ (TestSE=0.0028) \\ (TrainSE=0.0001)} & \shortstack{0.0471 \\ (TestSE=0.0009) \\ (TrainSE=0.0000)} & \shortstack{0.0007 \\ (TestSE=0.0005) \\ (TrainSE=0.0002)} & \shortstack{0.6864 \\ (TestSE=0.0047) \\ (TrainSE=0.0008)} & \shortstack{0.0996 \\ (TestSE=0.0032) \\ (TrainSE=0.0006)} & \shortstack{0.9491 \\ (TestSE=0.0010) \\ (TrainSE=0.0000)} \\
    Unemployment & $X_{i, t}^{\text{struct}}$ & Random Forest & \shortstack{-0.1906 \\ (TestSE=0.0029) \\ (TrainSE=0.0008)} & \shortstack{0.0470 \\ (TestSE=0.0009) \\ (TrainSE=0.0000)} & \shortstack{0.0017 \\ (TestSE=0.0007) \\ (TrainSE=0.0004)} & \shortstack{0.6981 \\ (TestSE=0.0045) \\ (TrainSE=0.0017)} & \shortstack{0.1051 \\ (TestSE=0.0031) \\ (TrainSE=0.0007)} & \shortstack{0.9491 \\ (TestSE=0.0010) \\ (TrainSE=0.0000)} \\
    \midrule
    Unemployment Plus & -- & LABOR-LLM & \shortstack{-0.3645 \\ (TestSE=0.0022) \\ (TrainSE=0.0008)} & \shortstack{0.1122 \\ (TestSE=0.0008) \\ (TrainSE=0.0003)} & \shortstack{0.0058 \\ (TestSE=0.0008) \\ (TrainSE=0.0031)} & \shortstack{0.8992 \\ (TestSE=0.0012) \\ (TrainSE=0.0003)} & \shortstack{0.8566 \\ (TestSE=0.0024) \\ (TrainSE=0.0004)} & \shortstack{0.8500 \\ (TestSE=0.0013) \\ (TrainSE=0.0004)} \\
    Unemployment Plus & $X_{i, t}^{\text{both}}$ & MLP & \shortstack{-0.3756 \\ (TestSE=0.0023) \\ (TrainSE=0.0010)} & \shortstack{0.1163 \\ (TestSE=0.0009) \\ (TrainSE=0.0003)} & \shortstack{0.0031 \\ (TestSE=0.0008) \\ (TrainSE=0.0011)} & \shortstack{0.8927 \\ (TestSE=0.0013) \\ (TrainSE=0.0005)} & \shortstack{0.8424 \\ (TestSE=0.0026) \\ (TrainSE=0.0013)} & \shortstack{0.8433 \\ (TestSE=0.0015) \\ (TrainSE=0.0006)} \\
    Unemployment Plus & $X_{i, t}^{\mathcal{E}}$ & MLP & \shortstack{-0.3770 \\ (TestSE=0.0023) \\ (TrainSE=0.0008)} & \shortstack{0.1166 \\ (TestSE=0.0009) \\ (TrainSE=0.0003)} & \shortstack{0.0036 \\ (TestSE=0.0008) \\ (TrainSE=0.0006)} & \shortstack{0.8925 \\ (TestSE=0.0013) \\ (TrainSE=0.0005)} & \shortstack{0.8412 \\ (TestSE=0.0026) \\ (TrainSE=0.0009)} & \shortstack{0.8429 \\ (TestSE=0.0014) \\ (TrainSE=0.0006)} \\
    Unemployment Plus & -- & CAREER & \shortstack{-0.3774 \\ (TestSE=0.0025)} & \shortstack{0.1152 \\ (TestSE=0.0009)} & \shortstack{0.0126 \\ (TestSE=0.0008)} & \shortstack{0.8916 \\ (TestSE=0.0015)} & \shortstack{0.8480 \\ (TestSE=0.0026)} & \shortstack{0.8471 \\ (TestSE=0.0014)} \\
    Unemployment Plus & $X_{i, t}^{\text{both}}$ & Logistic Regression & \shortstack{-0.3819 \\ (TestSE=0.0024) \\ (TrainSE=0.0005)} & \shortstack{0.1178 \\ (TestSE=0.0009) \\ (TrainSE=0.0001)} & \shortstack{0.0042 \\ (TestSE=0.0008) \\ (TrainSE=0.0007)} & \shortstack{0.8882 \\ (TestSE=0.0013) \\ (TrainSE=0.0001)} & \shortstack{0.8372 \\ (TestSE=0.0026) \\ (TrainSE=0.0004)} & \shortstack{0.8429 \\ (TestSE=0.0014) \\ (TrainSE=0.0002)} \\
    Unemployment Plus & $X_{i, t}^{\mathcal{E}}$ & Logistic Regression & \shortstack{-0.3819 \\ (TestSE=0.0024) \\ (TrainSE=0.0004)} & \shortstack{0.1177 \\ (TestSE=0.0009) \\ (TrainSE=0.0001)} & \shortstack{0.0032 \\ (TestSE=0.0007) \\ (TrainSE=0.0006)} & \shortstack{0.8882 \\ (TestSE=0.0013) \\ (TrainSE=0.0001)} & \shortstack{0.8361 \\ (TestSE=0.0027) \\ (TrainSE=0.0005)} & \shortstack{0.8430 \\ (TestSE=0.0014) \\ (TrainSE=0.0002)} \\
    Unemployment Plus & $X_{i, t}^{\text{both}}$ & Gradient Boosting & \shortstack{-0.4002 \\ (TestSE=0.0025) \\ (TrainSE=0.0008)} & \shortstack{0.1254 \\ (TestSE=0.0009) \\ (TrainSE=0.0001)} & \shortstack{0.0035 \\ (TestSE=0.0008) \\ (TrainSE=0.0006)} & \shortstack{0.8773 \\ (TestSE=0.0015) \\ (TrainSE=0.0003)} & \shortstack{0.8213 \\ (TestSE=0.0029) \\ (TrainSE=0.0005)} & \shortstack{0.8293 \\ (TestSE=0.0016) \\ (TrainSE=0.0005)} \\
    Unemployment Plus & $X_{i, t}^{\mathcal{E}}$ & Gradient Boosting & \shortstack{-0.4016 \\ (TestSE=0.0025) \\ (TrainSE=0.0008)} & \shortstack{0.1256 \\ (TestSE=0.0010) \\ (TrainSE=0.0001)} & \shortstack{0.0033 \\ (TestSE=0.0010) \\ (TrainSE=0.0005)} & \shortstack{0.8770 \\ (TestSE=0.0015) \\ (TrainSE=0.0003)} & \shortstack{0.8204 \\ (TestSE=0.0029) \\ (TrainSE=0.0006)} & \shortstack{0.8288 \\ (TestSE=0.0016) \\ (TrainSE=0.0003)} \\
    Unemployment Plus & $X_{i, t}^{\text{both}}$ & Random Forest & \shortstack{-0.4237 \\ (TestSE=0.0025) \\ (TrainSE=0.0007)} & \shortstack{0.1339 \\ (TestSE=0.0010) \\ (TrainSE=0.0003)} & \shortstack{0.0025 \\ (TestSE=0.0010) \\ (TrainSE=0.0008)} & \shortstack{0.8615 \\ (TestSE=0.0016) \\ (TrainSE=0.0005)} & \shortstack{0.7973 \\ (TestSE=0.0030) \\ (TrainSE=0.0006)} & \shortstack{0.8166 \\ (TestSE=0.0018) \\ (TrainSE=0.0004)} \\
    Unemployment Plus & $X_{i, t}^{\mathcal{E}}$ & Random Forest & \shortstack{-0.4237 \\ (TestSE=0.0026) \\ (TrainSE=0.0015)} & \shortstack{0.1334 \\ (TestSE=0.0010) \\ (TrainSE=0.0001)} & \shortstack{0.0057 \\ (TestSE=0.0010) \\ (TrainSE=0.0011)} & \shortstack{0.8635 \\ (TestSE=0.0015) \\ (TrainSE=0.0003)} & \shortstack{0.7989 \\ (TestSE=0.0030) \\ (TrainSE=0.0006)} & \shortstack{0.8163 \\ (TestSE=0.0017) \\ (TrainSE=0.0006)} \\
    Unemployment Plus & $X_{i, t}^{\text{struct}}$ & Gradient Boosting & \shortstack{-0.5576 \\ (TestSE=0.0031) \\ (TrainSE=0.0005)} & \shortstack{0.1873 \\ (TestSE=0.0014) \\ (TrainSE=0.0002)} & \shortstack{0.0066 \\ (TestSE=0.0016) \\ (TrainSE=0.0010)} & \shortstack{0.7246 \\ (TestSE=0.0028) \\ (TrainSE=0.0007)} & \shortstack{0.6214 \\ (TestSE=0.0042) \\ (TrainSE=0.0012)} & \shortstack{0.7251 \\ (TestSE=0.0031) \\ (TrainSE=0.0007)} \\
    Unemployment Plus & $X_{i, t}^{\text{struct}}$ & Random Forest & \shortstack{-0.5580 \\ (TestSE=0.0031) \\ (TrainSE=0.0006)} & \shortstack{0.1877 \\ (TestSE=0.0014) \\ (TrainSE=0.0002)} & \shortstack{0.0051 \\ (TestSE=0.0014) \\ (TrainSE=0.0017)} & \shortstack{0.7234 \\ (TestSE=0.0029) \\ (TrainSE=0.0006)} & \shortstack{0.6202 \\ (TestSE=0.0042) \\ (TrainSE=0.0015)} & \shortstack{0.7244 \\ (TestSE=0.0031) \\ (TrainSE=0.0007)} \\
    Unemployment Plus & $X_{i, t}^{\text{struct}}$ & MLP & \shortstack{-0.5580 \\ (TestSE=0.0030) \\ (TrainSE=0.0007)} & \shortstack{0.1875 \\ (TestSE=0.0014) \\ (TrainSE=0.0002)} & \shortstack{0.0073 \\ (TestSE=0.0017) \\ (TrainSE=0.0010)} & \shortstack{0.7240 \\ (TestSE=0.0028) \\ (TrainSE=0.0007)} & \shortstack{0.6200 \\ (TestSE=0.0042) \\ (TrainSE=0.0013)} & \shortstack{0.7241 \\ (TestSE=0.0031) \\ (TrainSE=0.0008)} \\
    Unemployment Plus & $X_{i, t}^{\text{struct}}$ & Logistic Regression & \shortstack{-0.5842 \\ (TestSE=0.0033) \\ (TrainSE=0.0012)} & \shortstack{0.1982 \\ (TestSE=0.0014) \\ (TrainSE=0.0003)} & \shortstack{0.0094 \\ (TestSE=0.0020) \\ (TrainSE=0.0010)} & \shortstack{0.6834 \\ (TestSE=0.0032) \\ (TrainSE=0.0006)} & \shortstack{0.5619 \\ (TestSE=0.0044) \\ (TrainSE=0.0020)} & \shortstack{0.7089 \\ (TestSE=0.0033) \\ (TrainSE=0.0025)} \\
    \end{longtable}
\end{landscape}
\normalsize